\documentclass[lettersize,journal]{IEEEtran}
\usepackage{amsmath,amsfonts}
\usepackage{algorithmic}
\usepackage{algorithm}
\usepackage{array}
\usepackage{textcomp}
\usepackage{stfloats}
\usepackage{url}
\usepackage{verbatim}
\usepackage{graphicx}
\usepackage{cite}
\usepackage{booktabs}
\usepackage{bm}
\usepackage{orcidlink}
\usepackage{hyperref}
\usepackage{subcaption}
\usepackage{xcolor}
\definecolor{bordergreen}{RGB}{0,155,0}  % Adjust RGB values as needed
\definecolor{borderred}{RGB}{225,0,0}    % Adjust RGB values as needed

\begin{document}

\title{SHAZAM: Self-Supervised Change Monitoring for \\ Hazard Detection and Mapping}

\author{Samuel Garske \orcidlink{0000-0002-0622-8255}, \IEEEmembership{Graduate Student Member, IEEE,}
        Konrad Heidler \orcidlink{0000-0001-8226-0727},
        Bradley Evans \orcidlink{0000-0001-6675-3118},
        KC Wong \orcidlink{0000-0002-4977-5611},\\
        %Lichao Mou,
        and Xiao Xiang Zhu \orcidlink{0000-0001-8107-9096}, \IEEEmembership{Fellow, IEEE} % <-this % stops a space
        % <-this % stops a space
%\thanks{Manuscript received April 19, 2021; revised August 16, 2021.}
\thanks{This work has been submitted to the IEEE for possible publication. Copyright may be transferred without notice, after which this version may no longer be accessible. This work was supported in part by the German Federal Ministry of Education and Research (BMBF) under Grant 01DD20001 through the international future AI lab “AI4EO – Artificial Intelligence for Earth Observation: Reasoning, Uncertainties, Ethics and Beyond,” which funded Samuel Garske's beyond fellowship at TUM. This work was also supported by the Australian Research Council through the Industrial Transformation Training Centre grant IC170100023 that funds the Australian Research Council (ARC) Training Centre for CubeSats, UAVs \& Their Applications (CUAVA). The work of Konrad Heidler was supported by the German Federal Ministry for Economic Affairs and Climate Action under Grant 50EE2201C in the framework of the “National Center of Excellence ML4Earth,” and by the Munich Center for Machine Learning. The work of Xiao Xiang Zhu was supported by the Munich Center for Machine Learning.}% <-this % stops a space
\thanks{Samuel Garske and KC Wong are with the School of Aerospace, Mechanical, and Mechatronic Engineering, The University of Sydney, NSW 2006, Australia (e-mail: sam.garske@sydney.edu.au). Bradley Evans is with the School of Environmental and Rural Science, The University of New England, Armidale, NSW 2350, Australia. Samuel Garske, Bradley Evans, and KC Wong are also with the Australian Research Council (ARC) Training Centre for CubeSats, UAVs \& Their Applications (CUAVA).}% <-this % stops a space
%\thanks{Bradley Evans is with the School of Environmental and Rural Science, The University of New England, Armidale, NSW 2350, Australia.}%
%\thanks{Samuel Garske, Bradley Evans, and KC Wong are also with the Australian Research Council (ARC) Training Centre for CubeSats, UAVs \& Their Applications (CUAVA).}
\thanks{Konrad Heidler, and Xiao Xiang Zhu are with the Chair of Data Science in Earth Observation (SiPEO), Department of Aerospace and Geodesy, School of Engineering and Design and Munich Center for Machine Learning, Technical University of Munich (TUM), 80333 Munich, Germany.}
}

%\markboth{Journal of \LaTeX\ Class Files,~Vol.~XX, No.~X, January~2025}%


\maketitle

\begin{abstract}
The increasing frequency of environmental hazards due to climate change underscores the urgent need for effective monitoring systems. Current approaches either rely on expensive labelled datasets, struggle with seasonal variations, or require multiple observations for confirmation (which delays detection). To address these challenges, this work presents SHAZAM - Self-Supervised Change Monitoring for Hazard Detection and Mapping. SHAZAM uses a lightweight conditional UNet to generate expected images of a region of interest (ROI) for any day of the year, allowing for the direct modelling of normal seasonal changes and the ability to distinguish potential hazards. A modified structural similarity measure compares the generated images with actual satellite observations to compute region-level anomaly scores and pixel-level hazard maps. Additionally, a theoretically grounded seasonal threshold eliminates the need for dataset-specific optimisation. Evaluated on four diverse datasets that contain bushfires (wildfires), burned regions, extreme and out-of-season snowfall, floods, droughts, algal blooms, and deforestation, SHAZAM achieved F1 score improvements of between 0.066 and 0.234 over existing methods. This was achieved primarily through more effective hazard detection (higher recall) while using only 473K parameters. SHAZAM demonstrated superior mapping capabilities through higher spatial resolution and improved ability to suppress background features while accentuating both immediate and gradual hazards. SHAZAM has been established as an effective and generalisable solution for hazard detection and mapping across different geographical regions and a diverse range of hazards. The Python code is available at: \href{https://github.com/WiseGamgee/SHAZAM}{https://github.com/WiseGamgee/SHAZAM}.
\end{abstract}


\newpage

\section{Introduction}

\IEEEPARstart{E}{nvironmental} hazards are any natural or human-induced phenomenon that has the potential to cause harm or damage to communities, ecosystems, or the environment as a whole. Common examples include floods, storms, landslides, oil spills, earthquakes, and bushfires (wildfires). Methods that automatically detect hazards allow organisations to actively monitor a region of interest (ROI), and mapping them provides crucial geographic information needed to respond to and manage their impact \cite{abid2021Mapping4ImpactManagement}. ROIs can include areas prone to hazards, regions that require environmental protection, or remote areas that lack ground-based monitoring infrastructure. Monitoring and managing environmental hazards is of increasing importance as hazardous events are becoming more frequent in a changing climate \cite{IPCC2021IncreasedFrequency}. This urgency is highlighted by the United Nations' ``Early Warnings for All" initiative, which aims to achieve global coverage by 2027, addressing the critical gap where one-third of the world's population lacks access to adequate warning systems for natural hazards \cite{UN2022EarlyWarnings}.

\begin{figure}[h!]
    \centering
    
    % Row labels
    \textbf{Satellite Image Time Series (SITS)}
    \vspace{0.3em}
    
    % Top row with RGB images
    \begin{subfigure}[b]{\linewidth}
        \begin{subfigure}[b]{0.32\textwidth}
            \centering
            \includegraphics[width=\textwidth]{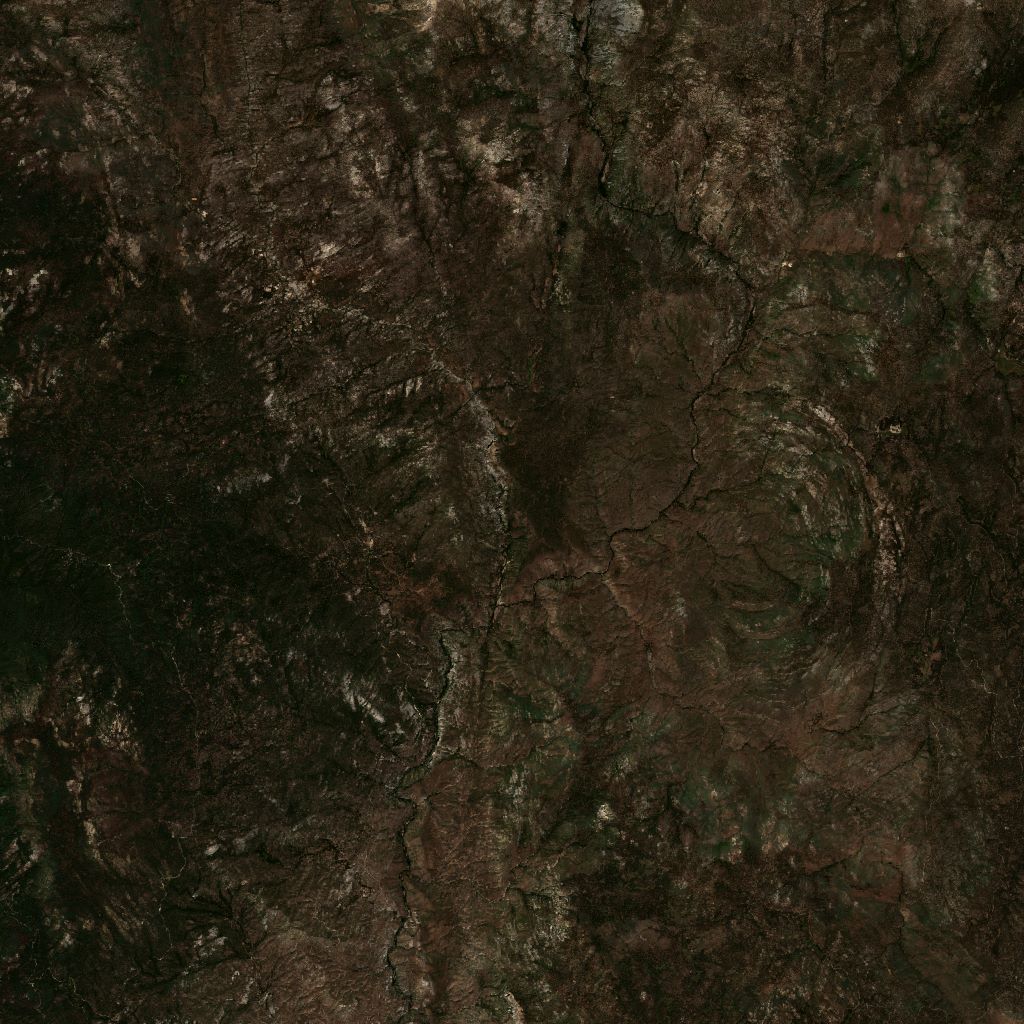}
            1 Aug, 2020 \\
            \textit{Normal}
        \end{subfigure}
        \hfill
        \begin{subfigure}[b]{0.32\textwidth}
            \centering
            \includegraphics[width=\textwidth]{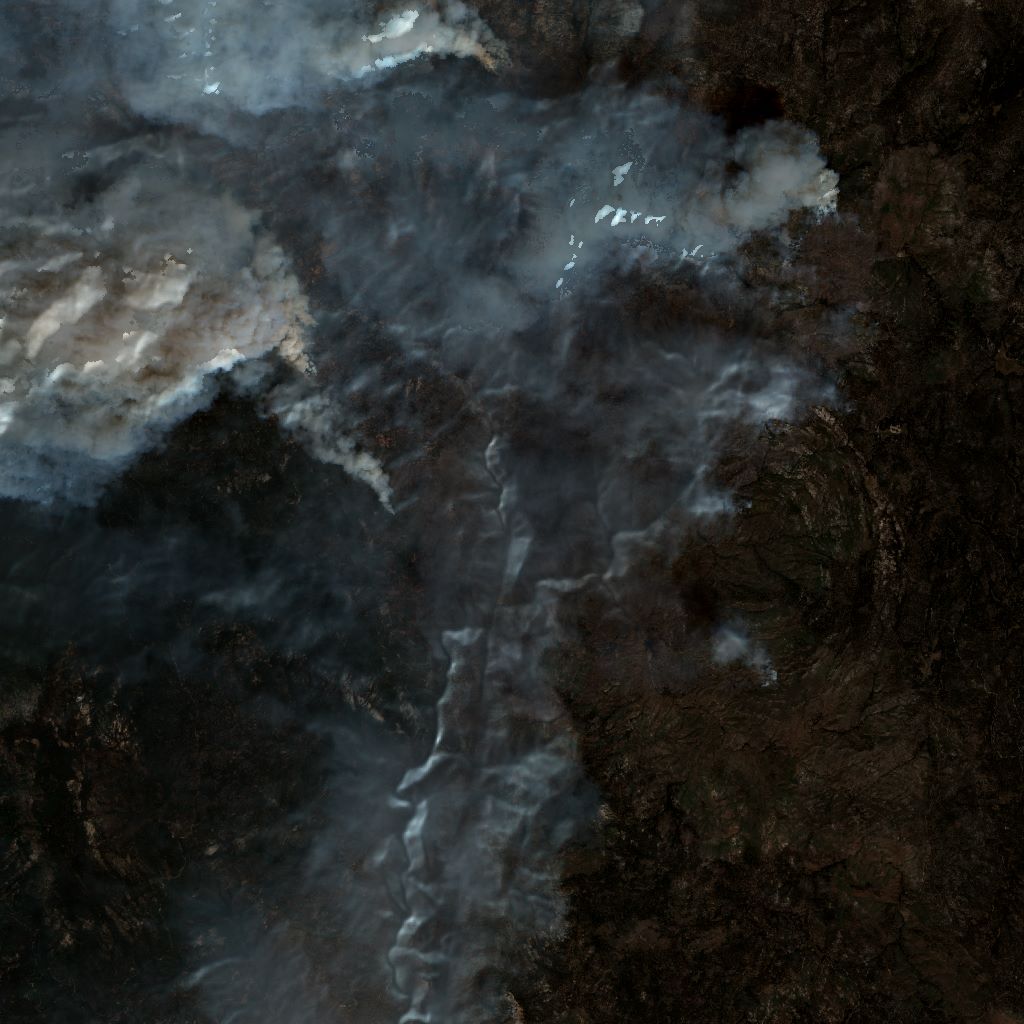}
            3 Sep, 2020 \\
            \textit{Bushfire}
        \end{subfigure}
        \hfill
        \begin{subfigure}[b]{0.32\textwidth}
            \centering
            \includegraphics[width=\textwidth]{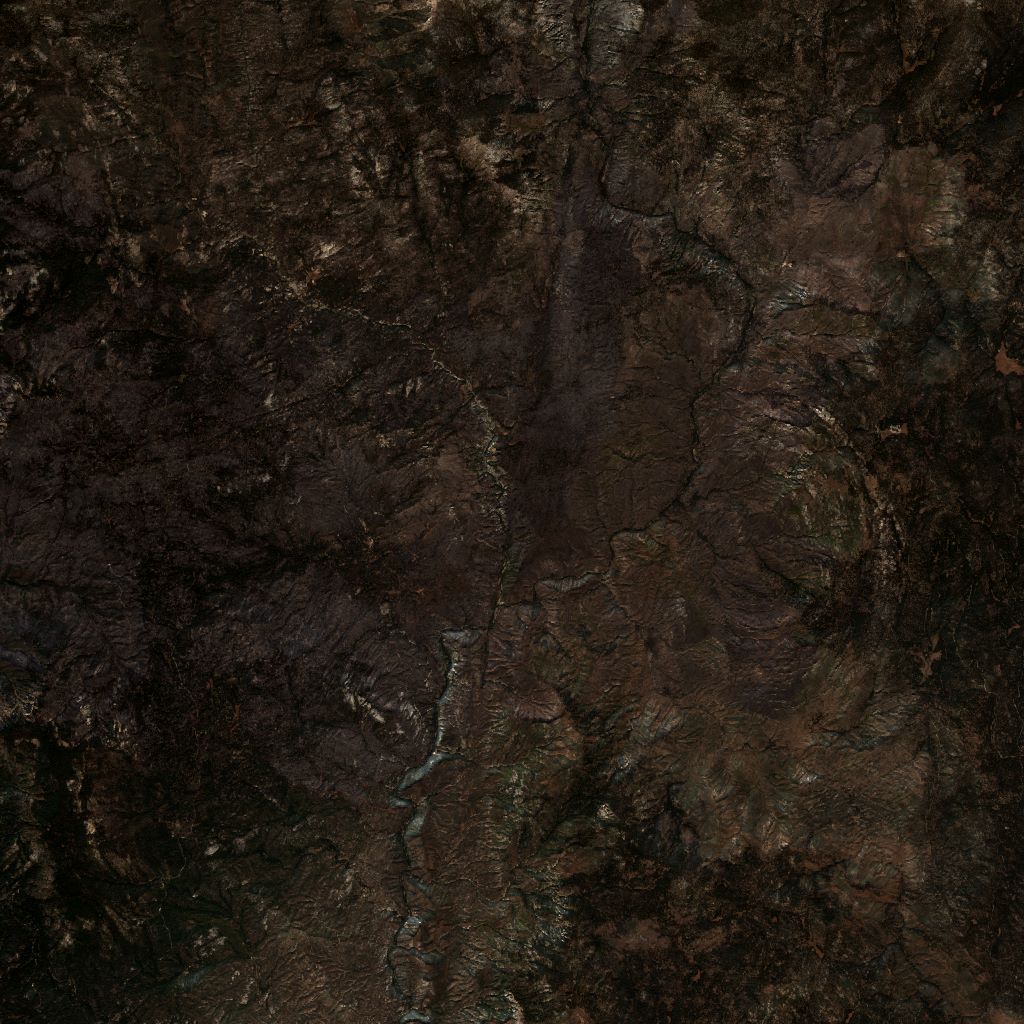}
            20 Oct, 2020 \\
            \textit{Burned Region}
        \end{subfigure}
    \end{subfigure}
    
    \vspace{0.8em}
    \textbf{SHAZAM Hazard Heatmaps}
    \vspace{0.3em}

    \begin{subfigure}[b]{\linewidth}
    \begin{subfigure}[b]{0.32\textwidth}
        \centering
        \setlength{\fboxsep}{0pt}  % Remove padding between frame and image
        \setlength{\fboxrule}{1.5pt}  % Set border thickness
        {\color{bordergreen}\fbox{\includegraphics[width=\textwidth]{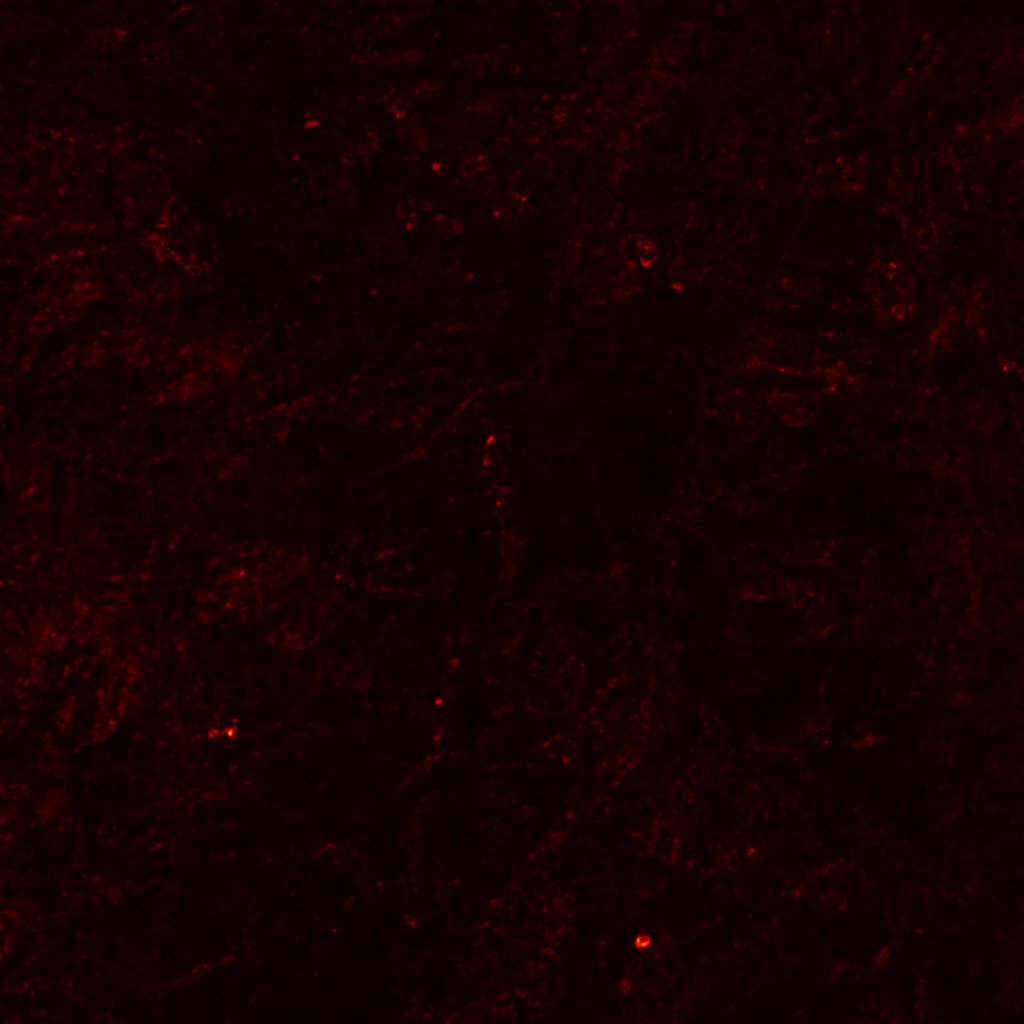}}}
        1 Aug, 2020 \\
        \textit{Normal}
    \end{subfigure}
    \hfill
    \begin{subfigure}[b]{0.32\textwidth}
        \centering
        \setlength{\fboxsep}{0pt}
        \setlength{\fboxrule}{1.5pt}
        {\color{borderred}\fbox{\includegraphics[width=\textwidth]{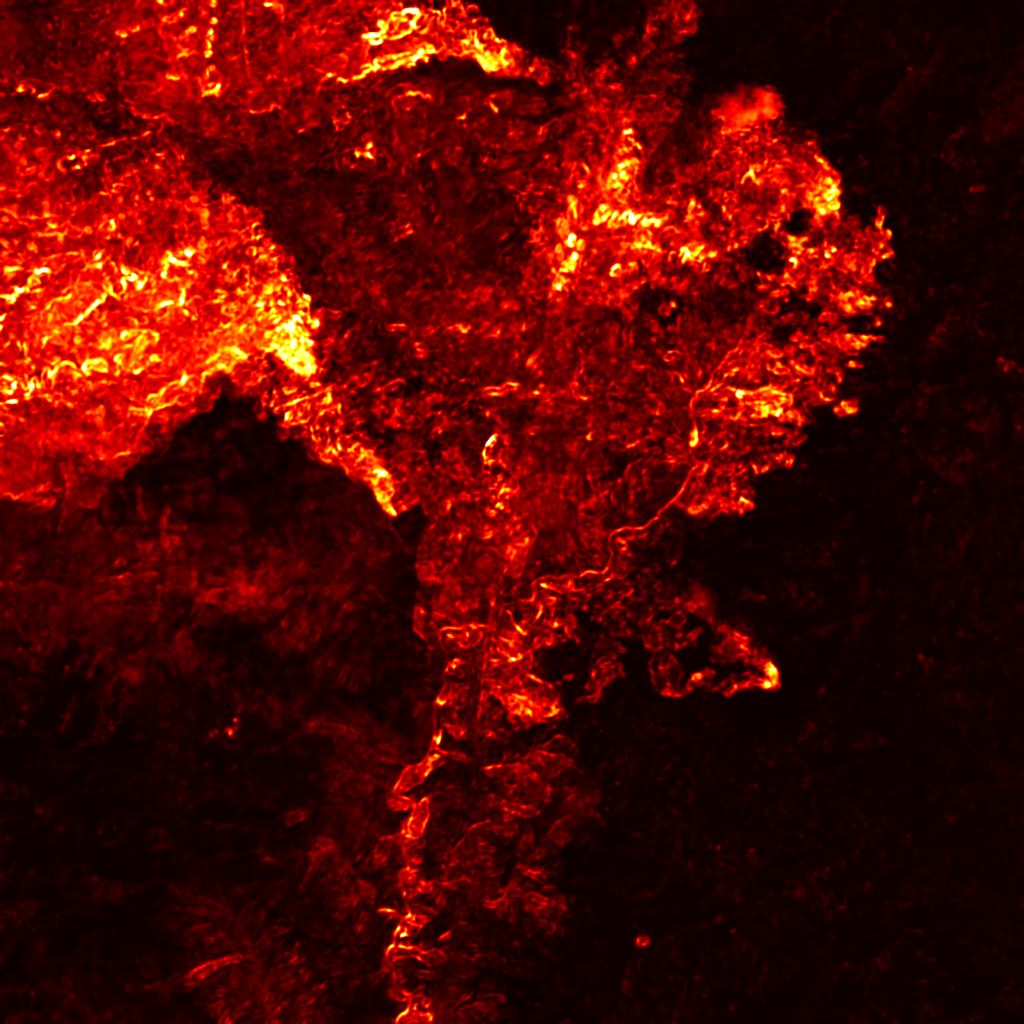}}}
        3 Sep, 2020 \\
        \textit{Hazard}
    \end{subfigure}
    \hfill
    \begin{subfigure}[b]{0.32\textwidth}
        \centering
        \setlength{\fboxsep}{0pt}
        \setlength{\fboxrule}{1.5pt}
        {\color{borderred}\fbox{\includegraphics[width=\textwidth]{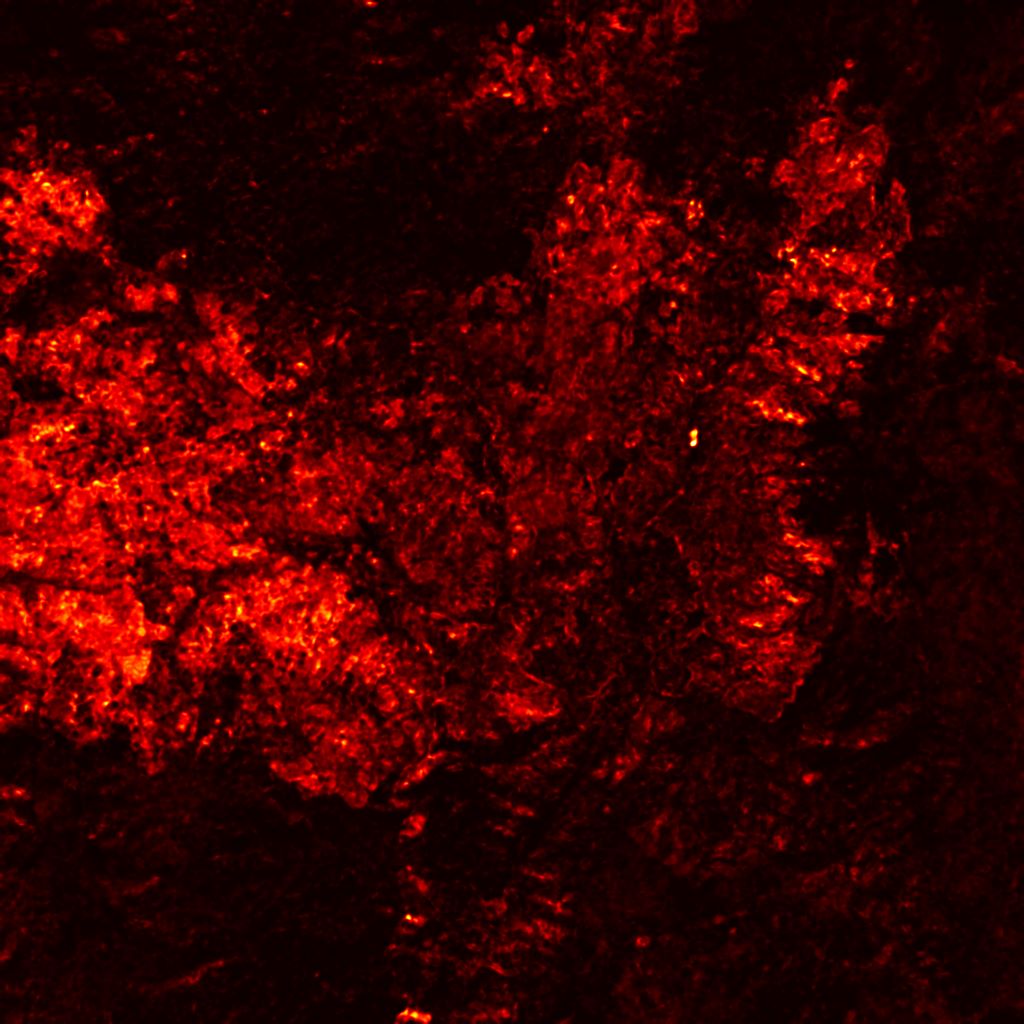}}}
        20 Oct, 2020 \\
        \textit{Hazard}
    \end{subfigure}
\end{subfigure}

    \vspace{0.6em}

    \begin{minipage}{\linewidth}
        \centering
        \includegraphics[width=\textwidth]{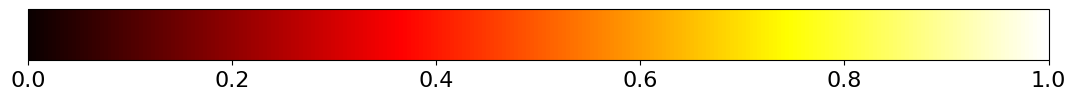}
    \end{minipage}%
    
    \caption{This Sentinel-2 SITS example shows a bushfire (wildfire) event in Sequoia National Park, California, progressing from normal conditions (left), to active bushfires (middle), and to burned regions (right). The corresponding hazard detection heatmaps were created by the proposed method, SHAZAM.}
    \label{fig:SHAZAM_intro}
\end{figure}

Satellites with onboard cameras are an effective tool for remotely monitoring an ROI, given their regular coverage of much of the Earth's surface. Regularly collecting images of the same ROI over time creates a satellite image time series (SITS), providing a temporal record of the landscape that can be used to identify unexpected changes and hazards (Figure \ref{fig:SHAZAM_intro}). The large and growing amount of available global historical SITS data provides unprecedented land monitoring capabilities, with missions such as Landsat and Sentinel-2 capturing high-resolution multispectral images \cite{goward2021landsathistory, sentinel2014missions}. Synthetic aperture radar (SAR) satellites like Sentinel-1 are also a powerful choice for hazard monitoring because they can penetrate cloud cover and capture images at night \cite{rambour2020SEN12-FLOOD}. Various methods for hazard detection and mapping with these satellite systems cover floods, landslides, deforestation, bushfires, and algal blooms \cite{rambour2020SEN12-FLOOD, casagli2023landslides, de2020Deforestation, rashkovetsky2021wildfires, caballero2020S2AlgalBlooms}. However, the sheer volume and complexity of SITS data have driven the development of automated computer vision and machine learning approaches for this task. 

Change detection has emerged as a fundamental approach for automated hazard monitoring using satellite imagery. Change detection is the process of identifying differences in the state of an object or phenomenon by observing it at different times \cite{singh1989ChangeDetectionDefinition}. In the context of remote sensing, change detection methods typically map the different pixels between two separate images of the same location. This approach is particularly relevant for hazard mapping, as models can learn to identify changes between satellite images that directly indicate a hazard.

However, current change detection approaches face several key limitations. Supervised methods require extensive labelled datasets, which are expensive and time consuming to create \cite{chi2016RSLabelQuality, tao2023RSLabelQuality}, particularly for the diverse range of potential hazards across different geographical locations. Unsupervised approaches eliminate the need for labels, but struggle to distinguish hazards from normal seasonal changes and are better suited to post-event mapping rather than active monitoring. Methods that attempt continuous monitoring, such as COLD \cite{zhu2020COLD} and NRT-MONITOR \cite{shang2022NRT-MONITOR}, address seasonal variations but are computationally intensive and require multiple post-hazard observations, hindering timely detection.

Recent advances in self-supervised learning and anomaly detection offer promising directions for addressing these challenges \cite{wang2022self}. These approaches can learn normal patterns from unlabelled data and identify deviations that may indicate hazards, as demonstrated in video anomaly detection \cite{jin2022andt, tran2024andt+}. However, adapting these methods to satellite imagery presents unique challenges, particularly in handling the irregular temporal sampling of SITS data and modelling seasonal dynamics throughout the year. To address these challenges, this work presents SHAZAM - a self-supervised change monitoring method for hazard detection and mapping. SHAZAM uses seasonal image translation to understand the normal dynamics of a region, allowing the detection and mapping of potential hazards without requiring labelled examples or multiple confirmatory observations. The contributions of this work are as follows:
\begin{enumerate}
    \item A seasonally integrated UNet architecture that learns to translate a baseline image of an ROI to an expected image for any day of the year. This enables direct differentiation between hazards and normal environmental changes.
    \item A modified structural similarity index measure (SSIM) that provides a region-wise anomaly score for detecting potential hazards, and a high resolution pixel-wise heatmap for mapping.
    \item A method for estimating a seasonal threshold to determine if an anomaly score is sufficiently high to indicate a potential hazard. The seasonal threshold varies across the year, as some seasons are more difficult to model due to larger variations in the landscape. This threshold is calculated using model performance on the training dataset.
    \item Extensive experimental validation on four new datasets across diverse geographical regions. These include bushfires, burned regions, extreme and out-of-season snowfall, algal blooms, drought, deforestation, and floods. These experiments demonstrate SHAZAM's effectiveness for hazard detection and mapping compared to similar approaches designed for monitoring.
\end{enumerate}

The remainder of this paper is organised as follows. Section \ref{sec:related_work} reviews related work on change detection and hazard mapping. Section \ref{sec:method} describes the proposed methodology, SHAZAM. Section \ref{sec:results} presents the experimental results and analysis. Finally, Section \ref{sec:conclusion} concludes the paper.

\section{Related Work} \label{sec:related_work}

Traditionally, change detection approaches encompassed algebra-based methods, transformation, classification, advanced models, GIS, and visual analysis \cite{lu2004TraditionalCDReview}. More recently, deep learning has become the dominant approach for supervised change detection due to its ability to learn complex spatial and spectral representations from satellite imagery \cite{shi2020AI4CDReview, shafique2022DL4CD, bai2023DL4ChangeDetection}. Supervised approaches span various architectures, such as encoder-decoder networks \cite{peng2019UNET++4CD, zheng2021CLNET4CD, wang2023WNet4CD}, Siamese networks \cite{daudt2018FCNNSiamese4CD, chicco2021SiameseNNOverview, bandara2022TransformerSiamese4CD, chen2020DASNetSiamese, chen2021BIT4CD, tang2023SMART4CD, yin2023SAGNet4CD, zhao2023Siam-DWENet4CD, yan2023HSSENET4CD, you2023CSViG4CD, zhou2023SIGNet4CD, li2023BuildingCD, yang2024BuildingCD}, generative adversarial networks \cite{niu2018cGAN4CD, du2023MTCDN, wu2023FCDGAN4CD}, recurrent neural networks \cite{lyu2016LSTM4CD, mou2018ReCNN4CD, chen2019LSTM4CD, khusni2020LSTM4CD, papadomanolaki2021LSTM4CD, shi2022LSTM4CD }, and lightweight architectures \cite{lei2023UltraLightweightCD, codegoni2023TinyCD, ding2024SAMCD}.

However, supervised methods face three key limitations for hazard monitoring. First, they typically rely on carefully selected bi-temporal image pairs, making them better suited for post-event mapping than active monitoring. Second, they lack mechanisms to handle seasonal variations, requiring images from similar seasons to avoid mistaking seasonal changes for hazards. Third, and most critically, they require extensive labelled datasets, which are expensive and time-consuming to create for the diverse range of potential hazards across different geographical locations \cite{chi2016RSLabelQuality, tao2023RSLabelQuality}. These limitations have motivated the development of unsupervised approaches, which is the focus of this work.

\subsection{Unsupervised Change Detection: Towards Continuous Monitoring}

Unsupervised methods identify changes between images without needing labelled change maps for training, eliminating the resource-intensive process of manual labelling. Traditional bi-temporal approaches use alternative learning objectives, such as the stacked sparse autoencoder proposed by \cite{touati2020sparseAEchangedetection}. This autoencoder reconstructs pixels from two sensor images, classifying changes based on reconstruction errors, the difference between model predictions and actual values, with changed pixels typically showing larger errors. Although avoiding labelled samples, it still depends on carefully chosen images from similar seasons, restricting continuous monitoring applications. Moving towards multi-temporal SITS methods, \cite{saha2020change} applied an LSTM framework using multiple SAR images to map flood events by reconstructing shuffled pixel time series sequences. Though representing progress in leveraging SITS, this approach lacked seasonal modelling and required selecting unaffected training regions retrospectively, making it more suitable for post-event mapping than monitoring.

Some methods now explicitly address seasonal variations in SITS. UTRNet \cite{yang2022UTRNet} embeds multi-year temporal information within a convolutional LSTM network, using both seasonal cycle position and temporal distance data with self-generated labels to differentiate genuine changes from seasonal variations. However, it requires costly retraining with all historical data for new observations, and its detection capacity is limited by its underlying pre-change model. Similarly, UCDFormer \cite{xu2023ucdformer} employs a lightweight transformer with domain adaptation, attempting to suppress seasonal effects through an affinity-weighted translation loss that measures both spectral distribution distances and structural content differences between images. This suppresses changes caused by seasonality, but still requires image selection near identified events rather than identifying the events themselves.

More recent work has focused on lightweight solutions suited to monitoring. RaVAEn \cite{ruuvzivcka2022ravaen} implements a variational autoencoder for onboard change event detection, compressing Sentinel-2 image patches to detect and map multiple hazards including floods, fires, landslides, and hurricanes. CLVAE \cite{yadav2024CLVAE} advances this using contrastive learning for SAR flood mapping, and similarly handling irregular capture times without labelled data. While these operational methods demonstrate progress for practical monitoring, they still lack robust seasonal integration to distinguish real changes from seasonal variations.

\subsection{Seasonal Modelling Through Disturbance Detection}

An area that provides insight into modelling seasonal dynamics when detecting potential hazards is disturbance detection, which can be considered a sub-field of unsupervised change detection. Disturbance detection has traditionally focused on detecting abrupt events that result in the sustained disruption of an ecosystem, community, or population \cite{ps1985DisturbanceDetection, turner2010DisturbanceDetection}. Methods such as the Breaks for Additive Seasonal and Trend algorithm (BFAST) \cite{verbesselt2010BFAST} and the Landsat-based detection of Trends in Disturbance and Recovery algorithm (LandTrendR) \cite{kennedy2010LandTrendR} model seasonal variations and long-term trends of a region using sequential multispectral satellite images, and can identify abrupt changes caused by natural or human-induced hazards. 

BFAST is a more statistical approach that identifies disturbances using breakpoints, which are times when the SITS-derived Normalised Difference Vegetation Index (NDVI) significantly alters from the modelled trend after the removal of seasonality. BFAST uses piecewise linear regression to model the NDVI trend and employs dummy variables to represent annual seasonality cycles. LandTrendR employs a geometric method to identify breakpoints in the NDVI and in the Normalised Burn Ratio (NBR) time series, using piecewise linear segments and angles between them to model trends. However, it does not directly model seasonality.

Disturbance-based methods have since been demonstrated for various applications, such as the detection and mapping of droughts, urbanisation, forest degradation and deforestation, land temperature anomalies, grassland mowing, impacts on coastal wetlands, and burned regions \cite{verbesselt2012NRTBFAST, zhu2014CCDC, ye2021DisturbanceDetectionKalman, hamunyela2016BFASTSpatial, li2022DisturbanceLandTemp, schwieder2022GrasslandDisturbance, yang2022CoastalWetlandDisturbance, fotakidis2023BurnedRegionMapping}. However, these approaches tend to use specific spectral indices and rule-based methods to detect disturbances in a particular biome. This limits their ability to detect and map multiple hazards across diverse geographic regions.

More generalised methods that use multiple spectral bands were developed for pixel-wise disturbance detection and monitoring such as COLD \cite{zhu2020COLD} and NRT-MONITOR \cite{shang2022NRT-MONITOR}. COLD uses a limited Fourier series to model intra-annual seasonality and inter-annual trends in Landsat images, predicting each band's reflectance values. The Fourier series models seasonal cycles over yearly, half-yearly, and four-month intervals. Deviations deemed statistically significant are identified as disturbances. NRT-MONITOR employs a similar Fourier series for seasonality on harmonised Landsat-Sentinel images but uses a forgetting factor for near real-time updates. This eliminates the requirement for storing complete historical time series of each pixel to identify potential hazards. It operates faster than COLD and flags disturbances with a shorter lag-time.

Despite efforts to develop faster near real-time methods \cite{zhu2020COLD, ye2021DisturbanceDetectionKalman, shang2022NRT-MONITOR, ye2023OBCOLD}, disturbance detection methods are extremely computationally demanding and are typically implemented on high-performance computers (HPCs). They also require multiple post-hazard observations to confirm a disturbance. This reduces false disturbance alarms, but comes at the cost of delayed hazard detection. This hinders the reaction of disaster and hazard management agencies, where a fast response time is critical. Pixel-wise approaches also lack the spatial context required for effective high-resolution image analysis \cite{hussain2013SpatialContextGood}. Few disturbance detection studies have investigated generalised spatially aware methods \cite{ye2023OBCOLD}, and none involve a framework that simultaneously flags multiple hazards at a regional level while mapping them at the pixel level.

\subsection{Self-Supervised Learning \& Anomaly Detection}

Although disturbance detection successfully models seasonality, self-supervised learning offers a more data-driven and efficient approach to change monitoring. Self-supervised learning is a subset of unsupervised learning in which a model is trained on a traditionally supervised task, such as regression or classification, without human-annotated labels \cite{he2020moco, misra2020selfsupervised, wang2022self}. Self-supervised learning often involves the use of machine-generated labels, contrastive learning, which separates different observations while clustering similar ones, or autoencoding, where models learn from reconstructing the input \cite{wang2022self}. Self-supervised learning has become popular for creating foundation models on large unlabelled datasets in remote sensing \cite{hong2024SpectralGPTFoundation, liu2024SSLRemoteClipFoundation, zhu2024SSLFoundationsRS}, which are later fine-tuned for downstream tasks with smaller and more niche datasets.

Self-supervised anomaly detection methods for videos offer a unique perspective towards change monitoring for an ROI, given their ability to identify unexpected events within time-series images. Anomaly detection methods are typically trained using only normal video footage, gaining a strong understanding of the expected dynamics within a given scene \cite{liu2018videoAD, chang2020videoAD, park2020videoAD,  zhang2020videoAD, huang2021videoAD, liu2021videoAD}. This could include normal traffic patterns for a certain intersection or normal human behaviour in CCTV footage. During testing, anomalous frames are flagged when the model struggles to represent them. \cite{yu2022videoADcontaminated} even demonstrated that the use of contaminated training data (that is, training data containing anomalous images) still performed well. This was based on the assumption that anomalies are rare occurrences and deep learning networks will bias themselves towards representing the majority of normal images. This is outlined as an important finding, as ensuring high-quality annotation is a difficult task in remote sensing \cite{chi2016RSLabelQuality, tao2023RSLabelQuality}.

In the realm of remote sensing, video-based anomaly detection methods have been adapted for aerial videos captured by drones. \cite{jin2022andt} proposed ANDT, a self-supervised hybrid transformer-convolutional network. ANDT uses six consecutive images to predict the subsequent image, using the prediction error to identify unexpected events. Images with prediction errors one standard deviation above the mean training error are classified as anomalies. The transformer-based encoder improved time-series feature mapping and future frame prediction, outperforming other well-known anomaly detection methods such as MemAE \cite{gong2019MAE} and GANomaly \cite{akcay2019ganomaly}. \cite{tran2024andt+} enhanced the approach by adding temporal cross-attention to the encoder outputs, improving spatial and temporal feature capture and surpassing ANDT's performance.

Although video-based detection methods offer promising directions, their direct application to SITS faces one major challenge. Unlike the uniform temporal sampling of video data, SITS typically presents in irregular intervals due to cloud cover and other environmental factors. This irregularity particularly affects transformer architectures, which rely on learning from dense, regular sequences to model temporal dependencies \cite{dosovitskiy2020ViT, sharir2021ViT4Video}.

Recent self-supervised transformer architectures for SITS have incorporated temporal encoding mechanisms, allowing models to handle the irregular and inconsistent timing between satellite image acquisitions \cite{yuan2020SITS-BERT, yuan2022SITS-Former, dumeur2024U-BARN}. These models employ input masking or corruption techniques during pre-training to learn spatial, spectral, and temporal features before tackling downstream supervised classification and segmentation tasks. A critical challenge in these approaches is to effectively encode seasonality. Most methods encode the day of the year using positional sine-cosine embeddings, borrowed from the original transformer architecture \cite{vaswani2017OGTransformer}. However, since these embeddings were originally designed to encode sequential positions rather than circular temporal patterns, they may not fully capture the cyclical nature of seasonal changes. An alternative approach, used by the Pretrained Remote Sensing Transformer (PRESTO), converts the month of image acquisition into cyclical coordinates \cite{tseng2023PRESTO}. This offers a more natural and direct embedding of seasonal patterns into the model, and is similar to the temporal modelling used in disturbance detection. It is clear that opportunities remain to develop more sophisticated seasonal embeddings specifically for anomaly detection.

Overall, the integration of self-supervised learning principles for SITS and video anomaly detection presents a promising direction for change monitoring. By modelling the normal spatio-temporal behaviour of an ROI across different seasons, self-supervised approaches could learn to distinguish genuine changes from regular seasonal variations without the need for labelled data. Directly embedding temporal information not only helps learn seasonal behaviour, but it offers a combined solution to address the challenge of irregularly spaced SITS.

\subsection{Summary}

This section's review of hazard detection and mapping approaches reveals four main directions: supervised change detection, unsupervised change detection, disturbance detection, and self-supervised video anomaly detection. Although supervised change detection methods provide a solid baseline for hazard mapping, they rely on costly and time-intensive human-annotated datasets and are better suited for post-event mapping than continuous monitoring. Unsupervised change detection approaches eliminate the need for labelled data and offer flexibility for detecting various types of hazards, but struggle to handle seasonal changes. Disturbance detection methods filter out seasonal changes but are computationally intensive and require multiple post-hazard observations, hindering timely detection. Self-supervised methods combined with video-based anomaly detection show promise for ROI-specific hazard monitoring, though challenges remain to handle the irregular capture times of SITS and model seasonal patterns. In the next section, we introduce SHAZAM to address these limitations.

\section{Proposed Method} \label{sec:method}

\subsection{SHAZAM Overview}

\begin{figure*}[t]
    \centering
    \includegraphics[width=\linewidth]{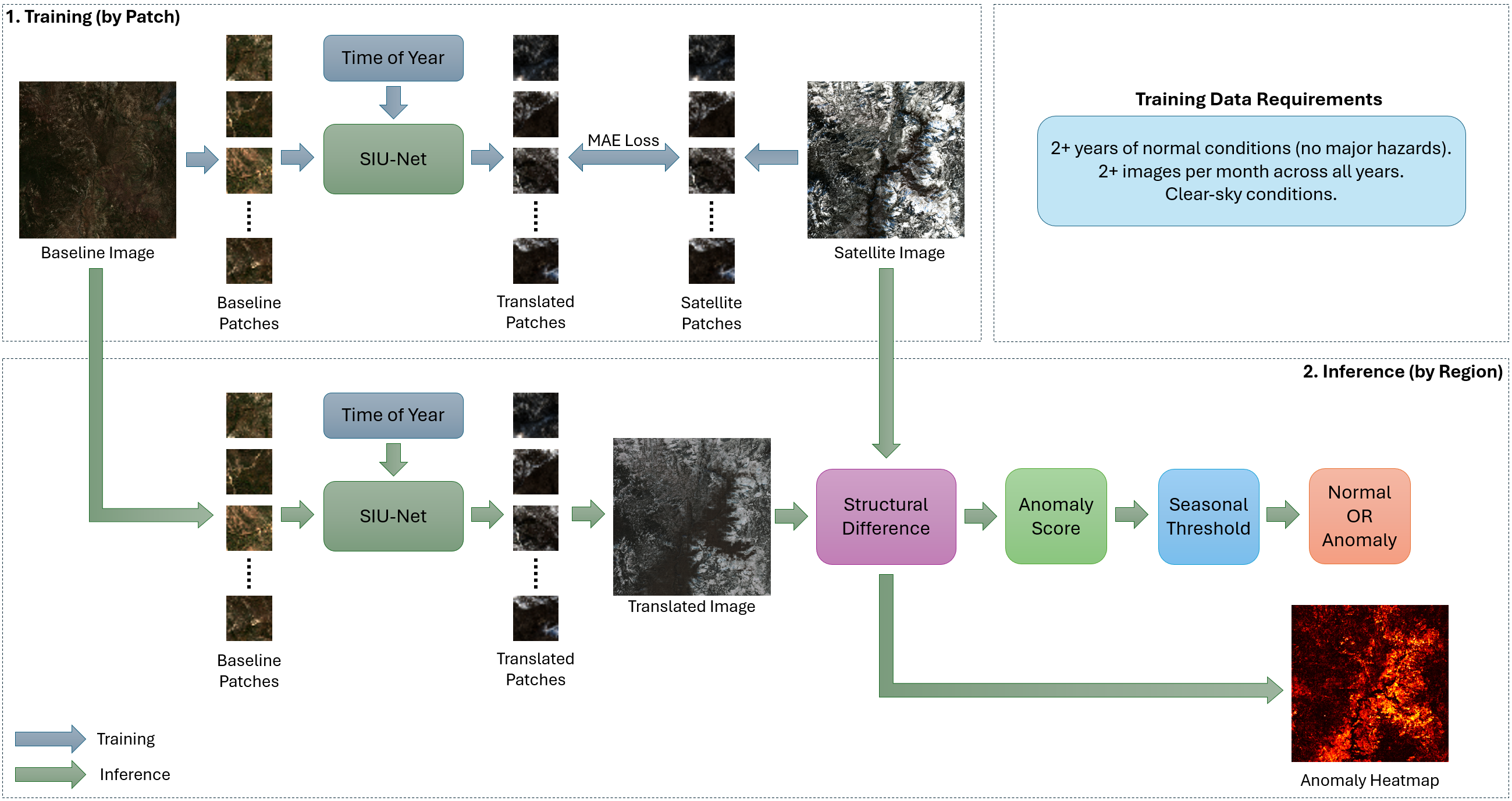}
    \caption[SHAZAM Overview]{SHAZAM - An overview of the proposed method. The top left visualises the training stage, the bottom the inference stage for monitoring an ROI, and the top right highlights training data requirements.}
    \label{fig:shazam}
\end{figure*}

This section introduces SHAZAM, a self-supervised change monitoring method for hazard detection and mapping (Figure \ref{fig:shazam}). By using 2+ years of satellite images without any major hazards present, SHAZAM learns the normal seasonal patterns of a region of interest (ROI). Using self-supervised learning, the seasonally integrated UNet model (SIU-Net) learns to translate a baseline image to each satellite image conditioned by the day of the year they were captured. It does this by each patch of 32x32 pixels, rather than ingesting the whole image at once. After training, SIU-Net can effectively generate an image of the ROI for any day of the year. 

In the inference phase (used for monitoring), SIU-Net generates an image of the ROI based on the same day of the year when a new satellite image is captured. A structural difference measure compares the satellite and SIU-Net images, producing an anomaly heatmap of unexpected pixel changes, serving as a hazard map. The average structural difference across all pixels in the anomaly heatmap gives the anomaly score, which is then used to detect a hazard. Due to seasonal variability in SIU-Net's translation performance, a seasonal error threshold is estimated from the training dataset and subtracted from the anomaly score. Final anomaly scores above zero indicate potential hazards, while those below zero are normal.

\subsection{Pre-Processing}

All 32×32 patches (both inputs and outputs) are normalised using a min-max scaling approach based on percentile values. The values of the 1st percentile ($x_1$) and 99th percentile ($x_{99}$) values are calculated using all patches in the training dataset (across all input channels). These same percentile values are then used to normalise the validation and test patches using the following equation:
\begin{equation}
x_{\text{norm}} = \frac{x - x_1}{x_{99} - x_1}, \quad x \in \mathbb{R}
\end{equation}

where $x$ represents the original reflectance value, $x_1$ and $x_{99}$ are the 1st and 99th percentiles of the training data, respectively, and $x_{\text{norm}}$ is the normalised value. After normalisation, values are clamped to the range [0, 1]:
\begin{equation}
x_{\text{final}} = \text{max}(0, \text{min}(1, x_{\text{norm}}))
\end{equation}

This robust normalisation approach effectively handles the heavy-tailed distributions typically present in satellite imagery, where extreme outliers (such as clouds, shadows, or strong reflections) could otherwise skew the scaling \cite{eoresearch2022NormalizeS2}. Using percentile-based normalisation instead of absolute minimum and maximum values helps preserve the meaningful variation in the data while reducing the impact of these outliers.

\subsection{Model Architecture}

\begin{figure*}[t]
    \centering
    \includegraphics[width=\linewidth]{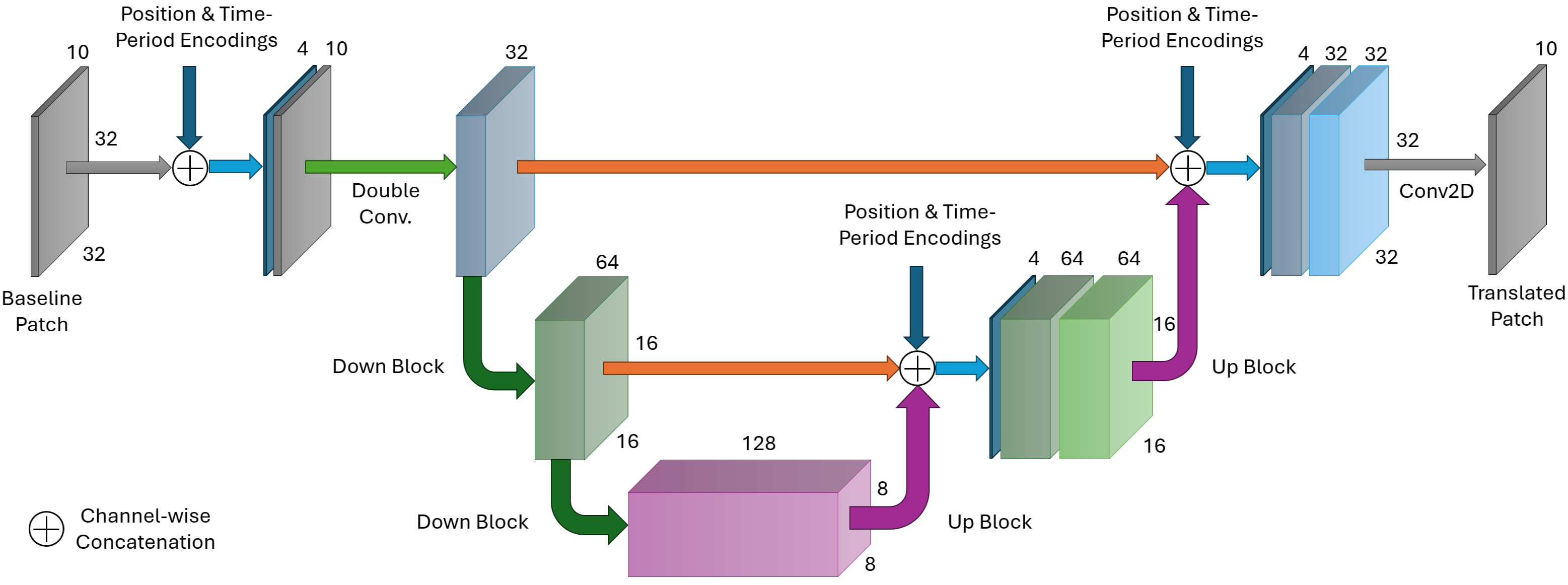}
    \caption{SIU-Net model architecture.}
    \label{fig:SIU-Net}
\end{figure*}

\begin{figure}[h!]
    \centering
    \includegraphics[width=\linewidth]{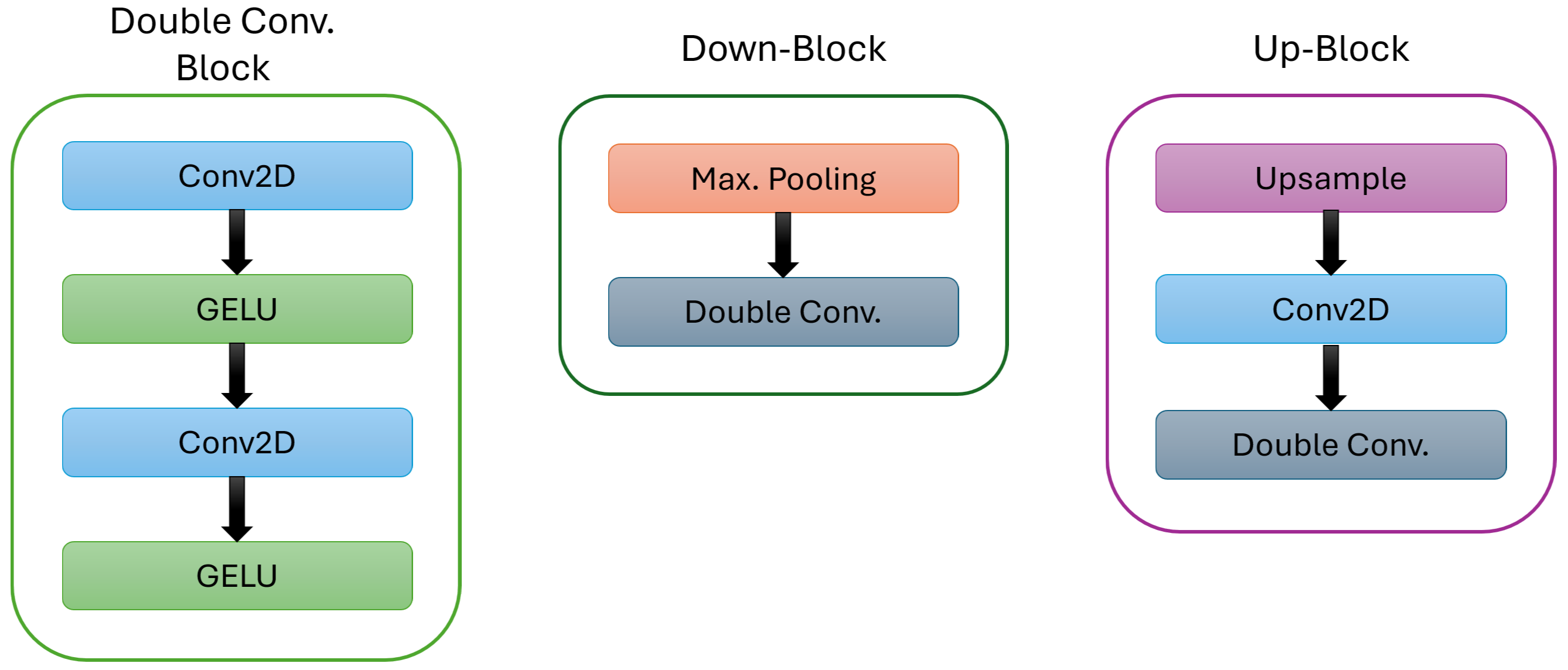}
    \caption{Network blocks.}
    \label{fig:SIU-Net_blocks}
\end{figure}

SHAZAM uses a conditional UNet-style model to translate 32x32 pixel baseline patches to their expected appearance for any day of the year, with an assumed 10 spectral channels for this work. This model is called Seasonally Integrated UNet (SIU-Net), and the architecture is shown in Figures \ref{fig:SIU-Net} and \ref{fig:SIU-Net_blocks}. UNet is an effective architecture for change detection because it preserves the spatial structure with skip connections while capturing features from multiple scales \cite{ronneberger2015UNetOG}. These two qualities are essential for providing high-quality, high-resolution outputs that accurately represent the ROI. 

The primary input of SIU-Net is a median patch calculated from the training dataset (excluding validation patches), referred to as a ``baseline" patch. This baseline patch provides the core spatial features of the region to SIU-Net (e.g., forests, rivers, lakes, mountains), serving as a reference point for temporal translation. This allows SIU-Net to act as a conditioned generative model, while enabling it to focus on seasonal transformation rather than memorising spatial features. It also allows for a much smaller model compared to fully-generative alternatives (e.g. cVAE, cGAN, stable diffusion), which would require significantly more parameters and data to accurately generate an ROI from scratch. 

To perform the seasonal translation efficiently, SIU-Net uses a lightweight architecture that integrates temporal information through time-position encodings concatenated as input channels. The network consists of a series of blocks (Figure \ref{fig:SIU-Net_blocks}); one double convolutional block (2×Conv2D with GELU), two down-blocks (MaxPool + double convolutional block) and two up-blocks (bilinear upsampling + double convolutional block), with skip connections between feature spaces. All double convolutional block layers use 3x3 kernels with a stride of 1 and a padding of 1. The MaxPool layers use 2x2 kernels and a stride of 2. Feature channels progressively expand from 32, to 64, to 128 through the down-blocks before returning to the 32x32x10 input dimensions via the up-blocks and a final Conv2D projection layer. The final Conv2D layer uses 1x1 kernels with a stride of 1 to reduce the concatenated feature maps, residuals, and time-position encodings to 10 channels. Skip connections play a dual role by concatenating both time-position encodings and residual features after each up-block, reinforcing temporal context while preserving the key spatial features of the patch during translation. SIU-Net uses the L1 loss function during training, also known as the Mean Absolute Error (MAE), which calculates the absolute difference between each translated patch and the true values.

\subsection{Season \& Position Encoding}

\begin{figure}[h!]
    \centering
    \includegraphics[width=0.9\linewidth]{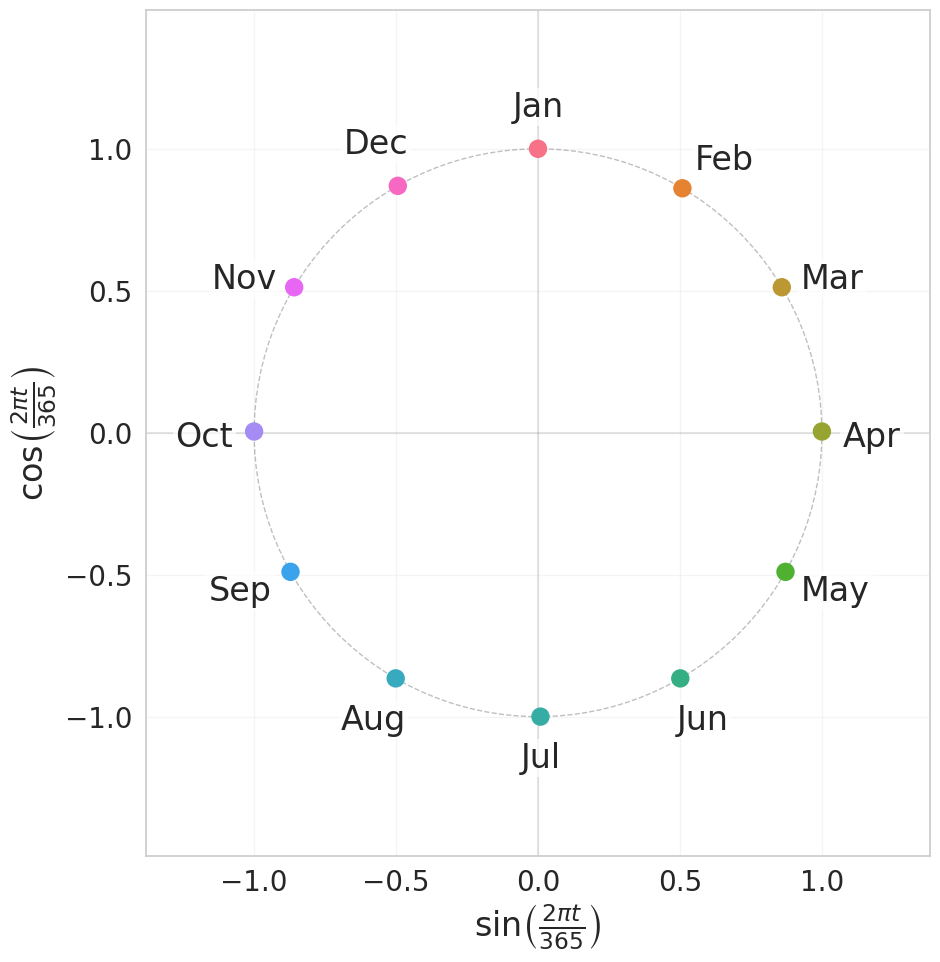}
    \caption[Seasonal Encodings]{Visualising the seasonal encodings for the day of the year, $t$.}
    \label{fig:seasonal_encodings}
\end{figure}

Capturing seasonal changes is one of the primary research gaps for change monitoring approaches. For SHAZAM, seasonality is injected into SIU-Net by converting the day of the year that an image was captured to a cyclical representation:
\begin{align}
    t_{\text{sin}} &= \sin(\frac{2 \pi t}{365})\\
    t_{\text{cos}} &= \cos(\frac{2 \pi t}{365})   
\end{align}

where $t$ is the current day of the year. These encodings directly capture the cyclical nature of seasonality, inferring that a day in late December (e.g., day 350) is near early January (day 5). This is illustrated in Figure \ref{fig:seasonal_encodings}. Another feature of these encodings is that they enable SIU-Net to interpolate between days that do not have corresponding images in the training set. This is crucial for handling irregular capture times with only a few years of historical SITS. Overall, these seasonal encodings allow SHAZAM to model both seasonal changes while handling irregular satellite observations.

The position encodings denote the location of each patch within the ROI image using normalised row and column coordinates:
\begin{align}
p_{\text{row}} &= \frac{\text{row}}{n_{\text{patches}}} \\
p_{\text{col}} &= \frac{\text{col}}{n_{\text{patches}}}
\end{align}

where $n_{\text{patches}}$ is the number of patches per dimension, assuming a square image. Position encoding allows SIU-Net to better model seasonal dynamics on a local level, enhancing image translation for each individual patch. This relative coordinate system provides a foundation for potential model extensions to handle multiple regions through geographic conditioning. Both the season and position encodings are concatenated as extra channels to the input and to the residuals after each SIU-Net convolutional block. As there are four encoding values (two seasonal and two position) there are four extra channels, and each value is duplicated across its respective channel (for example, a row encoding of 0.5 has all channel values set to 0.5).

\subsection{Anomaly Map \& Scoring}

Unlike change detection where the main evaluation metric is pixel-wise accuracy of the change map, SHAZAM's shift towards change monitoring focuses on image-wise event detection. Hazard mapping is a secondary objective. As such, SHAZAM treats hazard detection as a time-series anomaly detection problem, assigning each satellite image an anomaly score which can be used to detect significant changes in the ROI. To do this, SHAZAM uses the structural difference between the real satellite image and SIU-Net's generate image for the same day of the year. The structural difference measure is based on the Structural Similarity Index Measure (SSIM \cite{wang2004ssim}).

The SSIM compares the similarity of two images - in this case, the image generated by SIU-Net $\hat{y}$ and the real satellite image $y$ - using three key components: luminance, contrast, and structure. For each pixel, these components are calculated in a local window (typically 11×11 pixels) that moves across the image in a sliding window fashion. Mathematically, the SSIM score of each pixel is defined as:
\begin{equation}
\text{SSIM}(\hat{y},y) = \frac{(2\mu_{\hat{y}}\mu_y + k_1)(2\sigma_{\hat{y}y} + k_2)}{(\mu_{\hat{y}}^2 + \mu_y^2 + k_1)(\sigma_{\hat{y}}^2 + \sigma_y^2 + k_2)}
\end{equation}
where $\mu$ represents the window mean intensity computed using a Gaussian filter, $\sigma$ represents the window standard deviation, $\sigma_{\hat{y}y}$ is the window covariance between the predicted and real images, and $k_1=0.01$, $k_2=0.03$ are small constants to prevent division by zero. The SSIM is computed independently for each spectral channel, and the final image-wise SSIM score is obtained by averaging across all pixels and channels. The final $\text{SSIM}_{\text{score}}$ ranges from -1 to 1, where 1 indicates perfect structural similarity, 0 is not similar, and -1 is perfect dissimilarity.

As SSIM measures similarity, it is inverted and normalised to create the Structural Difference Index Measure (SDIM). SHAZAM computes two SDIM-based metrics: an image-wise anomaly score to flag if an image contains a potential hazard or not, and a pixel-wise anomaly map to map the hazard. The image-wise anomaly score is calculated by inverting and normalising the global SSIM score into a 0 to 1 range:
\begin{equation}
\text{SDIM}_{\text{score}} = \frac{1 - \text{SSIM}_{\text{score}}}{2}
\end{equation}
where $\text{SSIM}_{\text{score}}$ is the average SSIM across all pixels and channels.

A score of 0 means that the predicted image is identical to the true image. Scores increasing from 0 to 0.5 indicate increased differences between the images and, therefore, a higher indication of an anomaly. Scores between 0.5 and 1 indicate that the images are negatively correlated and dissimilar.

For the anomaly map, SHAZAM uses a few more steps. The anomaly map first computes the SSIM of each pixel across each spectral channel using the aforementioned local-window approach. It then calculates the average of these channel-wise maps to give a single score per pixel, and inverts them. The resulting values are clamped between 0 and 1, before being squared to suppress the background and emphasise the major structural differences:
\begin{align}
\text{SSIM}_{\text{avg}} &= \frac{1}{C} \sum\limits_{c=1}^{C} \text{SSIM}_c (\hat{y},y) \\
\text{SDIM}_{\text{map}} &= \text{max}(0, \text{min}(1, 1 - \text{SSIM}_{\text{avg}}))^2
\end{align}
where $C$ represents the number of channels. Overall, this two-pronged approach provides both a quantitative measure of unexpected change in the ROI, and a distinct map that highlights where these change have occurred.

\subsection{Seasonal Threshold}

Instead of a flat threshold for anomaly detection, a seasonal threshold is used to account for variations in SIU-Net's performance throughout the year. This is justified by the difficulty of reconstructing seasons with significant structural differences from the baseline image. For example, predicting the first snowfall in winter is challenging, as snow accumulation patterns vary annually and obscure landmarks differently each year. This makes it harder for SIU-Net to model the ROI, increasing the average anomaly score during this period. Therefore, a cyclical seasonal threshold is proposed that changes across the year.

To calculate the seasonal threshold, a linear regression is first fitted to the anomaly scores from the training period using circular day-of-year coordinates to ensure continuity across the year boundary:
\begin{align}
    \hat{m}_s (t) = a_1 \cdot t_{\text{sin}} + b_1 \cdot t_{\text{cos}} + c_1
\end{align}
where $\hat{m}_s(t)$ represents the expected anomaly score for day $t$.

A second linear regression is fitted to the monthly standard deviations of the anomaly scores, capturing the seasonal variation in SIU-Net's performance:
\begin{align}
    \hat{s}_s (t) = a_2 \cdot t_{\text{sin}} + b_2 \cdot t_{\text{cos}} + c_2
\end{align}
where $\hat{s}_s(t)$ represents the expected standard deviation for day $t$.

\begin{figure}[h!]
    \centering
    \includegraphics[width=\linewidth]{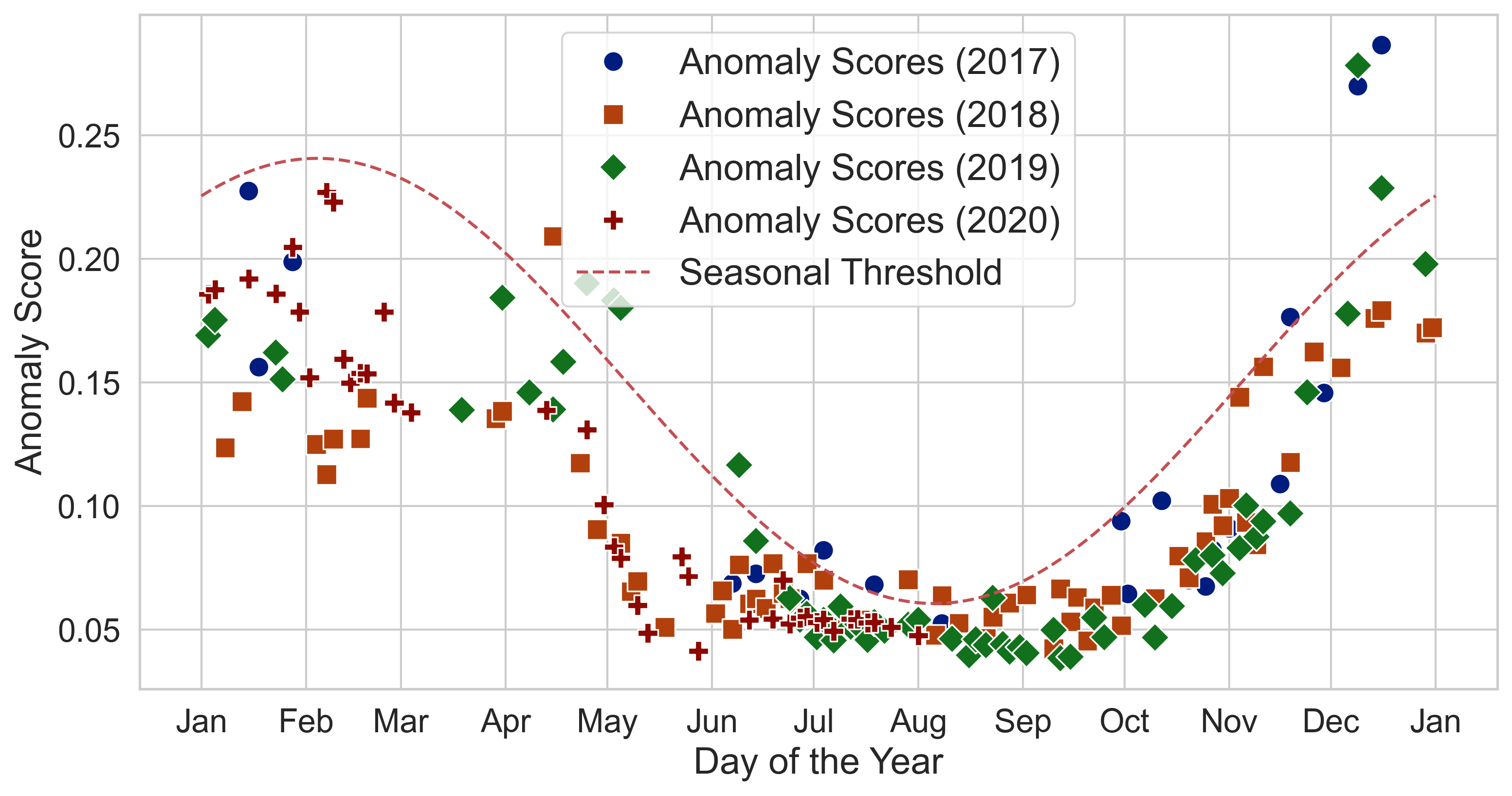}
    \caption[Seasonal Threshold Example]{Example of the seasonal threshold as calculated on training images from the SNP dataset covered in the following section.}
    \label{fig:seasonal_threshold}
\end{figure}

The seasonal threshold curve is then defined as:
\begin{align}
    \tau (t) = \hat{m}_s (t) + 1.64 \times \hat{s}_s (t)
\end{align}
where $\tau(t)$ is the threshold value for day $t$. The multiplier of 1.64 corresponds to the 95th percentile of a normal distribution (one-tailed), allowing the threshold to capture approximately 95\% of the training dataset's normal variations. Scores that exceed this threshold are classified as anomalies. Figure \ref{fig:seasonal_threshold} illustrates this adaptive threshold applied to the SNP dataset.

\section{Experimental Setup} \label{sec:experiments}

\subsection{Datasets}

\begin{figure*}[b]
    \centering
    \includegraphics[width=\linewidth]{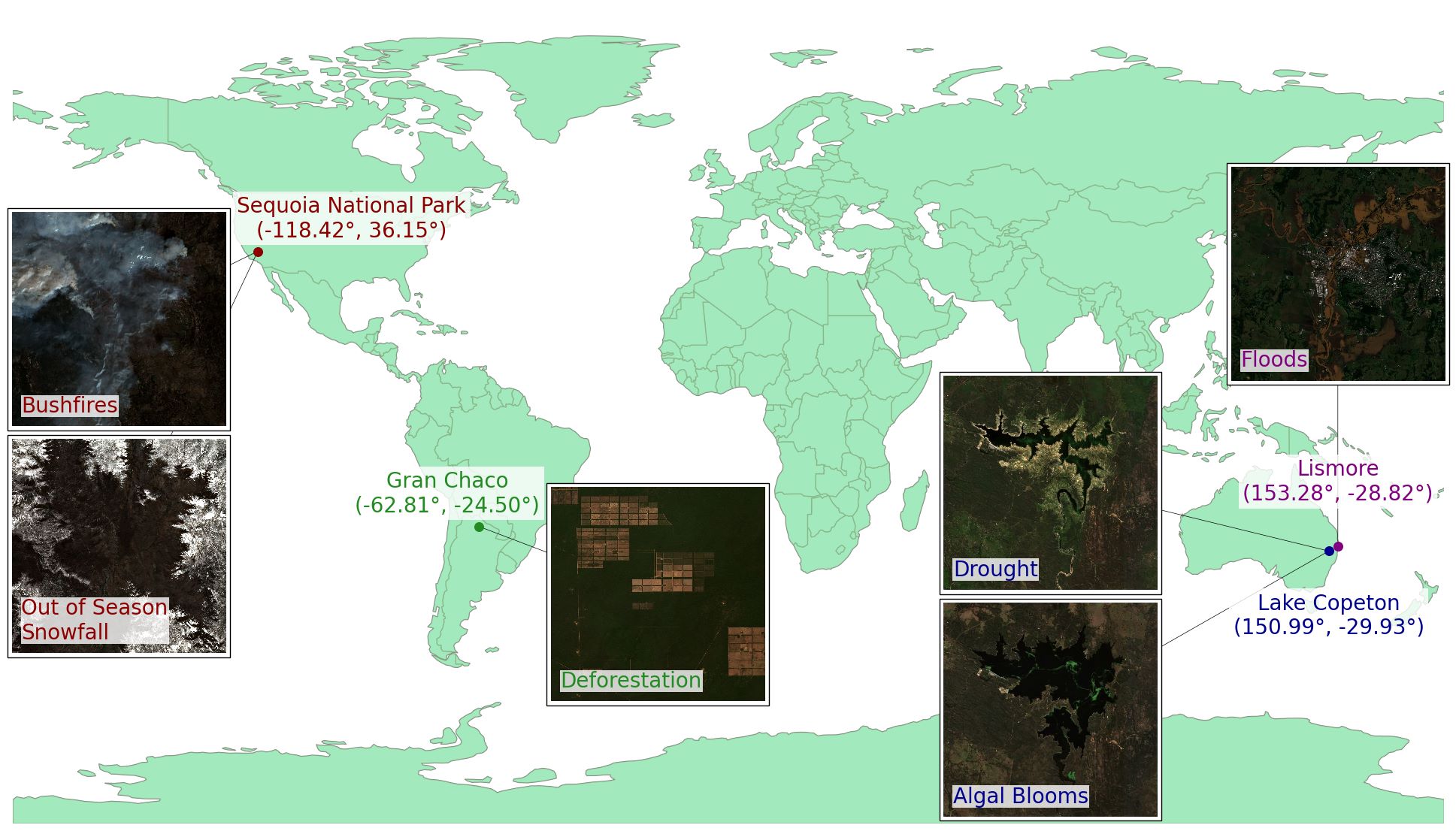}
    \caption[Datasets' Geographical Locations and Examples]{Geographical locations of the four datasets/ROIs. Examples of their respective hazards are shown.}
    \label{fig:dataset_map}
\end{figure*}

\begin{table*}[t]
\centering
\caption{Dataset Attributes for Each Region of Interest (ROI).}
\begin{tabular}{@{}ccccc@{}}
\toprule
Dataset Attributes       & Sequoia National Park (SNP)      & Gran Chaco            & Lake Copeton          & Lismore                \\ \midrule
Spatial Res. (Pixels)    & 1024x1024                   & 1024x1024             & 512x512               & 1024x1024              \\
Pixel Size               & 30m                         & 30m                   & 30m                   & 10m                    \\
Hazards                  & Bushfires, Extreme Snowfall & Deforestation         & Algal Blooms, Drought & Floods                 \\
Training Period          & Jan 2017 to July 2020       & Jan 2018 to June 2020 & Jan 2022 to Apr 2024  & June 2017 to June 2020 \\
No. of Training Images   & 186                         & 56                    & 78                    & 83                     \\
Test Period              & Aug 2020 to Aug 2023        & July 2020 to Apr 2024 & Jan 2020 to Dec 2021  & July 2020 to Aug 2024   \\
No. of Test Images       & 250                         & 110                   & 61                    & 105                    \\
Test Images with Hazards & 187                         & 108                   & 56                    & 24                     \\
Hazard Image Percentage  & 74.8\%                      & 98.2\%                & 91.8\%                & 22.9\%                 \\ \bottomrule
\end{tabular}
\label{tab:dataset_attributes}
\end{table*}

Four datasets were collected through the Sentinel-2 satellite constellation and extracted using Sentinel-Hub\footnote{\url{https://www.sentinel-hub.com/}} and its associated Python API. These datasets cover different geographical regions of interest with various types of hazards. The images all use Level-2a processing, which means that the pixel values have been radiometrically corrected to provide bottom-of-atmosphere surface reflectance. From the available 13 spectral bands, only the 10 bands with 10m or 20m resolution are used, excluding the three 60m resolution bands which are primarily used for atmospheric correction. The SNP, Gran Chaco, and Lake Copeton datasets are further downsampled to 30m pixel resolution, as this is sufficient for detecting their respective hazards while allowing coverage of larger regions of interest. The geographical locations of all datasets and hazard examples are shown in Figure \ref{fig:dataset_map}. Further details are provided in the following sections, with their key attributes shown in Table \ref{tab:dataset_attributes}.

\subsubsection{Sequoia National Park (SNP)}

This dataset was collected over a forested mountain region in the Sequoia National Park in California, the United States of America. This is a unique area that contains both wildfires and extreme out-of-season snowfall throughout the year. Detecting out-of-season snowfall is useful for nearby communities, which are at risk of flooding when the snow melts. Images captured when snow depth was above or below the maximum and minimum of the training period were labelled anomalies, based on data obtained via the National Water and Climate Center (NWCC) for the Leavitt Meadows ground station near to the ROI \cite{NWCC2024SNPSnowfall}. The wildfires were identified by visual inspection of RGB and infrared images. Images with an average Normalised Burn Ratio (NBR) below the training minimum were also labelled as anomalies, as they indicate large amounts of destroyed vegetation in the ROI (i.e. burned regions).

\subsubsection{Gran Chaco}

Gran Chaco is the second largest forest in South America, stretching into Argentina, Bolivia, Paraguay, and Brazil. It is a hot and dry forest that is sparsely populated and is under constant threat of deforestation to create agricultural land \cite{gasparri2009granchaco_deforestation, fehlenberg2017granchaco_deforestation}. Remote detection of deforestation allows local authorities to intervene and reduce further forest degradation. Deforestation is the primary anomaly present, with some minor fires used for crop burning. The test images were labelled as anomalies by visual inspection in combination with outliers identified using the Normalised Difference Vegetation Index (NDVI). This dataset has a consistent increase in deforestation over time and is therefore highly unbalanced, as almost all test images are considered anomalous.

\subsubsection{Lake Copeton}

Lake Copeton is a lake bounded by Copeton dam in regional New South Wales, Australia. Lake Copeton provides freshwater to nearby communities for agricultural, household, and recreational purposes and is also home to aquatic life \cite{WaterNSW2024AboutCopeton}. The use of upstream fertilisers has led to algal blooms and reduced water quality, especially during heavy rainfall. These water quality drops pose health risks to the ecosystem and community if not properly managed. In addition, the region is prone to drought and significant variations in the amount of water stored in the lake. 

Images with algal blooms and reduced water quality were partially labelled with periods corresponding to water quality alerts from the government organisation that manages Copeton Dam, WaterNSW \cite{WaterNSW2022CopetonWQ}. Visual inspection of RGB and Normalised Difference Chlorophyll Index (NDCI) \cite{mishra2012NDCI} was also used to label images with visible water disturbances, such as large sediment clouds. Images were labelled as drought/low water level when Copeton Dam had less than 50\% water capacity \cite{WaterNSW2024CopetonStorageLevels}.

\subsubsection{Lismore}

Lismore, a town in regional New South Wales, Australia, is prone to flooding, having experienced more than ten flood events in the last two decades \cite{lismore2022FloodHistory}. Although satellite images are not necessarily the most effective method for flood detection in populated areas, they provide valuable spatial information for mapping flood impacts. The 2022 floods altered the landscape by causing extensive damage to infrastructure and agricultural land, with an estimated cost of approximately \$1 billion AUD \cite{lismore2022FloodResponseReport}. While active flooding is captured in only a few images due to its brief duration, post-flood effects are visible through damaged structures and altered land surfaces. These floods are visually confirmed and documented in Lismore Council's records. 

The images are labelled as hazardous from the February/March 2022 floods until January 2023, as visual analysis of the satellite imagery showed larger flood-induced landscape alterations diminishing around this time. The dataset serves as an effective test case for SHAZAM's ability to detect both immediate flood impacts and persistent landscape changes. The area also undergoes gradual changes in land cover due to urban and agricultural development, adding complexity to the dataset as these changes may be detected as potential anomalies.

\subsection{Implementation \& Comparative Models}

\subsubsection{SHAZAM}

SHAZAM, using the SIU-Net architecture, was trained for 20 epochs with a batch size of 32. For all datasets, the training images were cut into 32x32 patches, with 90\% used to train SIU-Net for patch translation, and 10\% randomly selected to validate the model's translation performance. The L1 loss, also known as the Mean Absolute Error (MAE), was used as the loss function. An initial learning rate of $1e-4$ with the Adam optimiser was used (with default parameters $\beta_1=0.9$, $\beta_2=0.999$). The learning rate was progressively reduced by a factor of 0.1, whenever the validation loss plateaued for 3 epochs. The minimum learning rate was set to $1e-7$. Once trained patch-wise, SHAZAM was implemented image-wise to the training period so that the seasonal threshold could be estimated using the training image anomaly scores. Finally, SHAZAM was applied to the test period for final evaluation (as shown in the inference stage of Figure \ref{fig:shazam}). All models, including SHAZAM, were implemented in Python (with PyTorch Lightning\footnote{\url{https://github.com/Lightning-AI/pytorch-lightning}} used for deep learning). The layer weights are initialised with the PyTorch defaults.

\subsubsection{cVAE}

A conditional variational autoencoder (cVAE) serves as a baseline model, chosen for its ability to learn complex data distributions based on conditional inputs \cite{kingma2013OGVAE, sohn2015OGcVAE}. The cVAE employs a symmetric encoder-decoder structure with residual connections, where the encoder progressively downsamples the input through alternating convolutional and residual blocks while increasing channel depth (32, 64, 128). The output is then mapped to a normalised 256-dimensional latent space using the standard VAE reparameterisation trick.

The latent space is conditioned on four variables: the cyclical day-of-year encodings (sin, cos) and the scaled patch positions (row, column). These variables directly enable the model to capture seasonal patterns and regional variations across the ROI. The decoder mirrors the encoder's architecture, but uses transpose convolutions to upsample the latent features to the original patch size. During inference, only the decoder is used to generate patches given temporal-spatial conditions. Training is guided by an L1 reconstruction loss of the input and the Kullback-Leibler divergence ($\beta = 0.1$) to balance generation quality with a Gaussian latent space. For training, a batch size of 32 was used alongside 20 epochs. cVAE uses the same data pre-processing (including normalisation) and optimisation strategy as SHAZAM. The training phase begins with an initial learning rate of $1e-4$, decreasing by 0.1 when the validation loss plateaus for 3 epochs. For inference, the average L1 loss per image is used for ROI hazard detection, and the L1 loss per pixel is used for mapping. The hazard detection threshold is the mean L1 loss of the training images plus 1.64 standard deviations, following the theoretical justification as SHAZAM's threshold.

\subsubsection{RaVAEn}

RaVAEn provides the closest fit to the objective, given its direct design for change monitoring to detect natural hazards/disasters in Sentinel-2 time-series patches \cite{ruuvzivcka2022ravaen}. RaVAEn uses a VAE to compress patches into a 128-dimensional latent space, and then uses the minimum cosine distance between the current patch's latent features and historical patch latent features to score events. The three latest patches are used to detect change events.

Given that RaVAEn was designed for 32x32 pixel patches and the ROI images are much larger than this, two variants are implemented. The first, RaVAEn-Local, calculates the minimum cosine distance for each patch as in the paper (comparing the current patch to the three historical patches). The minimum cosine distance is then used to represent all pixels in the 32x32 patch, which is repeated for all patches in the ROI to create the anomaly heatmap. The average score across the anomaly heatmap is used to flag a hazard in the region. The second variant, RaVAEn-Global, computes and stores the cosine distances for the current patch and each of the three preceding patches. These values are used to create three anomaly heatmaps for each preceding image of the ROI. The heatmap with the lowest average score is then used to detect and map anomalies. The hazard detection threshold is the mean L1 loss of the training images plus 1.64 standard deviations, following the theoretical justification as SHAZAM's threshold. Cubic convolution is used to upsample the heatmaps to match the image resolution for visual inspection. For training, a batch size of 32 was used alongside 10 epochs, with a fixed learning rate of $1e-3$ for the Adam optimiser.

The s2cloudless cloud filtering algorithm was not used due to the low cloud coverage in the datasets, and because the algorithm risks filtering out snow-based pixels. This is particularly important given that the extreme and out-of-season snowfall is a key hazard in the SNP dataset. Nonetheless, the handling of cloud coverage remains a crucial requirement for optical SITS monitoring systems and is discussed in further detail in Section \ref{sec:future_work}.

\subsubsection{COLD}

The continuous monitoring of land disturbance algorithm (COLD) provides a good point of comparison with disturbance detection methods, as it is designed to detect various types of disturbance using multiple bands \cite{zhu2020COLD}. Designed for Landsat SITS, COLD models each band's time series independently for each pixel in the image. It captures trends and seasonality using a limited Fourier series. The author's implementation in the PYCOLD library is used\footnote{\url{https://github.com/GERSL/pycold}}.

Several adaptations were made to implement COLD with Sentinel-2 data for immediate hazard detection and mapping. First, only bands that closely match Landsat wavelengths were used (blue, green, red, near-infrared, short-wave infrared 1, and short-wave infrared 2, with no thermal band being used). Second, to balance computational efficiency with accuracy, the analysis was performed on the mean values of 8x8 pixel patches rather than individual pixels. Disturbance detection methods usually need HPCs to process an ROI promptly, which are unlikely to be available for constant monitoring. Third, instead of using COLD's approach of multiple observations to confirm a disturbance, the mean disturbance probability is used as an anomaly score. This allows COLD to be tested for detecting and mapping hazards on the first observation. Finally, like RaVAEn, COLD's cloud filtering is omitted because of the low cloud coverage and the need to detect snow-based hazards.

When evaluating each test image, COLD uses all available historical images from the datasets to model the time series of each 8x8 patch. This includes both the complete training dataset and any previous test images. However, COLD could not be evaluated on the Lake Copeton dataset, which lacked sufficient data prior to the test period. Since all SITS training data are required for modelling, it is not feasible to calculate a threshold similar to the approaches used by SHAZAM, cVAE, and the RaVAEn variants. Therefore, COLD is compared using only threshold-less metrics (discussed in the following section). Like RaVAEN, cubic convolutions are used to upsample the heatmaps.

\subsection{Metrics}

The precision, recall and F1 scores are used to evaluate performance of each model when using a specific threshold $\tau$. Precision is the ratio of true positives to all identified hazards, which assesses the accuracy of the model for hazard detection and the risk of false alarms. Minimising false alarms (and maximising precision) is crucial to avoid wasting resources on unnecessary hazard confirmations or responses. Recall measures correctly identified hazards against the total number of true hazardous images, considering false negatives (or missed hazards). Recall is crucial for hazard detection, as missing hazards can have severe consequences. The F1 score combines both precision and recall into a single aggregated metric:
\begin{align}
    \text{Precision} &= \frac{\text{True Positives}}{\text{True Positives} + \text{False Positives}} \\
    \text{Recall} &= \frac{\text{True Positives}}{\text{True Positives} + \text{False Negatives}} \\
    \text{F1 Score} &= \frac{2 \cdot \text{Precision} \cdot \text{Recall}}{\text{Precision} + \text{Recall}}
\end{align}
The F1 score can be considered a more accurate measure than accuracy, as it accounts for the imbalance between anomalous and normal images in the test dataset. In addition, the Area Under the Precision-Recall Curve (AUPRC) is used to evaluate the models across a range of possible thresholds. This is a robust secondary metric that assesses general detection performance without a specific threshold value, and is also unaffected by the imbalance between change events. Precision, recall, F1 score, and AUPRC range from 0 to 1, with higher values denoting better performance.

An important note is that SHAZAM's AUPRC is calculated using the residuals after removing the seasonal threshold. Since the seasonal threshold models SHAZAM's error rates throughout the year (mean fit plus 1.64 times the monthly standard deviation fit), removing it leaves the residual scores that determine if an image is anomalous (greater than 0) or normal (less than zero). This allows the AUPRC to evaluate performance across a range of constant thresholds, providing a fairer assessment in-line with the comparative models while allowing SHAZAM to benefit from its complete seasonal modelling.

As many of the datasets are heavily imbalanced, a ``coin-toss" random model is used as a baseline against which learning-based models perform the task. This random model simply guesses with a 50\% chance if any given image contains a hazard or not. The metrics for this random model can be theoretically calculated, yielding its precision, recall, and F1 score as follows:
\begin{align}
    \text{Precision}_{\text{rand}} &= \frac{p \cdot P}{p \cdot P + p \cdot (1 - P)} = P \\
    \text{Recall}_{\text{rand}} &= \frac{p \cdot P}{p \cdot P + (1-p) \cdot P} = 0.5 \\
    \text{F1 Score}_{\text{rand}} &= \frac{2 \cdot P \cdot p}{P + p} = \frac{P}{P + 0.5}
\end{align}
where $P \in [0,1]$ is the fraction of positive samples in the dataset (i.e. the fraction of images labelled as hazards), and $p = 0.5$ is the probability of the random model classifying an image as having a hazard. Given that the precision of the random model is constant at $P$ regardless of the classification threshold, and thus remains fixed for all values of recall, the AUPRC is simply the same as the precision: 
\begin{align}
    \text{AUPRC}_{\text{rand}} &= P
\end{align}

\subsection{Experiments \& Hardware}

Change monitoring differs from change detection by aiming to flag events at the image level, rather than mapping individual changed pixels. Given this focus on event detection for an ROI, the first experiment evaluates SHAZAM and the comparative models in hazard detection at the image level across all datasets. The detection performance of each model is evaluated using the quantitative metrics covered in the previous section, and a qualitative analysis of the anomaly heatmaps is used to evaluate the mapping ability. Particular attention is given to heatmaps of edge cases (missed hazards and false alarms) to identify SHAZAM's limitations and opportunities for improvement.

An ablation study evaluates SHAZAM's core components by systematically removing positional encodings, cyclical seasonal encodings, the structural difference module, and the seasonal threshold. Seasonal encodings are removed by replacing SIU-Net and the seasonal threshold's sine and cosine day-of-year embeddings with the normalised day-of-year (0 for January 1st to 1 for December 31st). To ablate the structural difference, it is replaced by the mean absolute error (MAE). This analysis helps validate design choices and explores potential simplifications of the architecture.

All experiments are conducted onboard a Windows desktop with an AMD Ryzen 9 3900X 12-Core Processor, 64GB of RAM, and an NVIDIA GeForce RTX 3080 graphics card with 10GB of VRAM. This is with the exception of COLD, which was implemented in a Google Colab environment (Linux), with access to 51GB of RAM and an 8-core CPU.

\section{Results \& Discussion} \label{sec:results}

\subsection{Hazard Detection Comparison}

The detection results of SHAZAM and the comparative models are shown in Table \ref{tab:detection_results}. Since the Lake Copeton dataset's test period predates the training period, COLD was not used because it needs sequential historical data for change detection.

\begin{table*}[t]
\centering
\caption{Detection Results}
\begin{tabular}{@{}c|ccccc|ccccc@{}}
\toprule
\multicolumn{1}{l|}{} & \multicolumn{5}{c|}{Sequoia National Park (SNP)}                          & \multicolumn{5}{c}{Gran Chaco}                                            \\
Model                & \#Params      & F1           & Precision    & Recall       & AUPRC        & \#Params      & F1           & Precision    & Recall       & AUPRC        \\ \midrule
Random Guess         & -             & $0.599$      & $0.748$      & $0.5$        & $0.748$      & -             & $0.662$      & $0.982$      & $0.5$        & $0.982$      \\
cVAE                 & 2.6M          & $0.245$      & $0.470$      & $0.169$      & $0.627$      & 2.6M          & $0.230$      & $\bm{1.000}$ & $0.130$      & $\bm{0.999}$ \\
COLD                 & 41.9M         & -            & -            & -            & $0.719$      & 41.9M         & -            & -            & -            & $0.995$      \\
RaVAEn-Local         & 617K          & $0.182$      & $\bm{0.864}$ & $0.102$      & $\bm{0.838}$ & 617K          & $0.108$      & $\bm{1.000}$ & $0.057$      & $0.967$      \\
RaVAEn-Global        & 617K          & $0.181$      & $0.826$      & $0.102$      & $0.821$      & 617K          & $0.108$      & $\bm{1.000}$ & $0.057$      & $0.967$      \\
SHAZAM               & \textbf{473K} & $\bm{0.771}$ & $0.767$      & $\bm{0.775}$ & $0.762$      & \textbf{473K} & $\bm{0.728}$ & $0.969$      & $\bm{0.583}$ & $0.987$      \\ \bottomrule
\end{tabular}

\vspace{0.25cm}

\begin{tabular}{@{}c|ccccc|ccccc@{}}
\toprule
\multicolumn{1}{l|}{} & \multicolumn{5}{c|}{Lake Copeton}                                         & \multicolumn{5}{c}{Lismore}                                               \\
Model                & \#Params      & F1           & Precision    & Recall       & AUPRC        & \#Params      & F1           & Precision    & Recall       & AUPRC        \\ \midrule
Random Guess         & -             & $0.647$      & $0.918$      & $0.5$        & $0.918$      & -             & $0.314$      & $0.229$      & $0.5$        & $0.229$      \\
cVAE                 & 2.6M          & $0.324$      & $0.917$      & $0.196$      & $0.918$      & 2.6M          & $0.069$      & $0.200$      & $0.042$      & $0.204$      \\
COLD                 & -             & -            & -            & -            & -            & 41.9M         & -            & -            & -            & $\bm{0.391}$ \\
RaVAEn-Local         & 617K          & $0.486$      & $\bm{1.000}$ & $0.321$      & $\bm{0.949}$ & 617K          & $0.061$      & $0.111$      & $0.042$      & $0.226$      \\
RaVAEn-Global        & 617K          & $0.172$      & $\bm{1.000}$ & $0.094$      & $0.935$      & 617K          & $0.054$      & $0.077$      & $0.042$      & $0.246$      \\
SHAZAM               & \textbf{473K} & $\bm{0.881}$ & $0.906$      & $\bm{0.857}$ & $0.928$      & \textbf{473K} & $\bm{0.466}$ & $\bm{0.304}$ & $\bm{1.000}$ & $0.389$      \\ \bottomrule
\end{tabular}
\label{tab:detection_results}
\end{table*}

Table \ref{tab:detection_results} demonstrates SHAZAM's superior performance on the SNP dataset, achieving both the highest F1 score and recall. Although its precision is comparable to the random baseline, SHAZAM's substantially higher recall than all the other models indicates an enhanced ability to detect hazards events with minimal misses. In contrast, the RaVAEn variants achieve the highest precision and AUPRC, showing excellent performance when minimising false alarms, but their very low recall indicates they miss most hazardous images. The disconnect between RaVAEn's strong AUPRC and poor recall and F1 score suggests the potential for improvement through threshold optimisation. However, SHAZAM's overall stronger and balanced metrics demonstrate that the use of its seasonal threshold provides an effective empirical trade-off between false alarms and missed hazards.

The results of the Gran Chaco dataset follow a similar pattern, with SHAZAM demonstrating superior performance through a higher F1 score and recall. Although SHAZAM has marginally lower precision compared to other models, the precision is notably high across all approaches. This aligns with the region's characteristics of persistent change from gradual deforestation, making false alarms almost impossible. The extremely low recall scores for most models indicate that they miss many hazards due to the default threshold (mean + 1.64 standard deviations) being overly conservative. While the high AUPRC scores suggest strong hazard separation capabilities across all models, this is less meaningful given that setting a low threshold in this region naturally yields high performance due to the high hazard percentage.

SHAZAM demonstrates superior performance across all metrics for the Lake Copeton dataset, except precision. Its F1 score exceeds the random baseline by 0.234 and exceeds the next best model, RaVAEn-Local, by 0.395, marking a substantial improvement. There is a notable discrepancy between the recall and the F1 scores of the two RaVAEn variants. Despite similar AUPRCs and perfect precision values, investigation revealed that RaVAEn-Global's threshold was significantly higher due to elevated anomaly score means and standard deviations in the training set. Further analysis showed this was caused by the Lake Copeton dataset's more prominent cloud coverage within images, and also extended gaps between images due to a more irregular sequence. RaVAEn-Global's higher sensitivity to these factors results in more pronounced seasonal changes, leading to inflated anomaly scores during threshold calculation. This finding emphasises the importance of managing cloud coverage, seasonality, and the irregularity of optical SITS.

SHAZAM demonstrates superior performance across all metrics for the Lismore dataset, with COLD being the only model to achieve comparable AUPRC. SHAZAM's perfect recall indicates that it correctly detects all instances of hazardous images, though its lower precision suggests a tendency toward false alarms. While both Lismore and Gran Chaco experience persistent changes, Lismore's changes are predominantly non-hazardous, stemming from routine human activities such as constructing more buildings and agricultural modifications. This makes false alarms more likely as these regular land cover changes are flagged despite not being hazardous. For regions experiencing such persistent non-hazardous changes, updating the baseline image or using another method to capture trend-based changes could be a viable solution. The comparative models perform significantly worse than the random baseline in terms of F1 score, precision, and recall. While threshold optimisation could marginally improve their results, SHAZAM clearly demonstrates more robust performance.

Across all datasets, SHAZAM consistently demonstrates superior detection performance compared to the cVAE and RaVAEn variants, with F1 score improvements ranging from 0.066 on the Gran Chaco dataset, to 0.234 on the Lake Copeton dataset. While SHAZAM achieves comparable or superior precision scores, its stronger performance is primarily driven by higher recall scores, indicating greater effectiveness in detecting hazards while reducing missed detections. COLD could only be evaluated using AUPRC due to dataset limitations preventing threshold calculation, but within the AUPRC scores, SHAZAM typically matched or slightly trailed other models (except Lake Copeton where it leads). This consistent performance, coupled with the fact that cVAE and RaVAEn variants perform below the random baseline, highlights both the efficacy of the seasonal threshold and the broader importance of threshold selection in change monitoring. The fundamental limitations of these competing approaches are explored in detail through visual comparison in the following section. SHAZAM's lightweight architecture makes it resource and cost-efficient, an essential requirement for practical monitoring systems. Overall, these results demonstrate that SHAZAM successfully balances the critical trade-off between false alarms and missed hazards while maintaining robust detection performance across diverse geographical regions and hazard types.

\subsection{Visual Mapping Evaluation}

\begin{figure*}[p]
    \centering
    
    % Row with 6 images
    \begin{subfigure}[b]{\textwidth}
        \begin{subfigure}[b]{0.16\textwidth}
            \includegraphics[width=\textwidth]{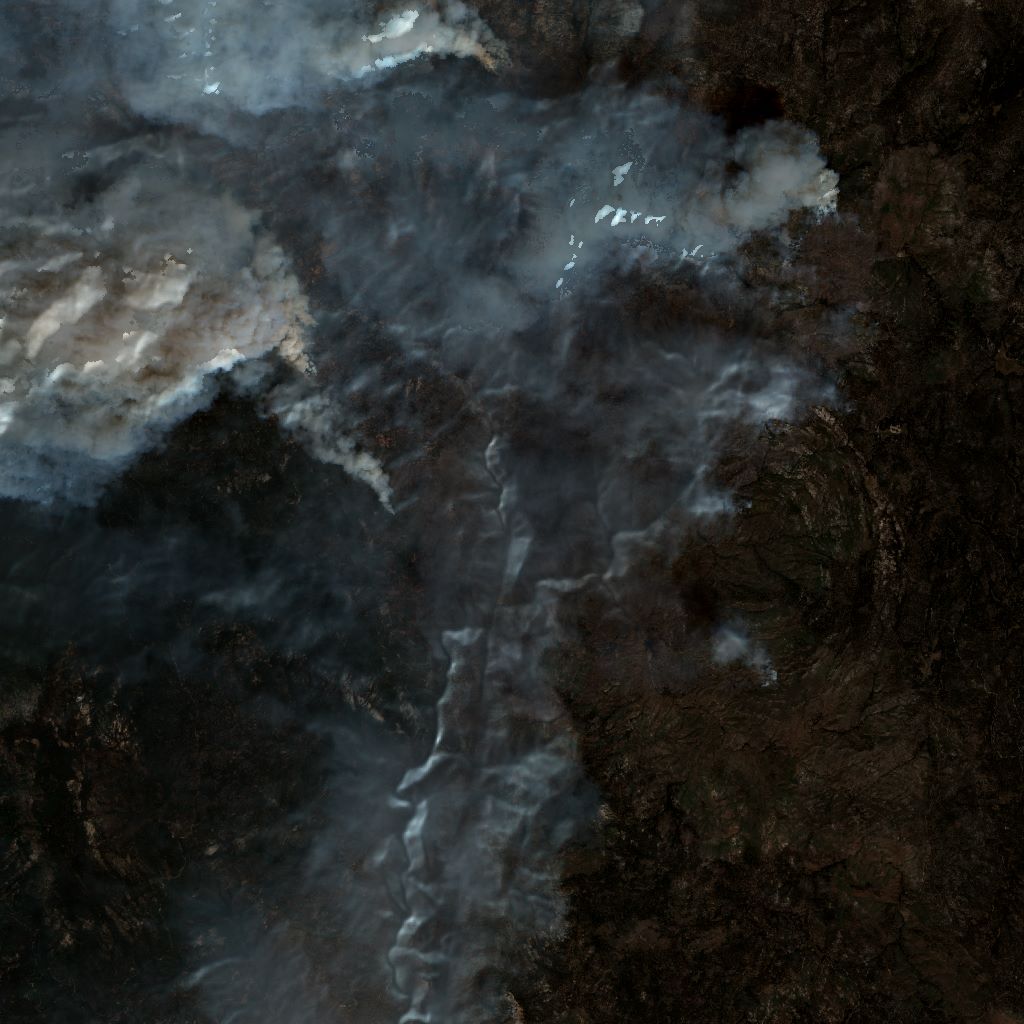}
        \end{subfigure}
        \hfill
        \begin{subfigure}[b]{0.16\textwidth}
            \includegraphics[width=\textwidth]{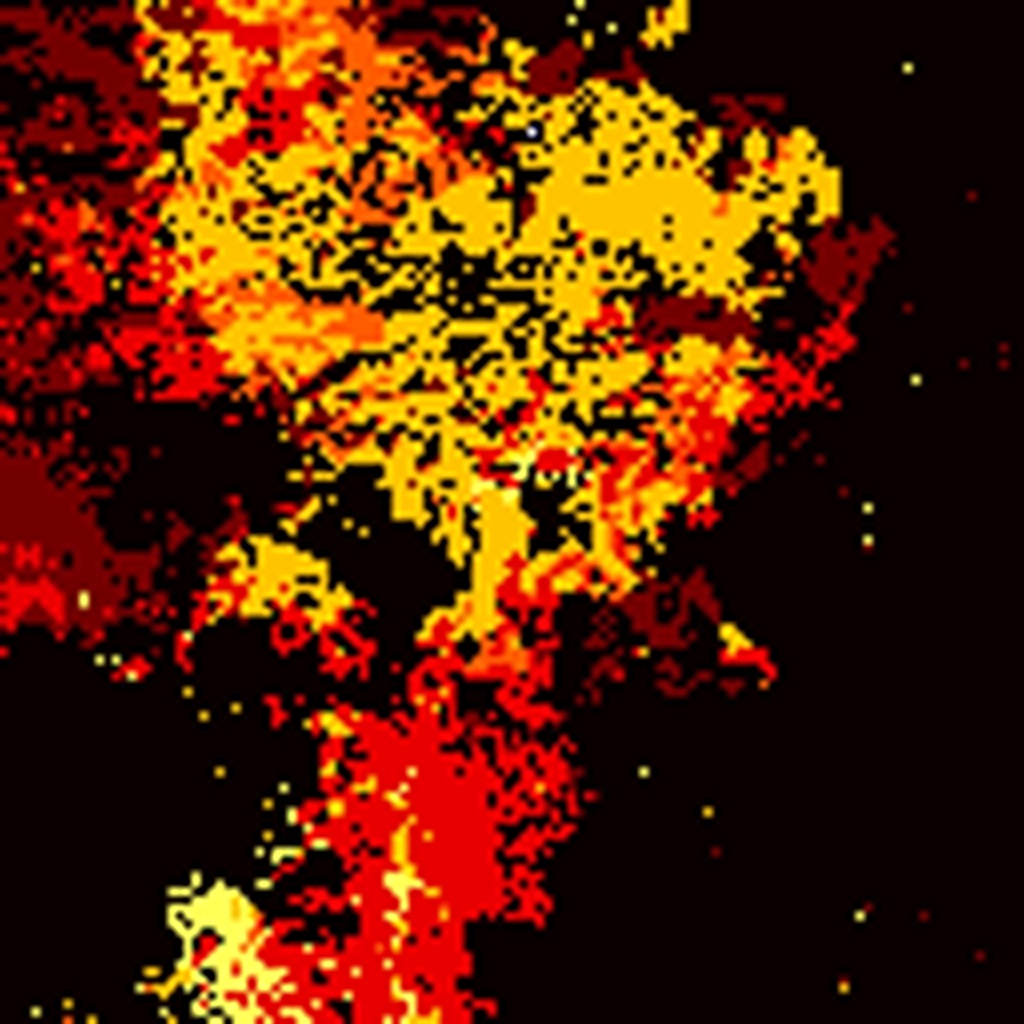}
        \end{subfigure}
        \hfill
        \begin{subfigure}[b]{0.16\textwidth}
            \includegraphics[width=\textwidth]{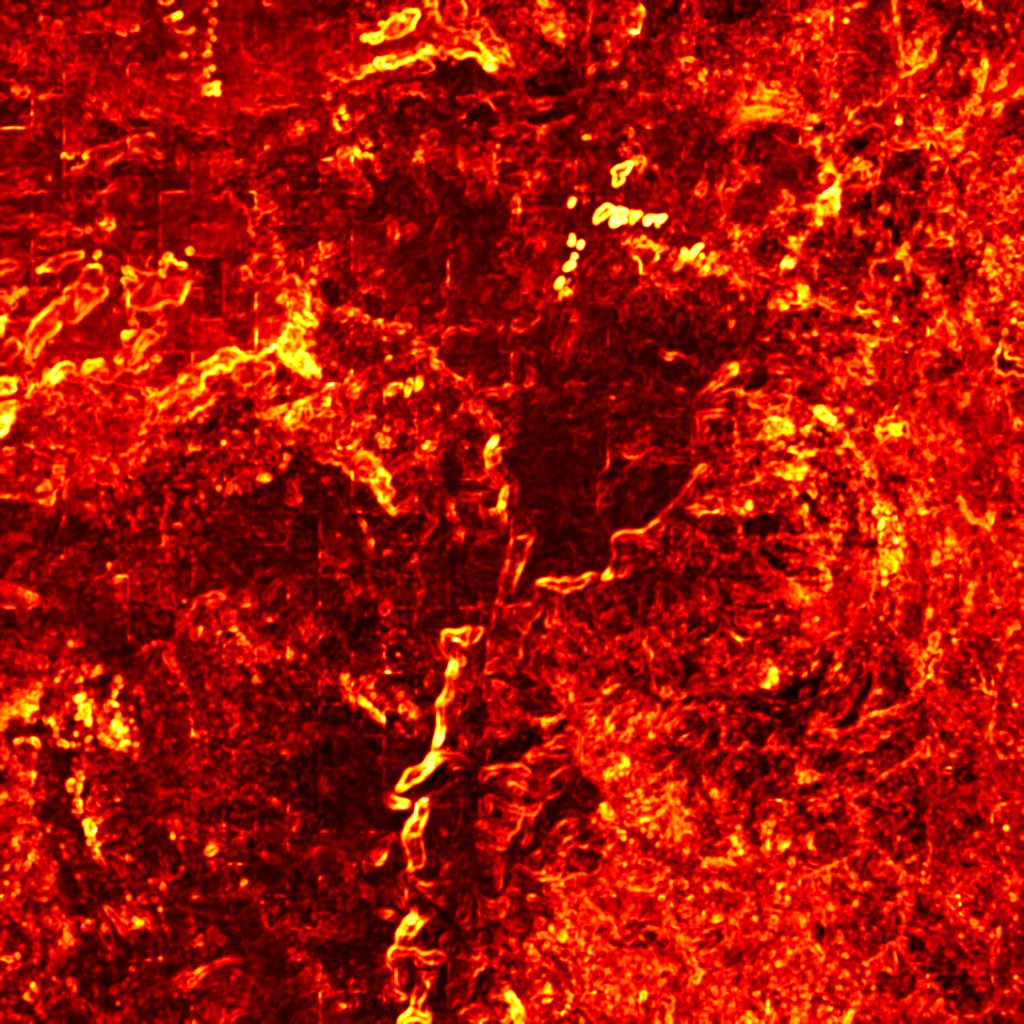}
        \end{subfigure}
        \hfill
        \begin{subfigure}[b]{0.16\textwidth}
            \includegraphics[width=\textwidth]{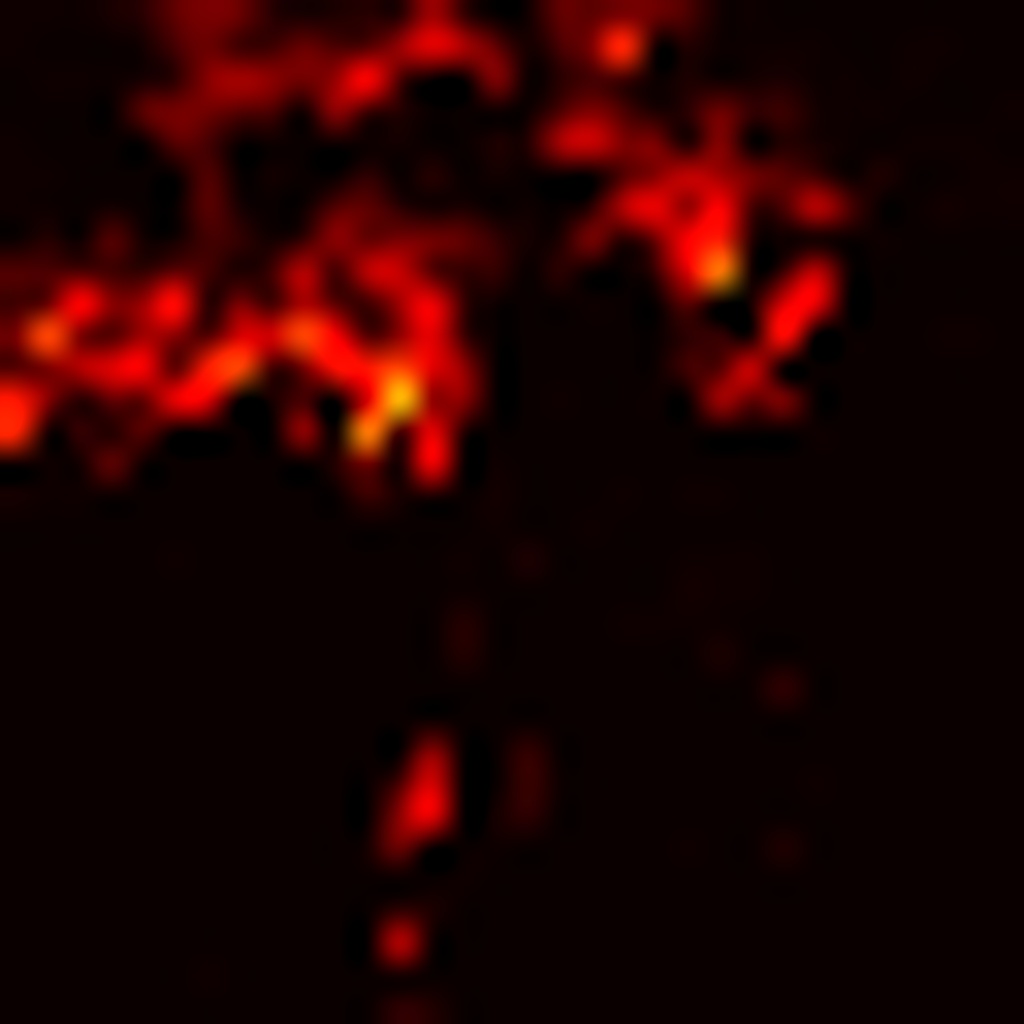}
        \end{subfigure}
        \hfill
        \begin{subfigure}[b]{0.16\textwidth}
            \includegraphics[width=\textwidth]{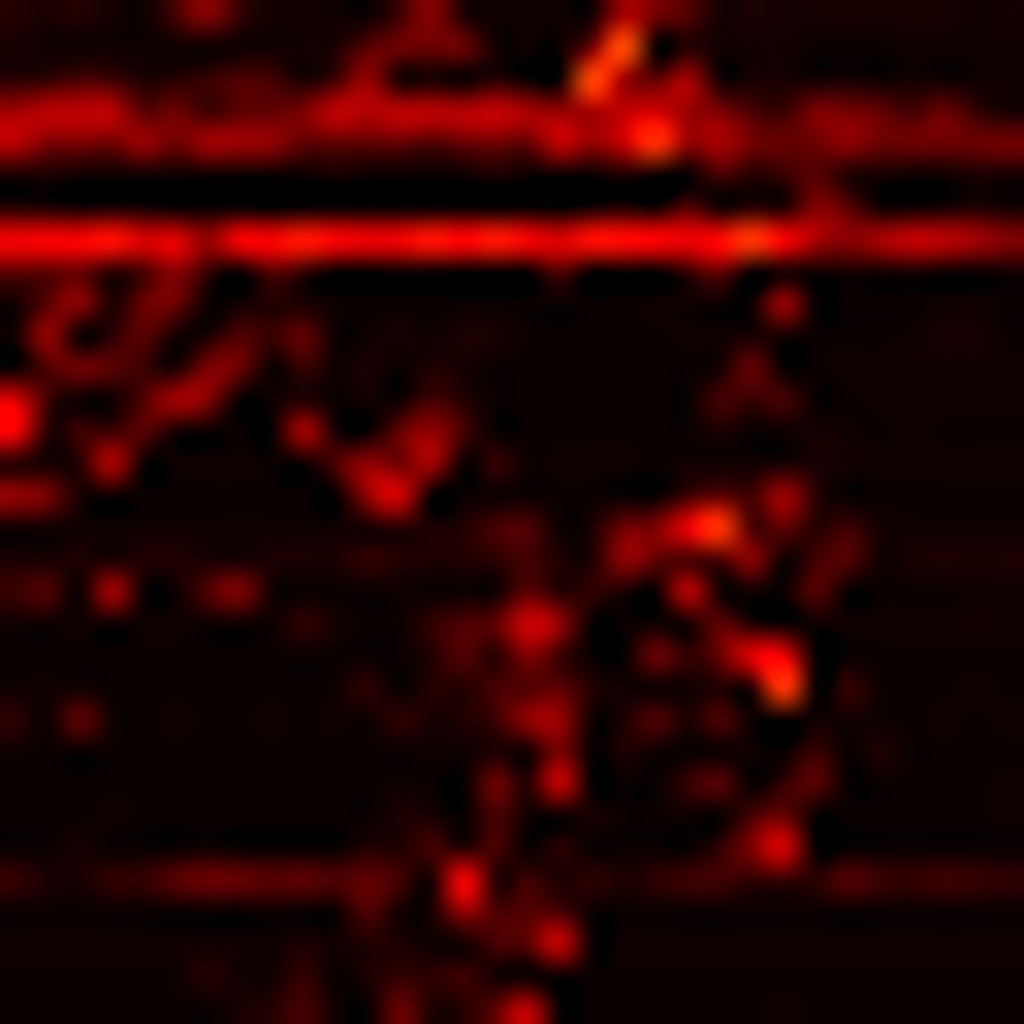}
        \end{subfigure}
        \hfill
        \begin{subfigure}[b]{0.16\textwidth}
            \includegraphics[width=\textwidth]{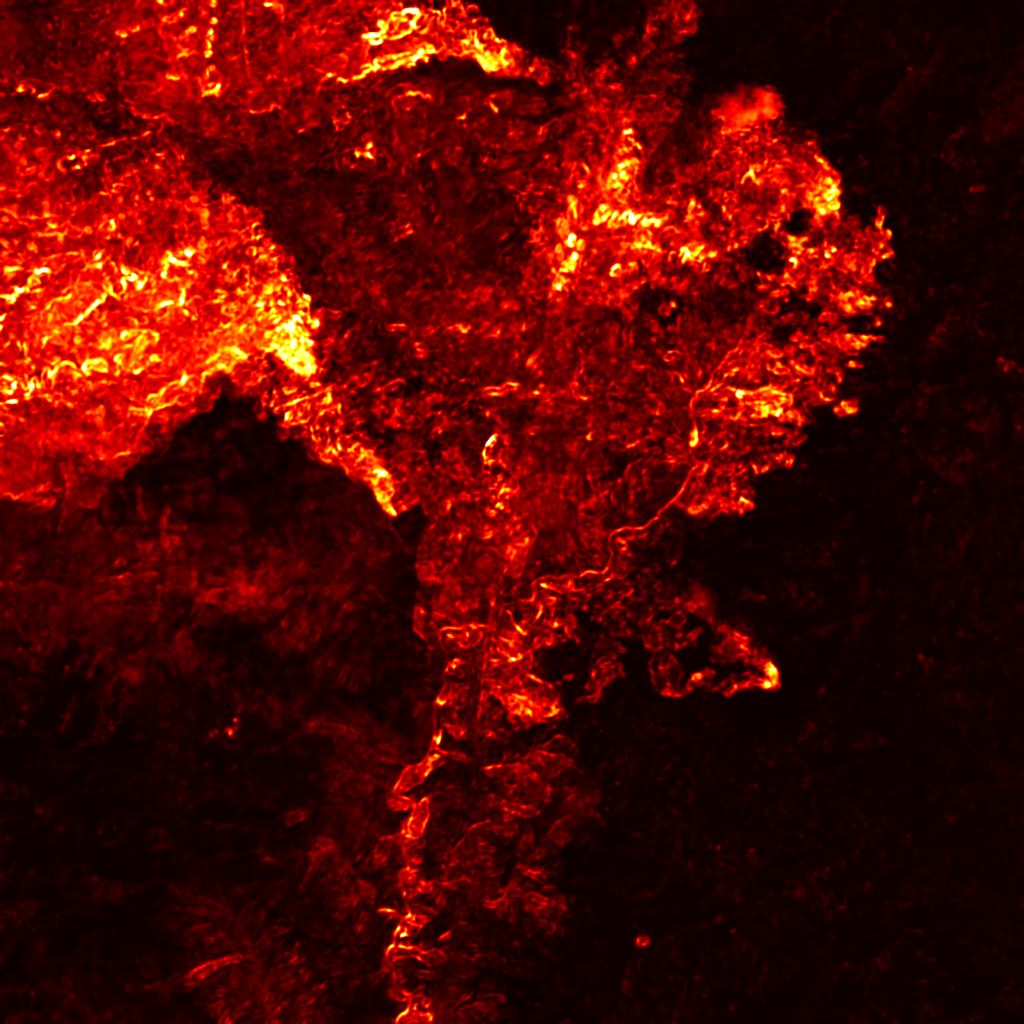}
        \end{subfigure}
    \end{subfigure}
    
    \vspace{0.4em}
    
    % Row with 6 images
    \begin{subfigure}[b]{\textwidth}
        \begin{subfigure}[b]{0.16\textwidth}
            \includegraphics[width=\textwidth]{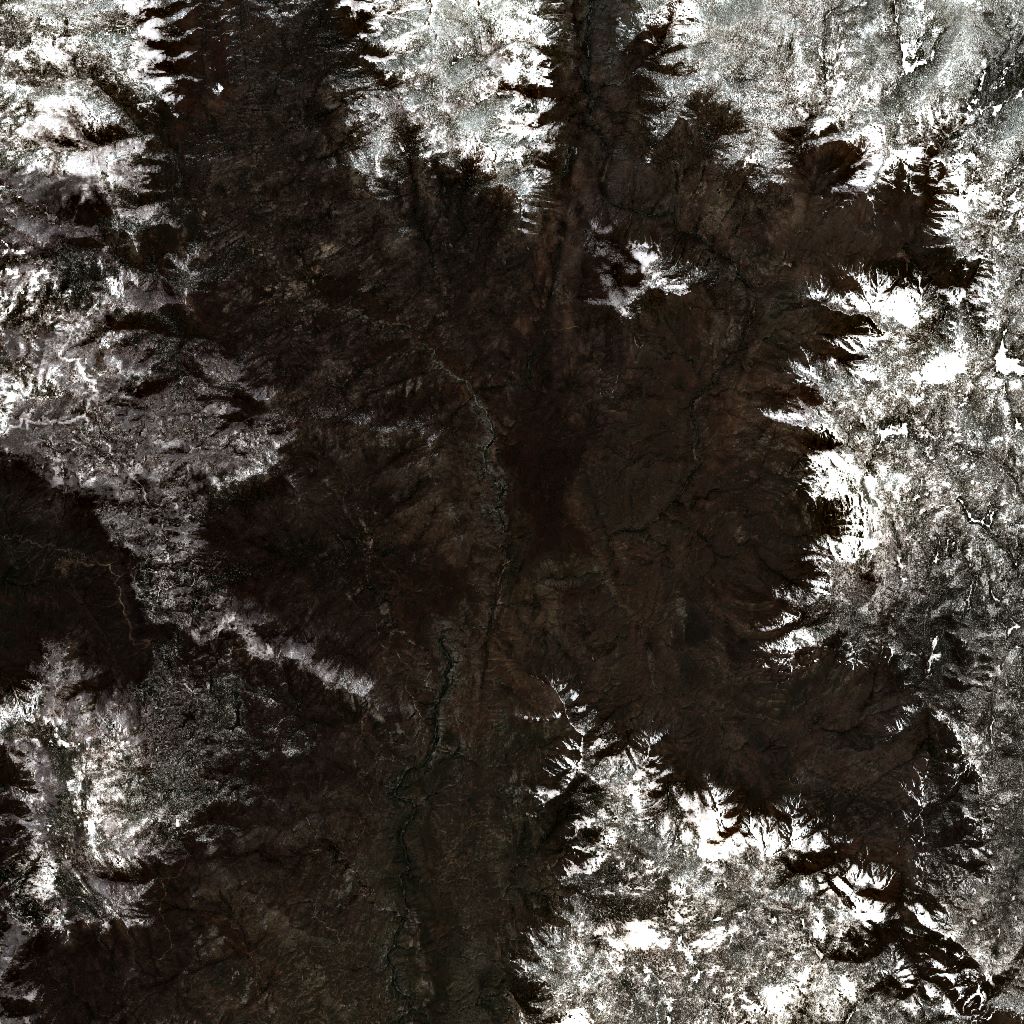}
        \end{subfigure}
        \hfill
        \begin{subfigure}[b]{0.16\textwidth}
            \includegraphics[width=\textwidth]{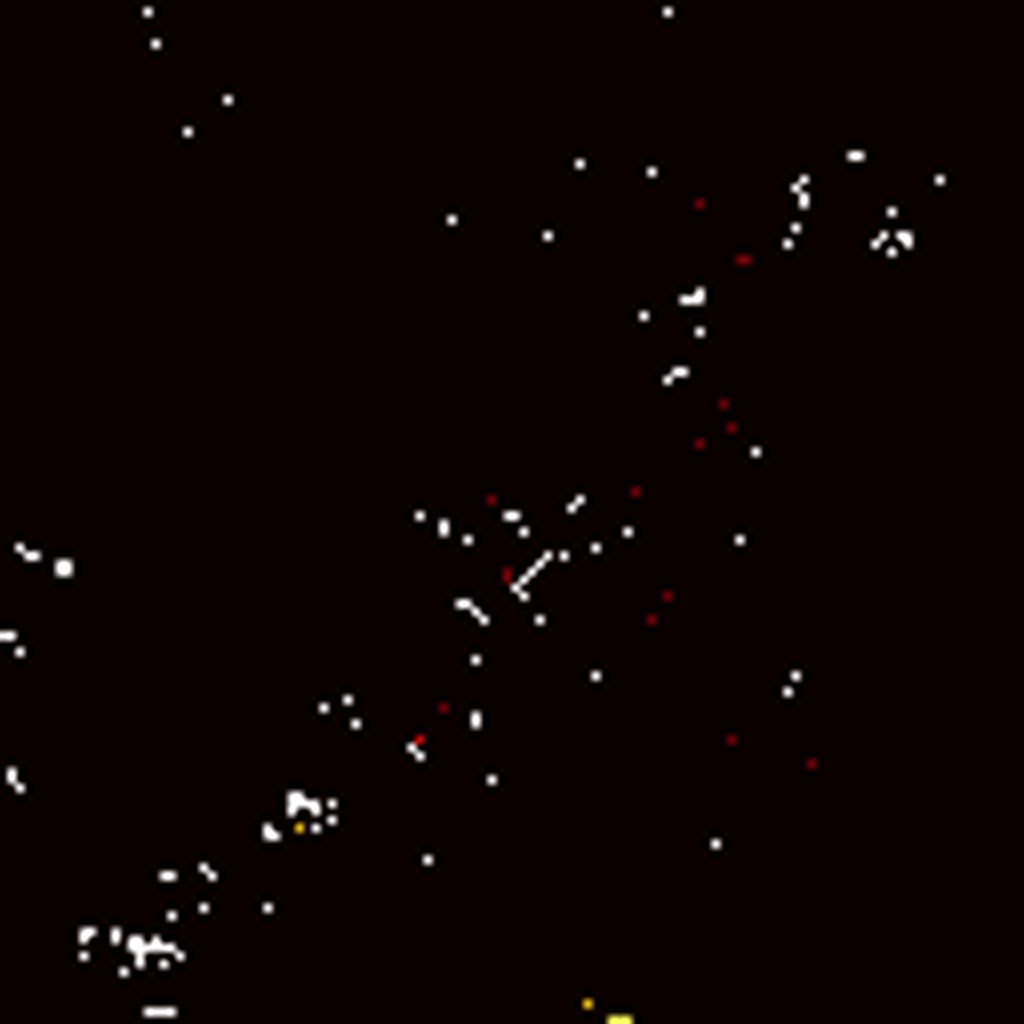}
        \end{subfigure}
        \hfill
        \begin{subfigure}[b]{0.16\textwidth}
            \includegraphics[width=\textwidth]{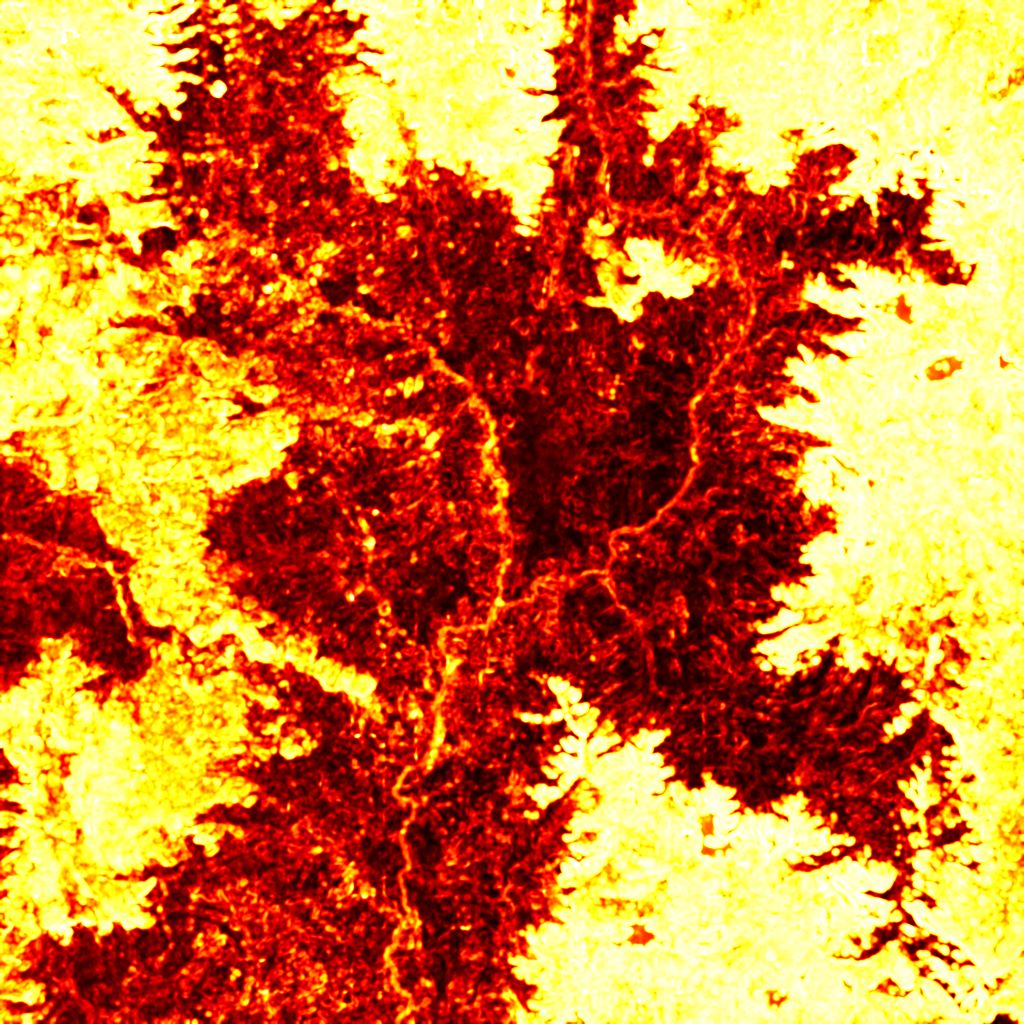}
        \end{subfigure}
        \hfill
        \begin{subfigure}[b]{0.16\textwidth}
            \includegraphics[width=\textwidth]{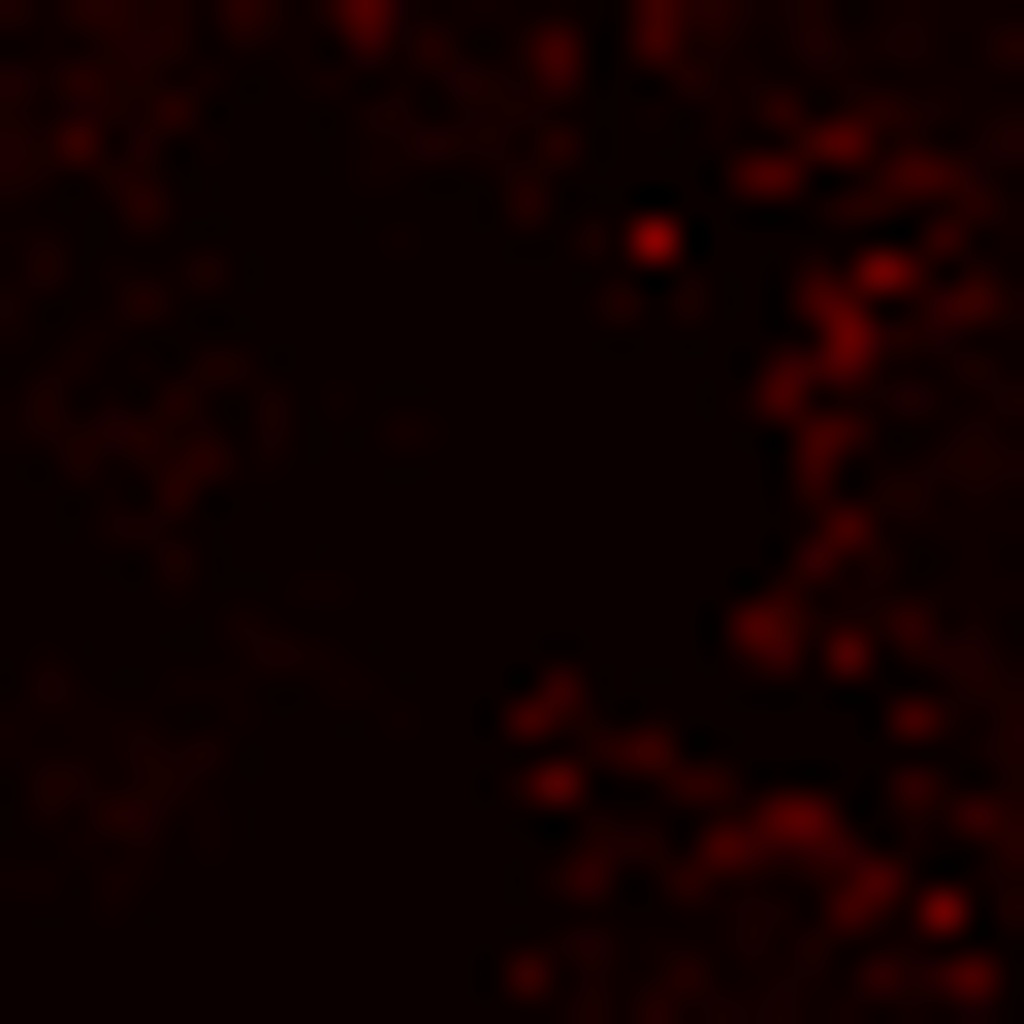}
        \end{subfigure}
        \hfill
        \begin{subfigure}[b]{0.16\textwidth}
            \includegraphics[width=\textwidth]{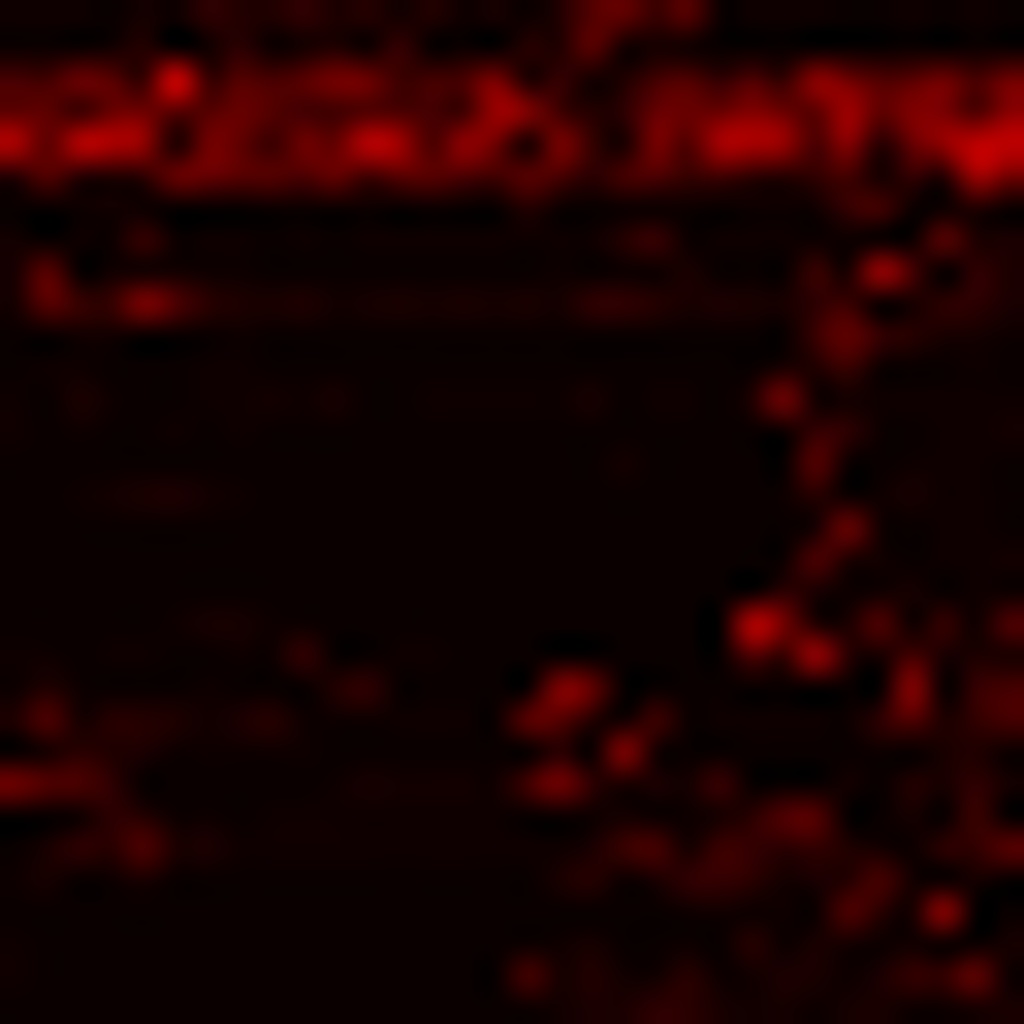}
        \end{subfigure}
        \hfill
        \begin{subfigure}[b]{0.16\textwidth}
            \includegraphics[width=\textwidth]{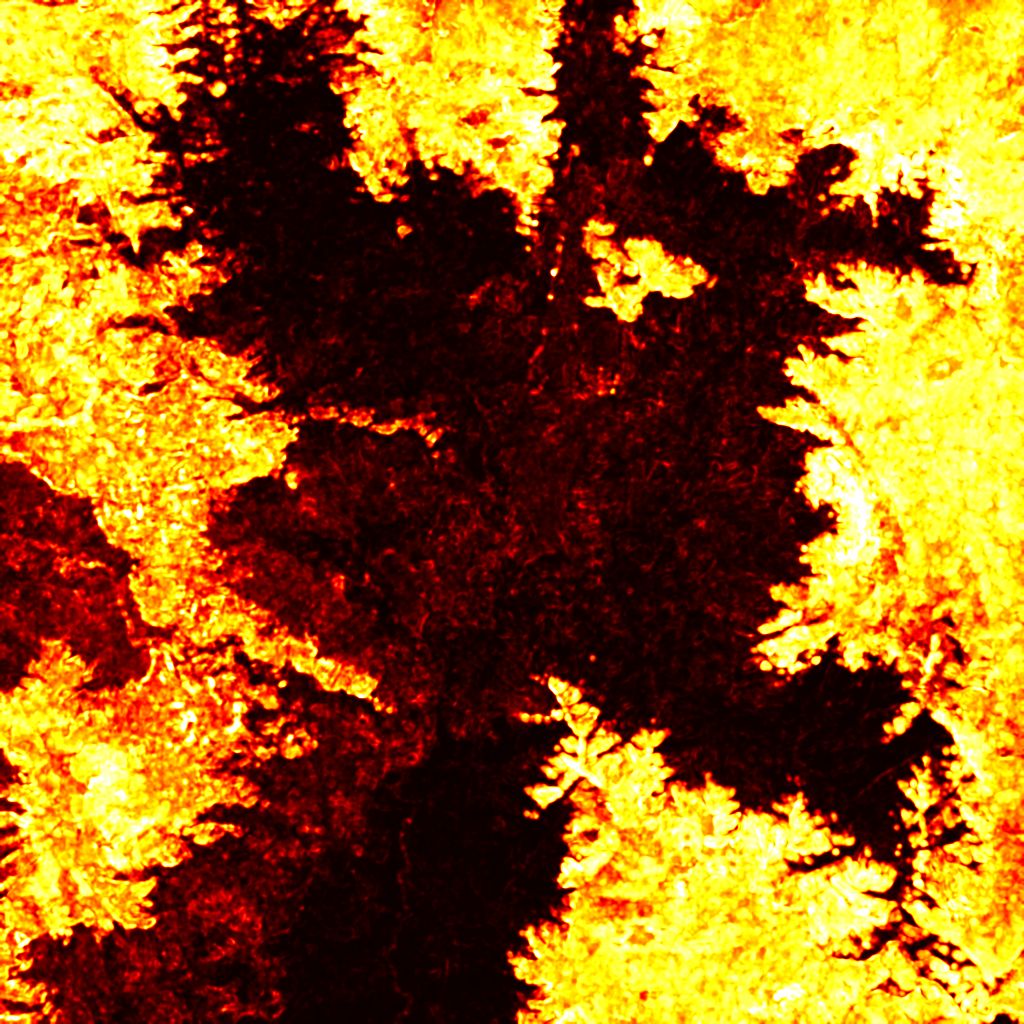}
        \end{subfigure}
    \subcaption{SNP dataset heatmap comparison. The top row shows heatmaps for a bushfire event, while the bottom row shows out-of-season snowfall.}
    \label{fig:snp_heatmaps}
    \end{subfigure}

    \vspace{0.8em}

    \begin{subfigure}[b]{\textwidth}
        \begin{subfigure}[b]{0.16\textwidth}
            \includegraphics[width=\textwidth]{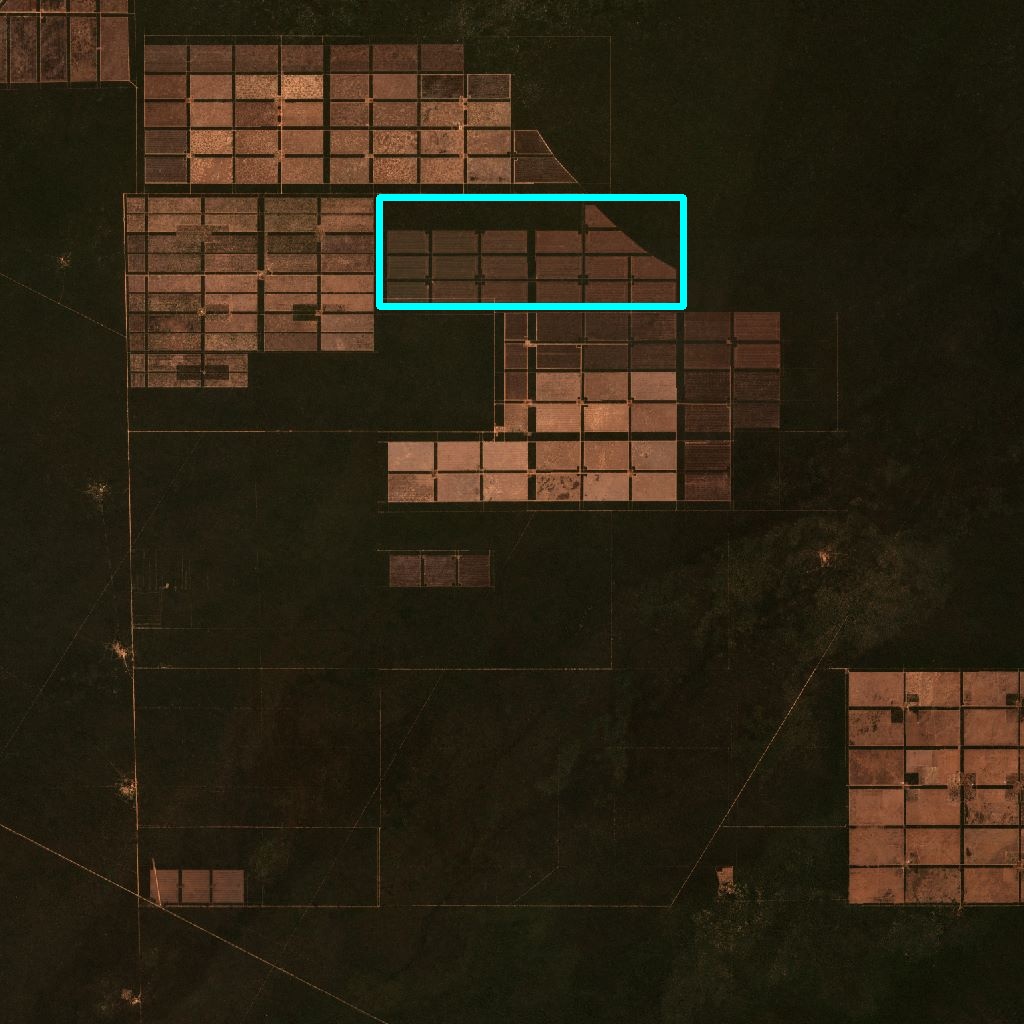}
        \end{subfigure}
        \hfill
        \begin{subfigure}[b]{0.16\textwidth}
            \includegraphics[width=\textwidth]{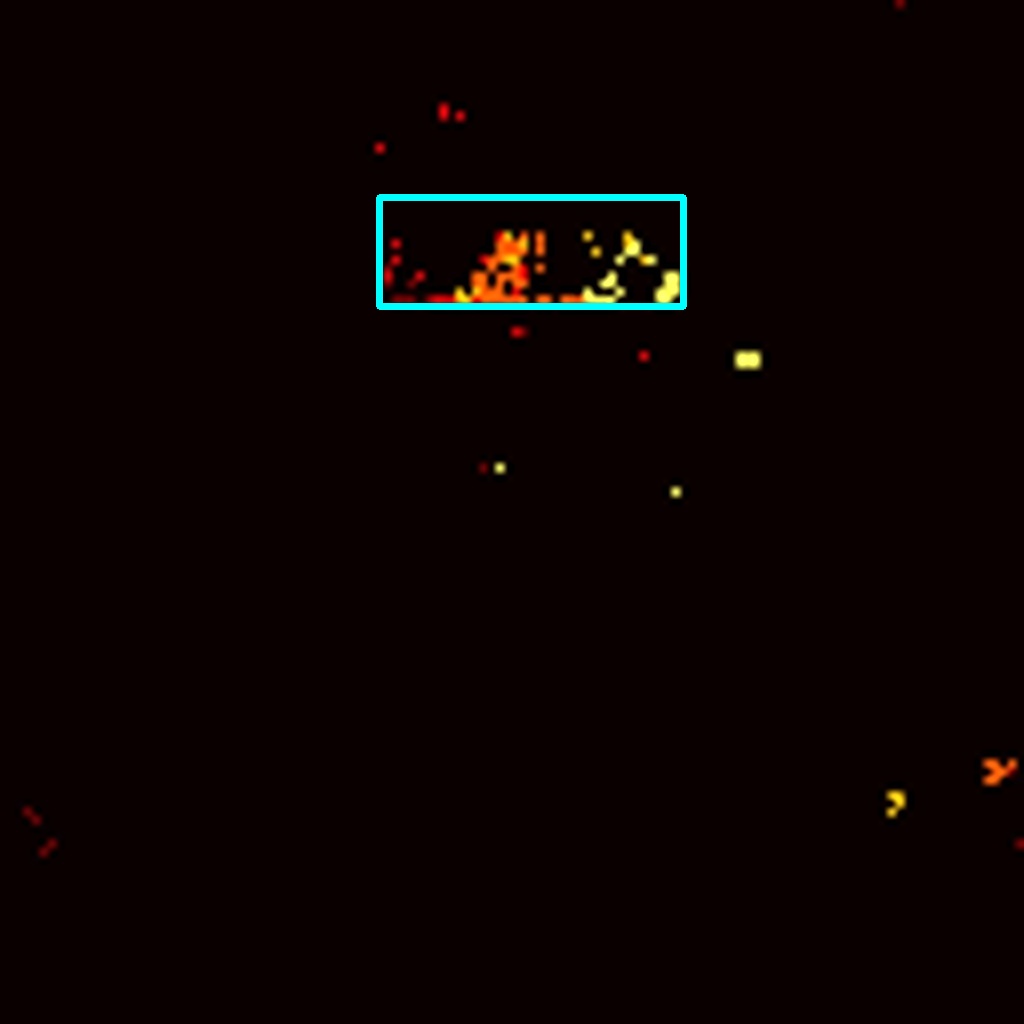}
        \end{subfigure}
        \hfill
        \begin{subfigure}[b]{0.16\textwidth}
            \includegraphics[width=\textwidth]{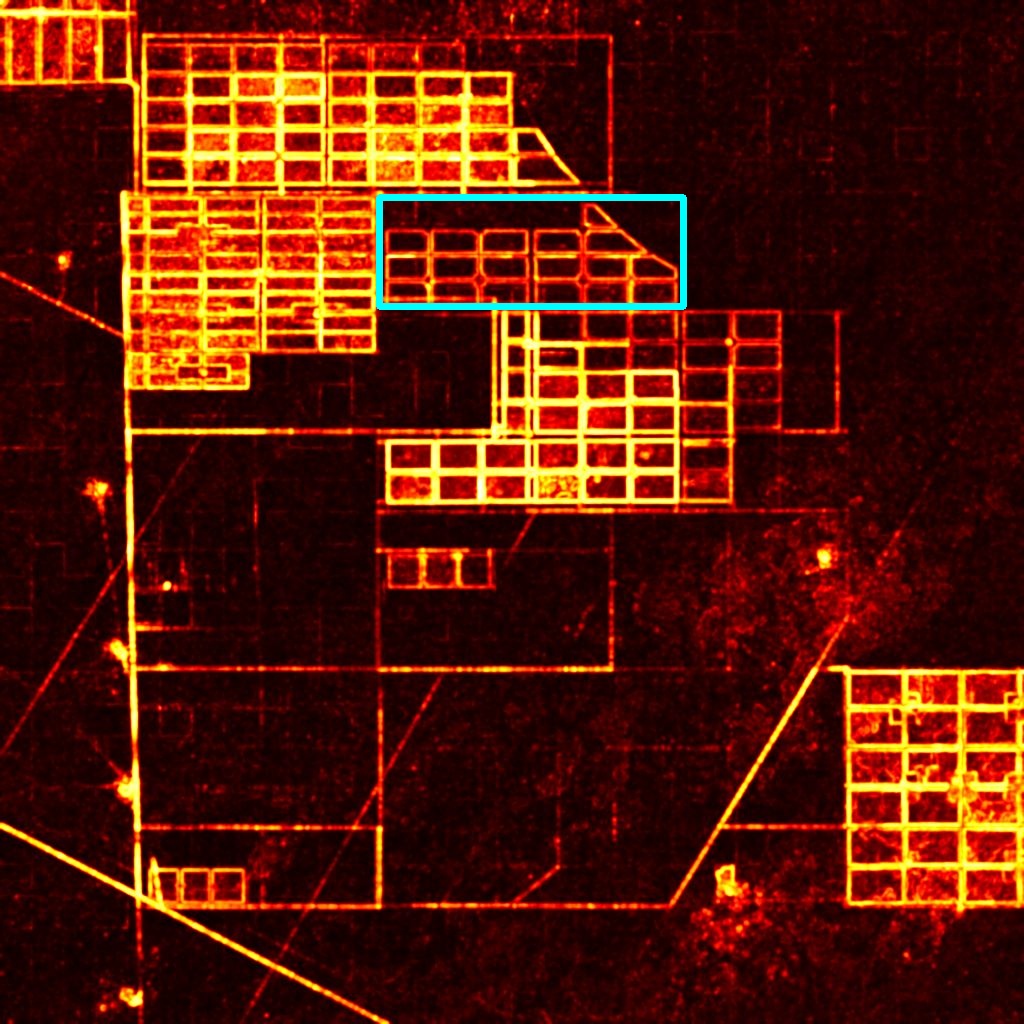}
        \end{subfigure}
        \hfill
        \begin{subfigure}[b]{0.16\textwidth}
            \includegraphics[width=\textwidth]{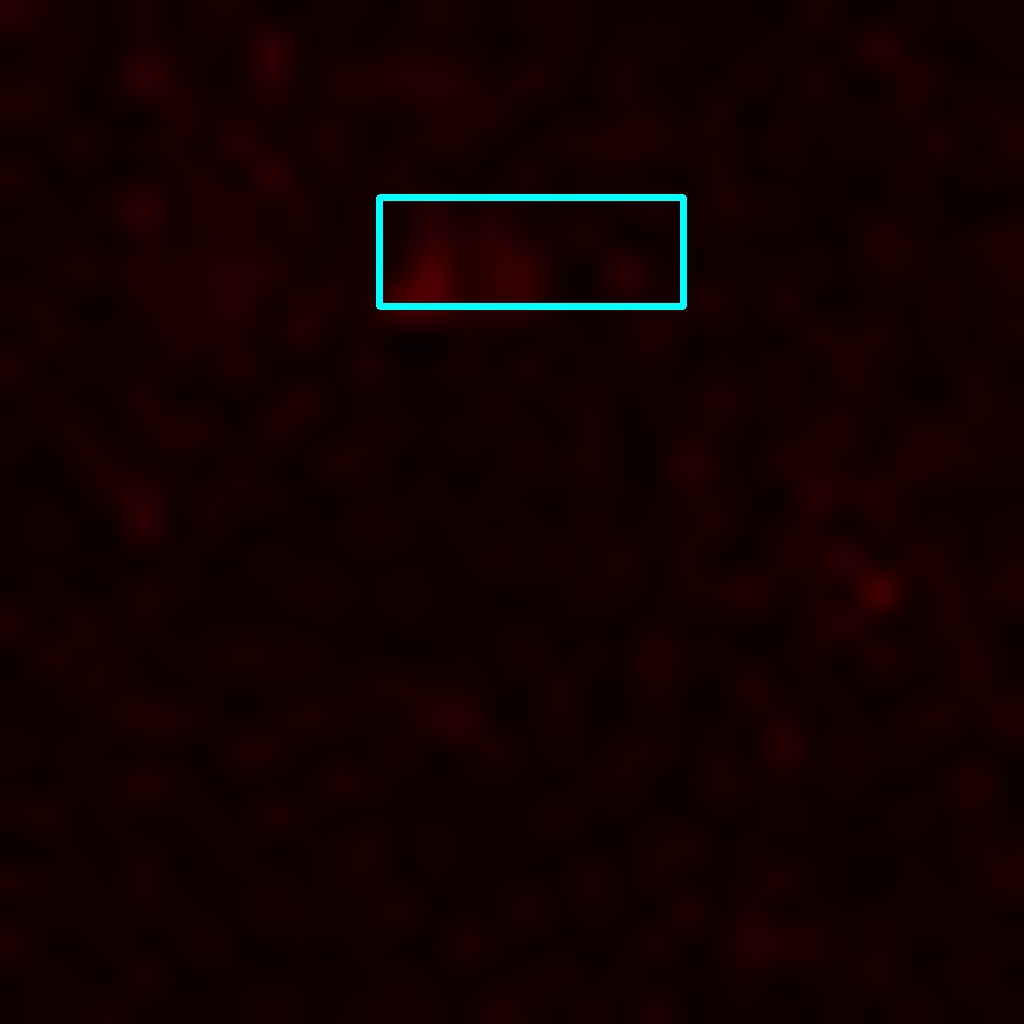}
        \end{subfigure}
        \hfill
        \begin{subfigure}[b]{0.16\textwidth}
            \includegraphics[width=\textwidth]{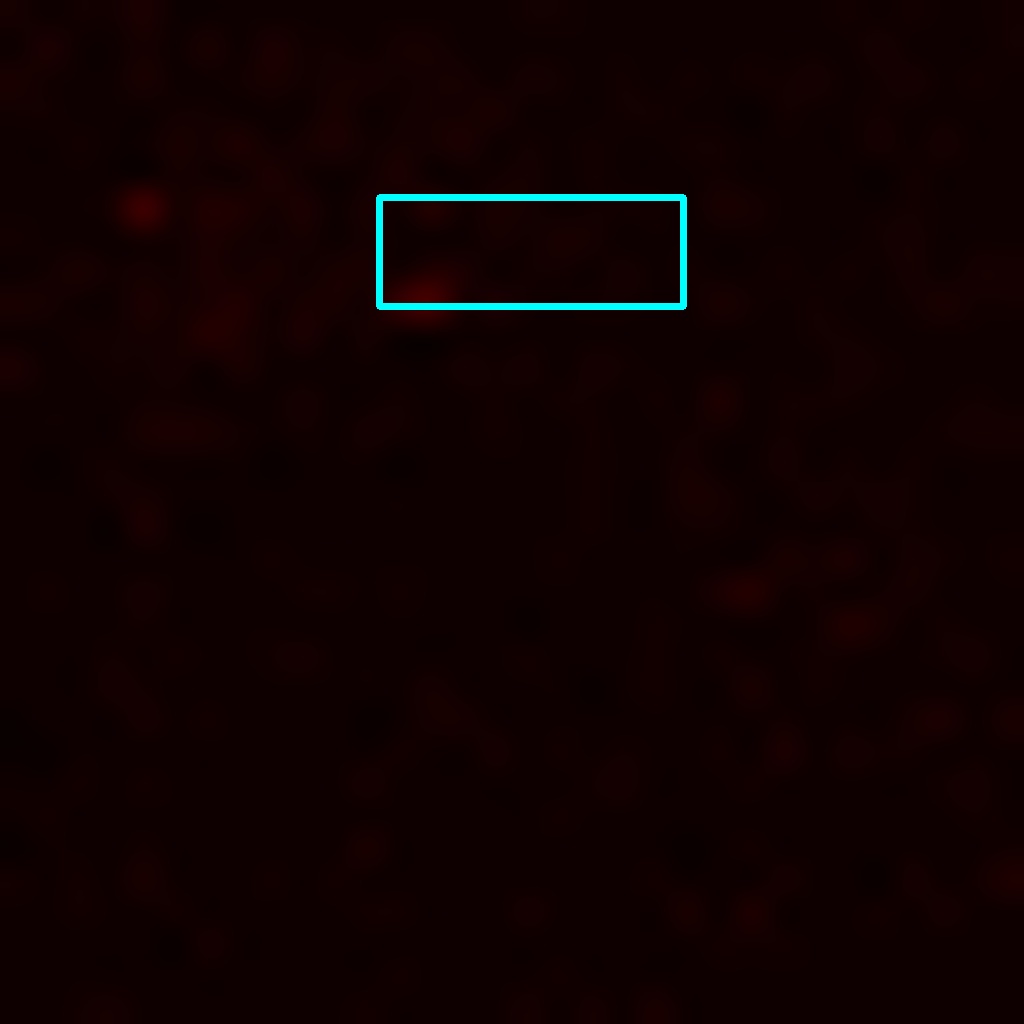}
        \end{subfigure}
        \hfill
        \begin{subfigure}[b]{0.16\textwidth}
            \includegraphics[width=\textwidth]{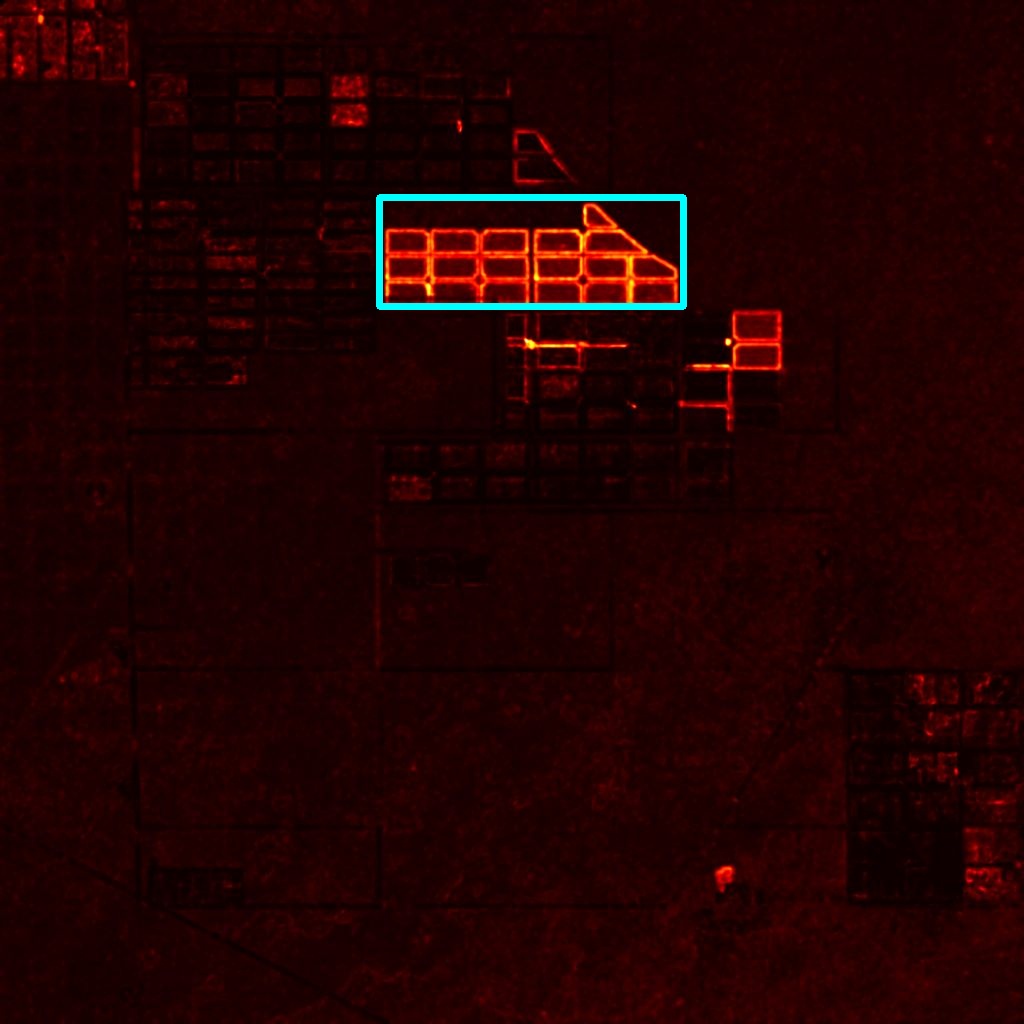}
        \end{subfigure}
    \subcaption{Gran Chaco dataset heatmap comparison, showing deforestation mapping. The cyan box highlights the primary area of new deforestation since the training period.}
    \label{fig:gc_heatmaps}
    \end{subfigure}

    \vspace{0.8em}

    \begin{subfigure}[b]{\textwidth}
    \centering
        \begin{subfigure}[b]{0.16\textwidth}
            \includegraphics[width=\textwidth]{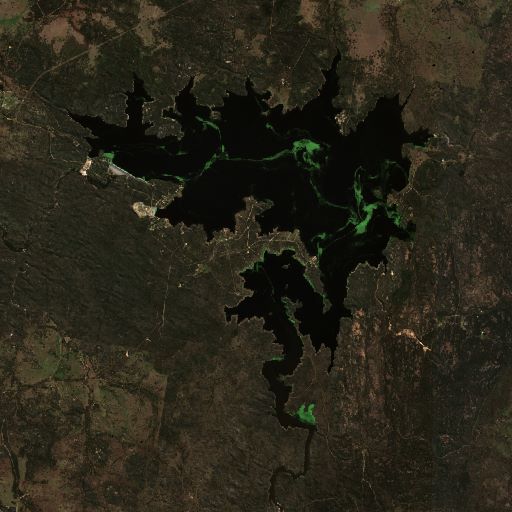}
        \end{subfigure}%
        \hfill
        \hspace{0.16\textwidth}
        \hfill
        \begin{subfigure}[b]{0.16\textwidth}
            \includegraphics[width=\textwidth]{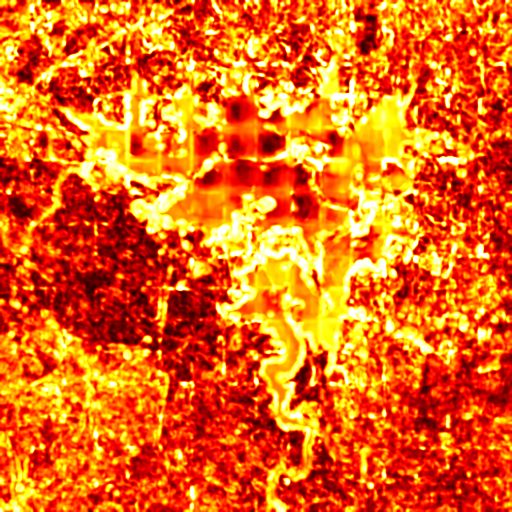}
        \end{subfigure}%
        \hfill
        \begin{subfigure}[b]{0.16\textwidth}
            \includegraphics[width=\textwidth]{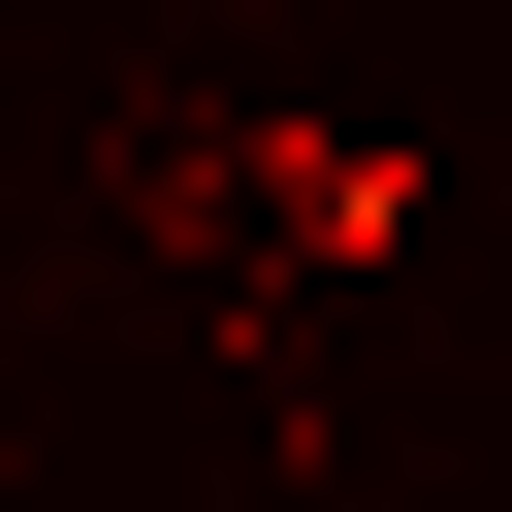}
        \end{subfigure}%
        \hfill
        \begin{subfigure}[b]{0.16\textwidth}
            \includegraphics[width=\textwidth]{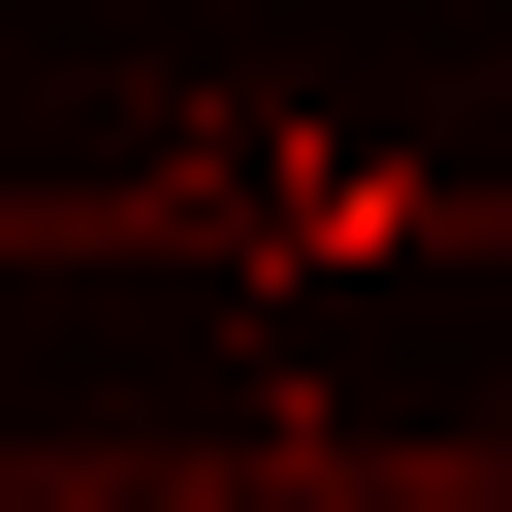}
        \end{subfigure}%
        \hfill
        \begin{subfigure}[b]{0.16\textwidth}
            \includegraphics[width=\textwidth]{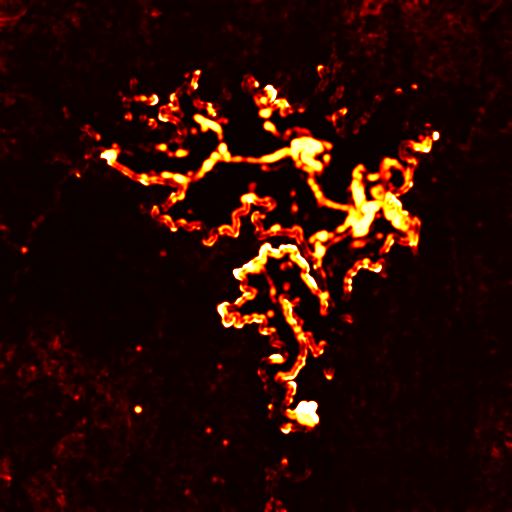}
        \end{subfigure}%
    \end{subfigure}
    
    \vspace{0.4em}

    \begin{subfigure}[b]{\textwidth}
    \centering
        \begin{subfigure}[b]{0.16\textwidth}
            \includegraphics[width=\textwidth]{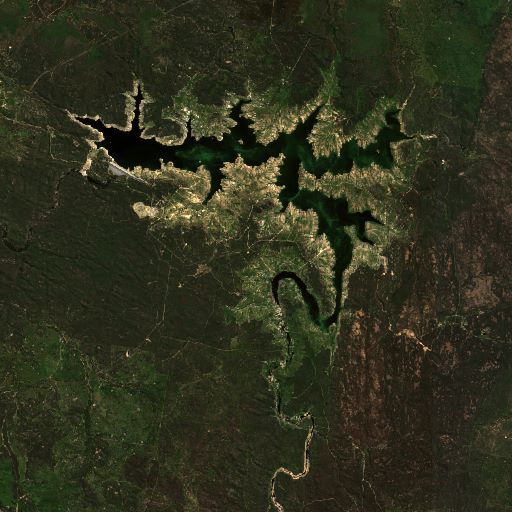}
        \end{subfigure}%
        \hfill
        \hspace{0.16\textwidth}
        \hfill
        \begin{subfigure}[b]{0.16\textwidth}
            \includegraphics[width=\textwidth]{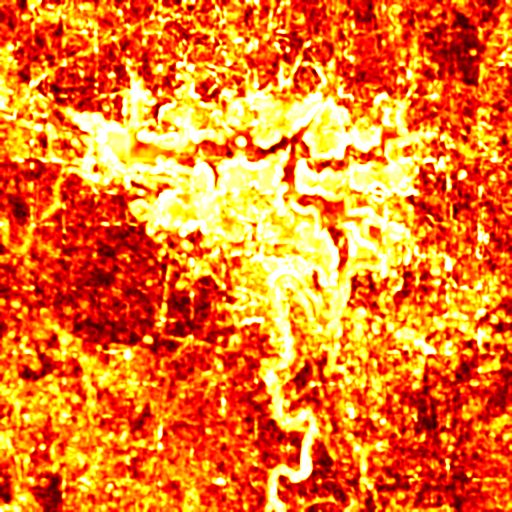}
        \end{subfigure}%
        \hfill
        \begin{subfigure}[b]{0.16\textwidth}
            \includegraphics[width=\textwidth]{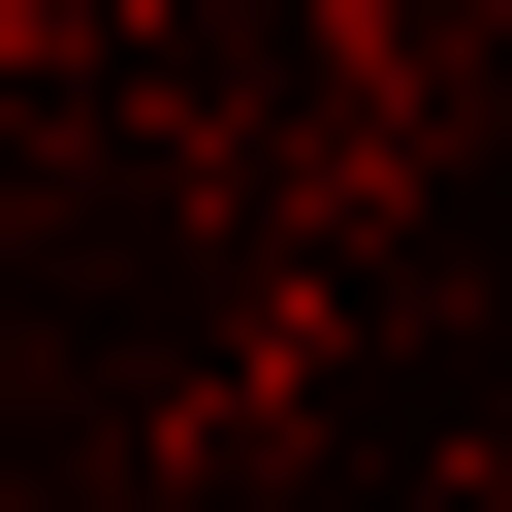}
        \end{subfigure}%
        \hfill
        \begin{subfigure}[b]{0.16\textwidth}
            \includegraphics[width=\textwidth]{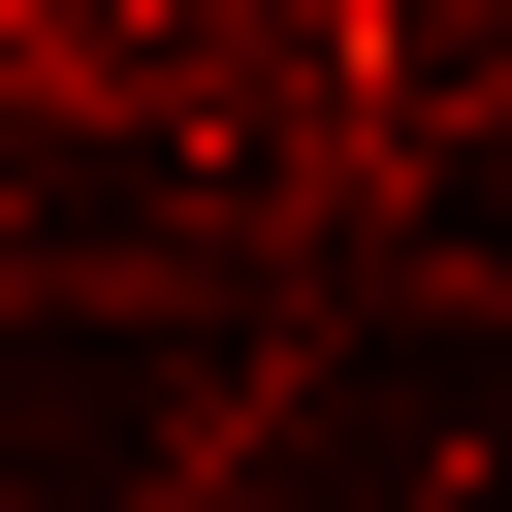}
        \end{subfigure}%
        \hfill
        \begin{subfigure}[b]{0.16\textwidth}
            \includegraphics[width=\textwidth]{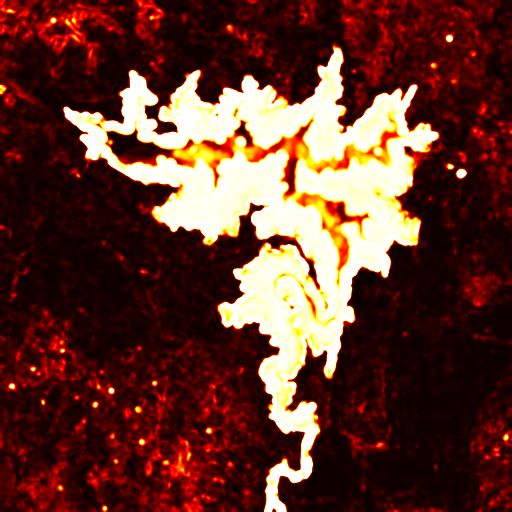}
        \end{subfigure}
    \subcaption{Lake Copeton dataset heatmap comparison. The top row shows heatmaps for an algal bloom, while the bottom row shows demonstrates drought mapping. COLD is absent from this dataset due to no pre-test images in Lake Copeton dataset.}
    \label{fig:lc_heatmaps}
    \end{subfigure}

    \vspace{0.8em}

    \begin{subfigure}[b]{\textwidth}
        \begin{subfigure}[b]{0.16\textwidth}
            \includegraphics[width=\textwidth]{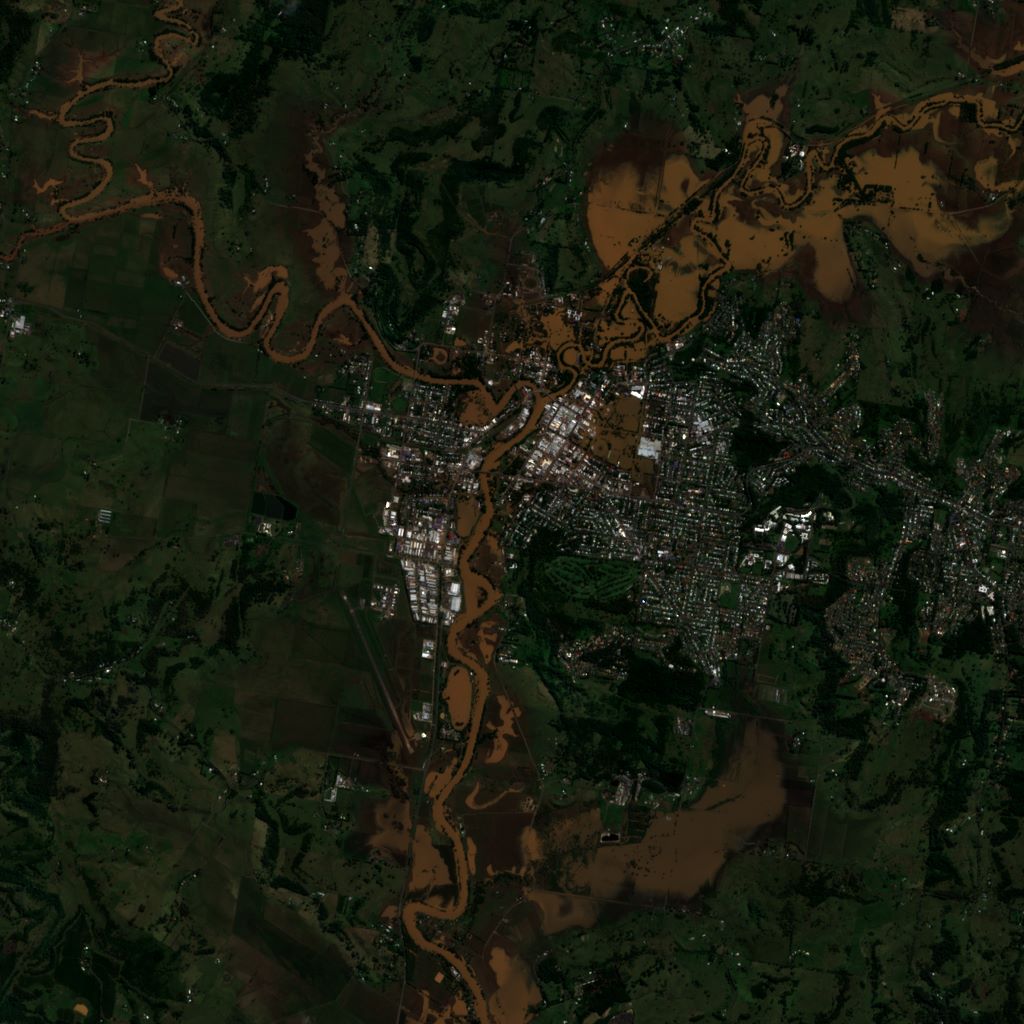}
        \end{subfigure}
        \hfill
        \begin{subfigure}[b]{0.16\textwidth}
            \includegraphics[width=\textwidth]{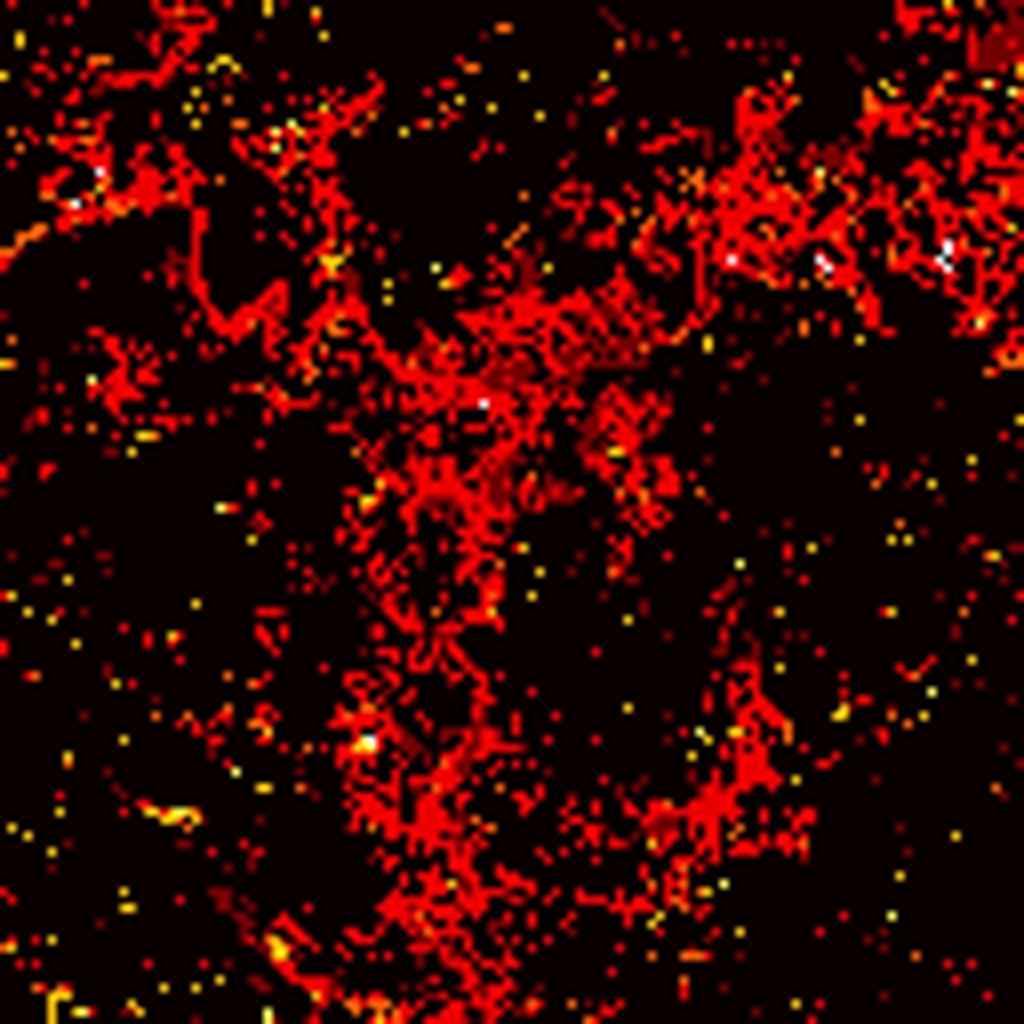}
        \end{subfigure}
        \hfill
        \begin{subfigure}[b]{0.16\textwidth}
            \includegraphics[width=\textwidth]{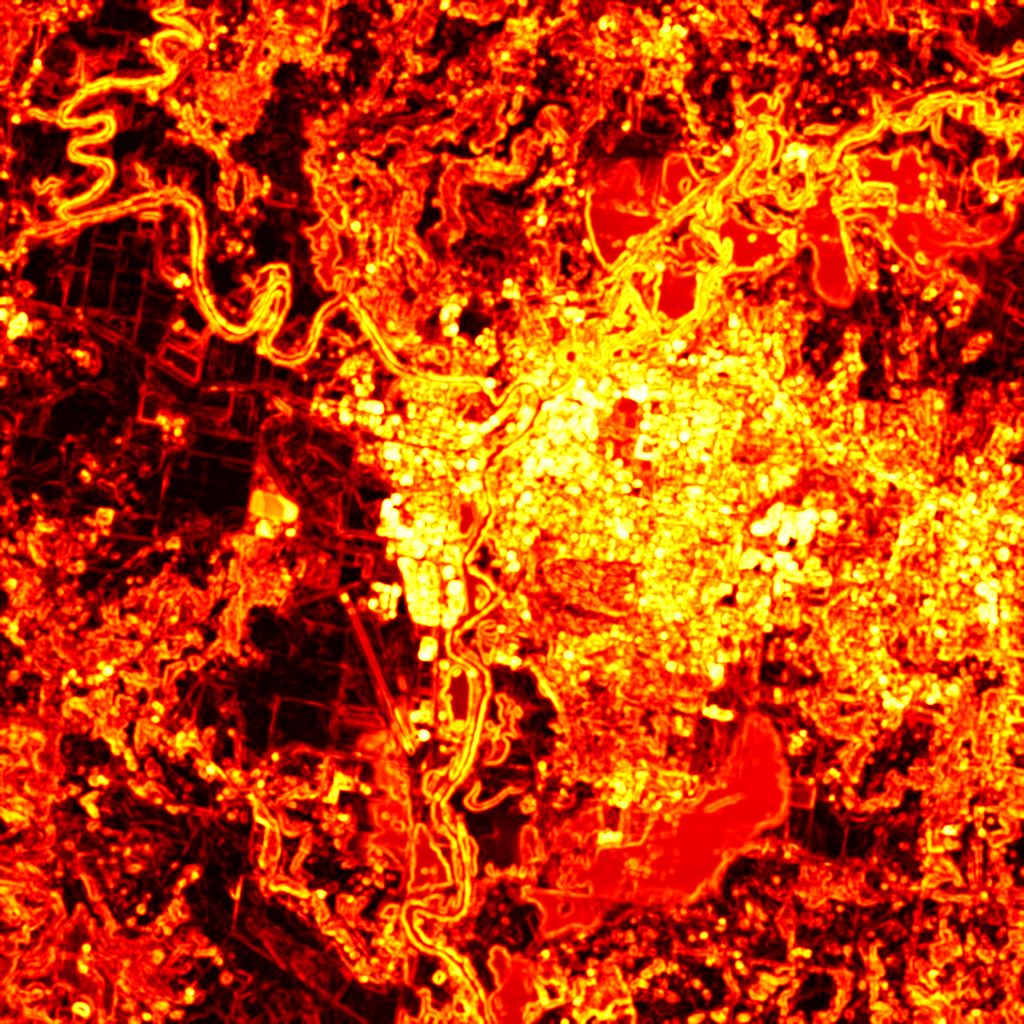}
        \end{subfigure}
        \hfill
        \begin{subfigure}[b]{0.16\textwidth}
            \includegraphics[width=\textwidth]{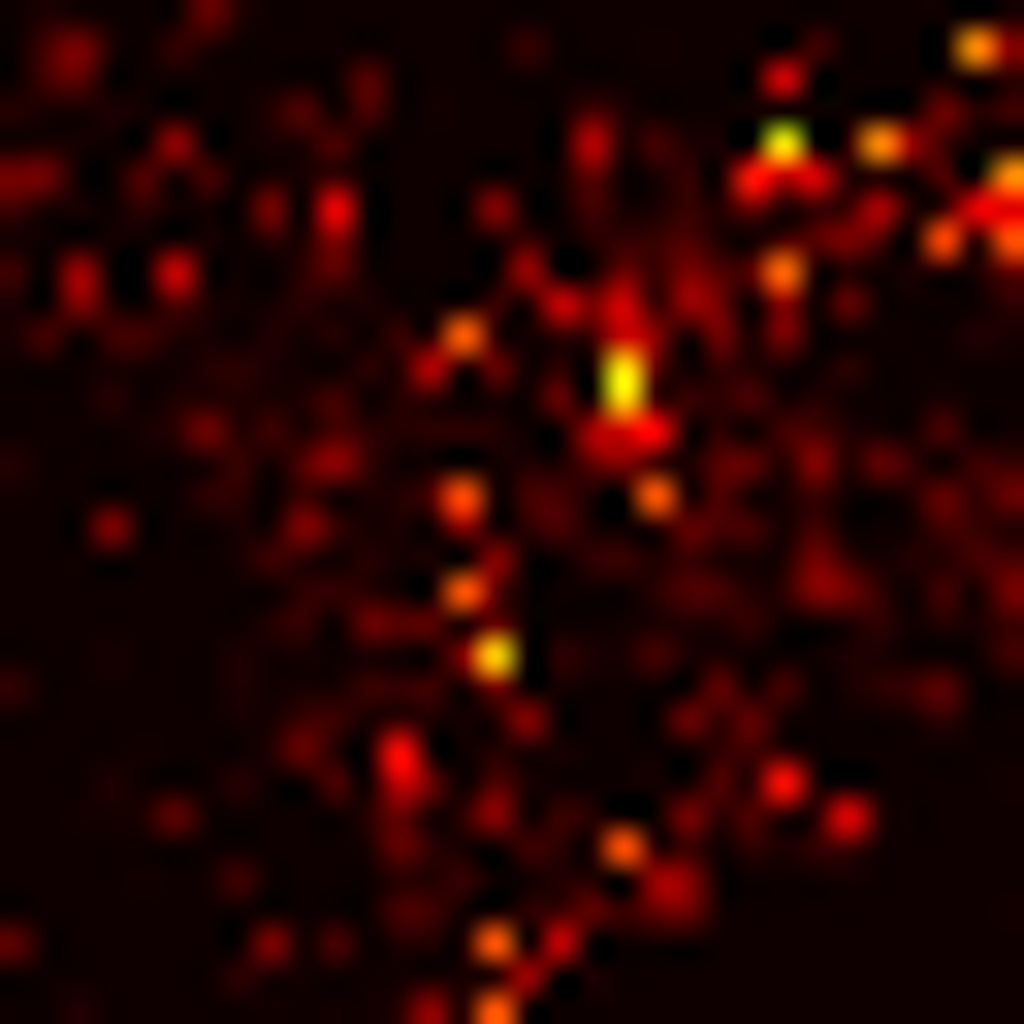}
        \end{subfigure}
        \hfill
        \begin{subfigure}[b]{0.16\textwidth}
            \includegraphics[width=\textwidth]{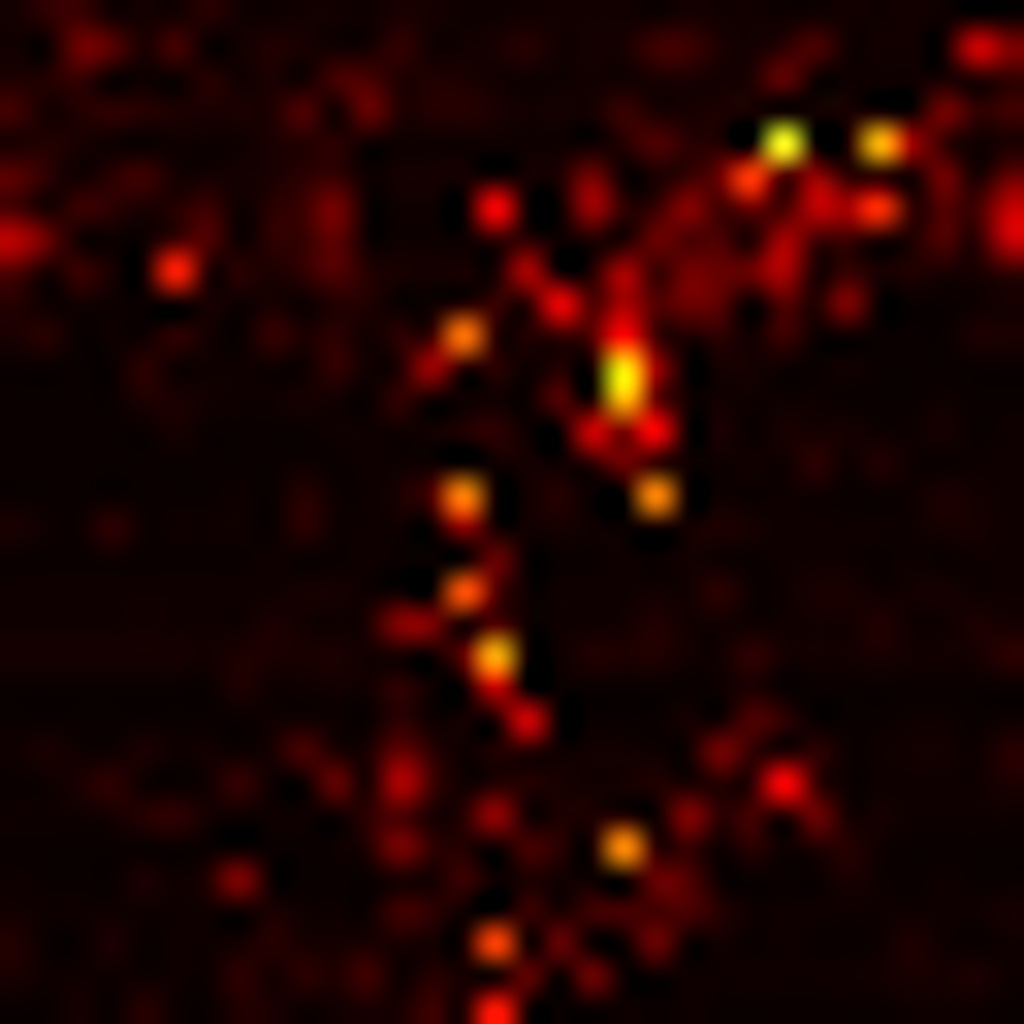}
        \end{subfigure}
        \hfill
        \begin{subfigure}[b]{0.16\textwidth}
            \includegraphics[width=\textwidth]{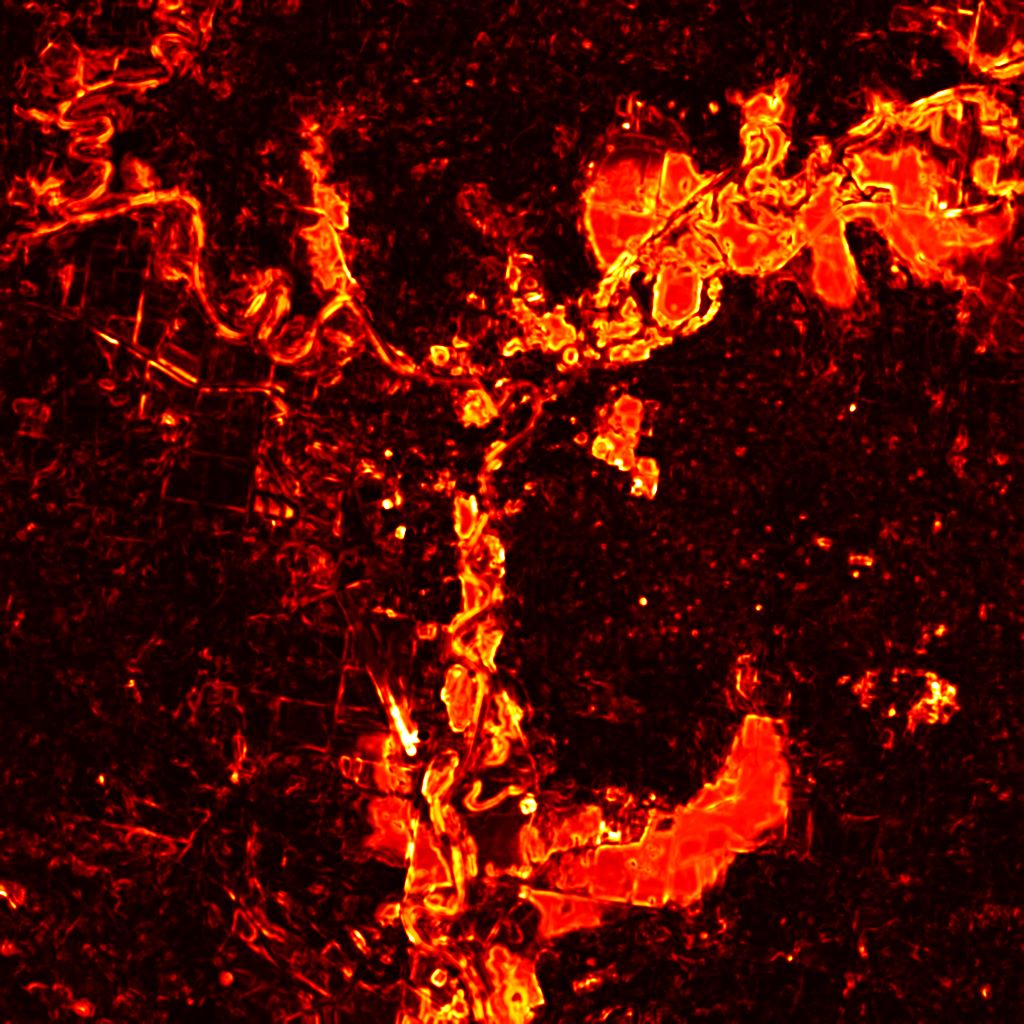}
        \end{subfigure}
    \end{subfigure}

    \vspace{0.2em}
    
    \begin{subfigure}[b]{\textwidth}
        \begin{subfigure}[b]{0.16\textwidth}
            \centering
            Hazard Image
        \end{subfigure}
        \hfill
        \begin{subfigure}[b]{0.16\textwidth}
            \centering
            COLD
        \end{subfigure}
        \hfill
        \begin{subfigure}[b]{0.16\textwidth}
            \centering
            cVAE
        \end{subfigure}
        \hfill
        \begin{subfigure}[b]{0.16\textwidth}
            \centering
            RaVAEn-Global
        \end{subfigure}
        \hfill
        \begin{subfigure}[b]{0.16\textwidth}
            \centering
            RaVAEn-Local
        \end{subfigure}
        \hfill
        \begin{subfigure}[b]{0.16\textwidth}
            \centering
            SHAZAM
        \end{subfigure}
    \end{subfigure}

    \vspace{0.6em}

    \begin{subfigure}{0.6\textwidth}
        \centering
        \includegraphics[width=\textwidth]{Images/Results/Visual_Comparison/hot_colourbar.png}
        \subcaption{Lismore dataset heatmap comparison, showing flood mapping.}
        \label{fig:lm_heatmaps}
    \end{subfigure}%
    \caption{Change/hazard heatmap examples of different hazards from each dataset.} 

\end{figure*}

The mapping results are grouped by dataset. Figure \ref{fig:snp_heatmaps} compares the heatmaps of each model on the SNP dataset, using a live bushfire and out-of-season snowfall as examples. Figure \ref{fig:gc_heatmaps} shows a small section of deforestation mapping on the Gran Chaco dataset, while Figure \ref{fig:lc_heatmaps} shows an algal bloom and drought in Lake Copeton. Finally, Figure \ref{fig:lm_heatmaps} compares the flood mapping capabilities of each model.

SHAZAM has distinctively strong performance when mapping both the bushfires and the out-of-season snowfall in Figure \ref{fig:snp_heatmaps}. For bushfire detection, SHAZAM achieves pixel-precise mapping of both burned areas and active fires/smoke plumes, effectively differentiating between active fire locations and their effects. While COLD's disturbance probability map identifies affected areas, it lacks the ability to distinguish active fire fronts. The cVAE output, though highlighting active fires, fails to maintain the scene's spatial structure, making it difficult to differentiate fire-related features from the background. RaVAEn-Global's strength in capturing dynamic events is evident in its rough detection of active fires and smoke, but its lower-resolution output makes it much less precise, and it's dynamic design misses crucial burned regions. RaVAEn-Local, designed for patch-wise scoring, introduces considerable noise by incorporating signals from all three prior temporal images, in contrast to RaVAEn-Global's more focused single-image comparison approach.

For the out-of-season snowfall, the conventional detection methods struggle. The COLD algorithm does not identify the snow coverage, while cVAE, despite leveraging the high reflectance of snow, is unable to suppress the background terrain. The RaVAEn variants demonstrate some success in delineating the snowfront boundaries, but struggle to map this out-of-season anomaly. In contrast, SHAZAM is very effective at suppressing the background while accentuating the unexpected snow, demonstrating a strong ability to model seasonality.

In analysing deforestation mapping in the Gran Chaco dataset (Figure \ref{fig:gc_heatmaps}), conventional methods show varying limitations. COLD sporadically maps some of the new deforestation within the cyan box while effectively suppressing the background, but fails to reveal true structure. cVAE highlights all deforested areas, including those from the training dataset, due to its inability to generate accurate structural representations of the region of interest. Both RaVAEn variants achieve good background suppression, but are unable to map the newly deforested areas. RaVAEn-Global's mapping is slightly more visible than RaVAEn-local but the deforestation patterns are still not visible, a limitation of its coarse patch-wise resolution.

SHAZAM demonstrates superior detection of newly deforested regions, with clear mapping of recent land-use changes. Notably, it identifies anomalies within existing agricultural fields that correspond to recent re-clearing or burning activities. The detection of two fields outside and to the bottom right of the cyan box, which were cleared before the test period, reveals an important characteristic of SHAZAM's methodology. As the baseline represents an average image across the entire training period, SHAZAM's ability to detect structural changes depends on this reference, suggesting that some recent features may not be captured if they are partially incorporated into the baseline.

The Lake Copeton dataset demonstrates SHAZAM's ability to map algal blooms, a type of hazard that the other models had not previously evaluated (Figure \ref{fig:lc_heatmaps}). cVAE maps the blooms within the water but struggles to distinguish them from the background. The RaVAEn variants show some ability to map the blooms, but at the noticeably lower resolution. Although they capture larger clusters of algae, these methods miss the finer strains present in the scene. SHAZAM, in contrast, precisely maps all algal blooms present on the surface. SHAZAM maintains a high spatial resolution that captures both the larger clusters, and the finer-scale threads between them and along the lake's edge.

The drought example shows marked differences in mapping and detection capabilities (Figure \ref{fig:lc_heatmaps}, bottom row). cVAE and both RaVAEn variants fail to map the drought conditions. While cVAE's failure stems from poor ROI structure representation, RaVAEn's limitation is inherent to its design focus on rapid ``change events". By comparing the current satllite image to only a small window of recent images, RaVAEn lacks the historical context to establish normal water levels for the lake, making it unsuitable for detecting and mapping hazards that gradually develop, such as drought. SHAZAM, using its baseline-derived ROI with seasonal translation, clearly delineates the drought-affected region. The increased background signal in the anomaly heatmap is consistent with drought conditions, reflecting the broader impact of extreme dry seasons on the surrounding terrain.

The Lismore dataset demonstrates the capability of each method in flood mapping (Figure \ref{fig:lm_heatmaps}). COLD, despite not being specifically designed for flood detection or immediate hazard response, performs notably well in identifying flooded rivers and shows clusters of anomalous pixels in inundated areas, though it fails to capture the complete extent of the flood. cVAE is ineffective on this dataset, while the RaVAEn variants identify the larger flooded regions but miss finer details due to their characteristic lower resolution. RaVAEn-local shows a better ability to suppress the background in this case. SHAZAM achieves comprehensive flood mapping with high spatial resolution, clearly mapping both all flood-affected areas.

Across diverse datasets and hazard types, SHAZAM demonstrates consistently superior mapping capabilities compared to existing methods. Although COLD shows some ability to map floods, fires, and some deforestation, its reliance on multiple disturbance observations often results in partial or missed hazard detection. cVAE exhibits generally poor performance across all hazards; despite some understanding of seasonality, it struggles to learn the spatial structure of the scene and suppress background features. The RaVAEn variants, though effective at background suppression and mapping abrupt changes, are limited by their coarse resolution and reliance on recent temporal context, particularly evident in their inability to detect gradual changes such as drought and deforestation. Although useful for detecting dynamic change events, their lack of seasonal understanding is further demonstrated by their inability to detect out-of-season snowfall. 

SHAZAM effectively combines seasonal modelling with high spatial resolution to accurately map all example hazards. This includes immediate events such as floods and fires, to more gradual changes such as drought and deforestation. This consistent performance across diverse scenarios, coupled with its ability to capture both fine-scale details and broader patterns, demonstrates SHAZAM's versatile and seasonally-aware approach to hazard mapping and detection.

\subsection{Ablation Study}

\begin{figure}[t]
    \centering
    
    % Row with 6 images
    \begin{subfigure}[b]{\linewidth}
        \begin{subfigure}[b]{0.32\textwidth}
            \includegraphics[width=\textwidth]{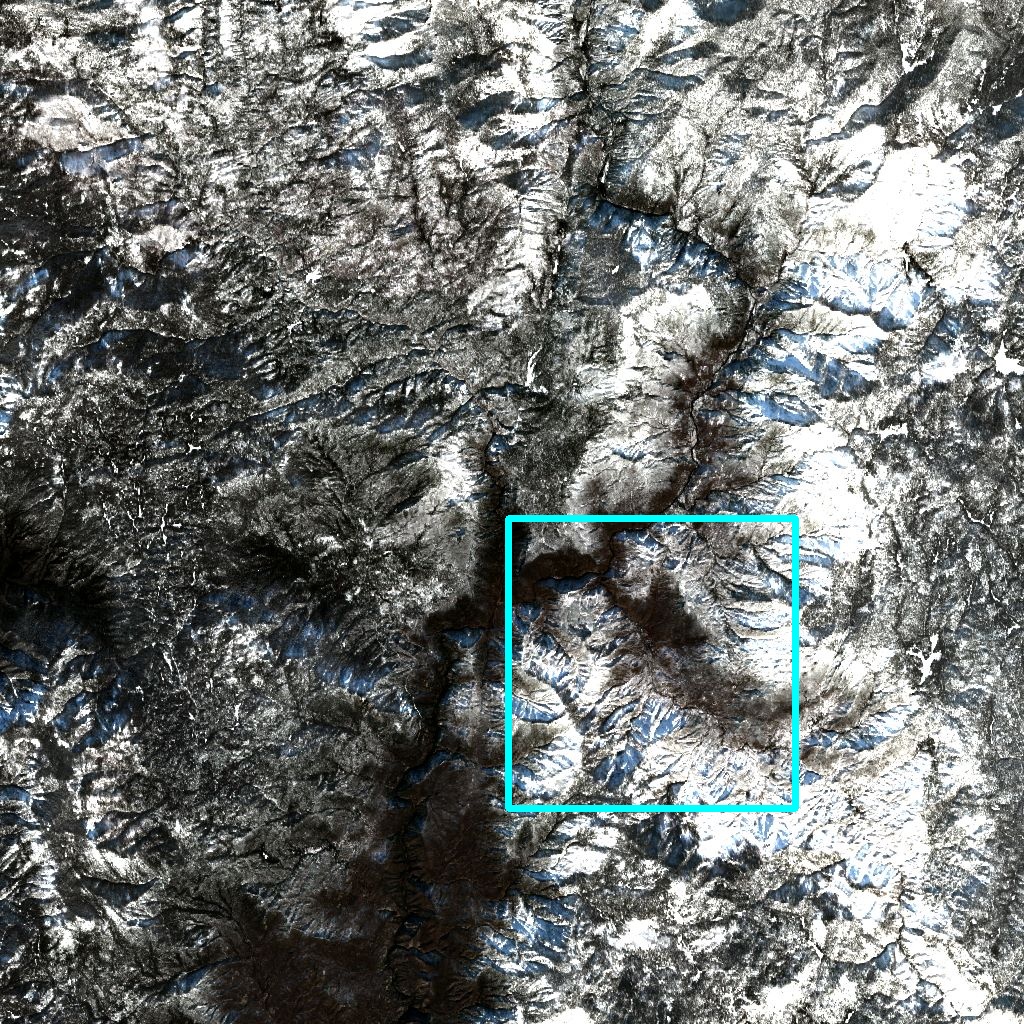}
        \end{subfigure}
        \hfill
        \begin{subfigure}[b]{0.32\textwidth}
            \includegraphics[width=\textwidth]{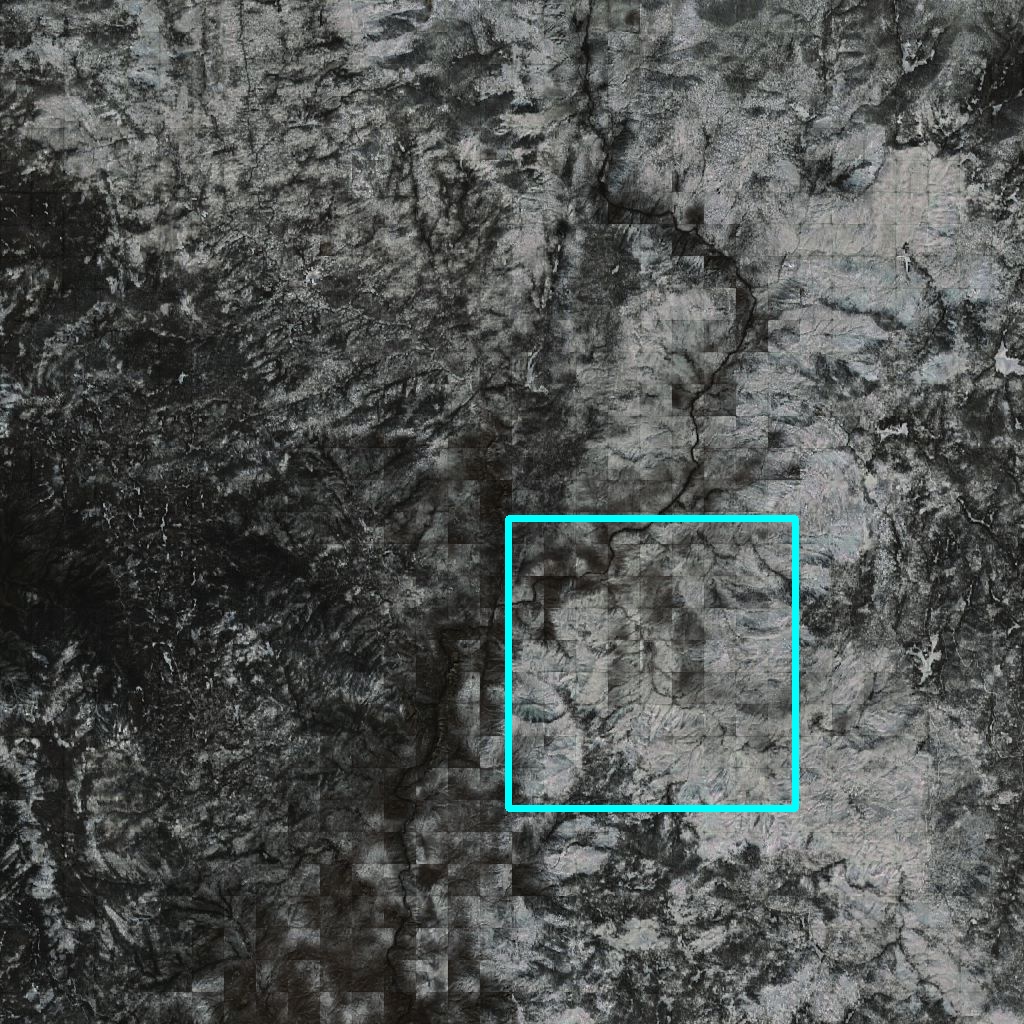}
        \end{subfigure}
        \hfill
        \begin{subfigure}[b]{0.32\textwidth}
            \includegraphics[width=\textwidth]{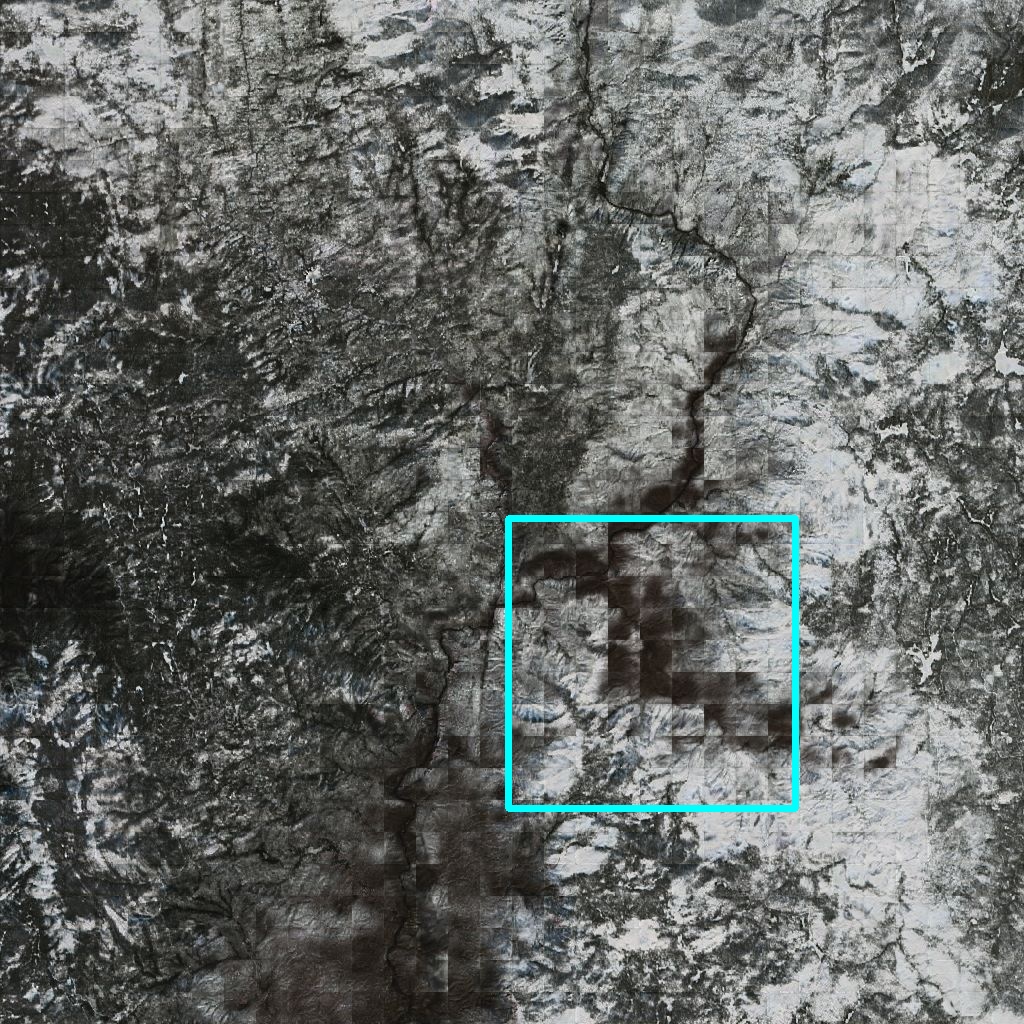}
        \end{subfigure}
    \end{subfigure}

    \begin{subfigure}[b]{\linewidth}
        \begin{subfigure}[b]{0.32\textwidth}
            \centering
            Real Image
        \end{subfigure}
        \hfill
        \begin{subfigure}[b]{0.32\textwidth}
            \centering
            No Pos. Encodings
        \end{subfigure}
        \hfill
        \begin{subfigure}[b]{0.32\textwidth}
            \centering
            SHAZAM
        \end{subfigure}
    \end{subfigure}
    
    \caption{Comparison of SHAZAM's generated output with and without positional encodings. The cyan box shows an area where removing position encodings leads to a coarser representation of expected snow cover.}
    \label{fig:as_posvnopos}
\end{figure}

The effects of removing positional encodings, seasonal encodings, SSIM score and the seasonal threshold are shown in Table \ref{tab:ablation_study}. Removing the position encodings shows minimal degradation in F1 scores and even marginal improvements in AUPRC (a 0.026 increase for SNP), qualitative analysis reveals that positional encodings contribute to more precise spatial representations (Figure \ref{fig:as_posvnopos}), also shown by slightly improved MAE and SSIM scores in the training phase. This suggests that the position encodings may improve feature representation at a local level, though their impact on final classification metrics is dataset-dependent. For smaller SITS datasets, there may be a risk of overfitting.

\begin{table*}[t]
\centering
\caption{SHAZAM Ablation Study - F1 and AUPRC Scores.}
\begin{tabular}{@{}c|cc|cc|cc|cc@{}}
\toprule
\multicolumn{1}{l|}{} & \multicolumn{2}{c|}{SNP}                  & \multicolumn{2}{c|}{Gran Chaco}           & \multicolumn{2}{c|}{Lake Copeton}         & \multicolumn{2}{c}{Lismore}              \\
Model                 & F1           & \multicolumn{1}{l|}{AUPRC} & F1           & \multicolumn{1}{l|}{AUPRC} & F1           & \multicolumn{1}{l|}{AUPRC} & F1           & \multicolumn{1}{l}{AUPRC} \\ \midrule
No Position           & $0.762$      & $\bm{0.788}$               & $0.714$      & $0.993$                    & $0.891$      & $0.923$                    & $0.460$      & $0.390$                   \\
Linear Time           & $0.490$      & $0.684$                    & $0.579$      & $\bm{0.995}$               & $\bm{0.917}$ & $0.954$                    & $0.495$      & $0.392$                   \\
MAE Score             & $0.705$      & $0.755$                    & $\bm{0.876}$ & $0.993$                    & $0.870$      & $\bm{0.965}$               & $0.512$      & $\bm{0.414}$              \\
Flat Threshold        & $0.378$      & $0.666$                    & $0.468$      & $0.987$                    & $0.865$      & $0.958$                    & $\bm{0.533}$ & $0.355$                   \\
SHAZAM                & $\bm{0.771}$ & $0.762$                    & $0.728$      & $0.989$                    & $0.881$      & $0.928$                    & $0.466$      & $0.389$                   \\ \bottomrule
\end{tabular}
\label{tab:ablation_study}
\end{table*}

Replacing the cyclical day-of-year encodings with linear temporal representations reveals more noticeable impacts on model performance. While marginal improvements are seen in the Lake Copeton and Lismore datasets (F1 increases of 0.036 and 0.029, respectively), there is a significant performance drop in SNP (F1 decrease of 0.281 and AUPRC decrease of 0.078) and Gran Chaco (F1 decrease of 0.135). Overall, these results show that cyclical day-of-year encodings provide a robust approach for incorporating seasonality across diverse locations and ROIs, and are particularly effective in regions with more pronounced seasonal changes (such as the SNP dataset).

The comparison between SSIM and MAE scoring mechanisms highlights the robustness of SHAZAM, with different similarity metrics still showing strong results for hazard detection. Using the MAE to provide an anomaly score achieves comparable AUPRC scores to SHAZAM's default SSIM on the SNP and Gran Chaco datasets, and superior AUPRC scores on Lake Copeton and Lismore. In terms of the F1 scores, there is a notable drop in performance on the SNP dataset, but a notable improvement on the Gran Chaco and Lismore datasets. This validates the efficacy of SHAZAM's broader approach of comparing generated images to the true images, and shows that further research into more effective scoring methods offers promise. Nevertheless, SHAZAM's structural difference module does provide an advantage for hazard mapping by capturing the differences in the local structure of the generated and true images, as well as suppressing the background.

The comparison between SSIM and MAE scoring mechanisms provides important insights into SHAZAM's detection framework. Using MAE for anomaly scoring achieves marginally lower AUPRC scores on SNP (decrease of 0.007) and Gran Chaco (decrease of 0.004), while showing improved performance on Lake Copeton (increase of 0.037) and Lismore (increase of 0.025). The F1 scores reveal a trade-off: MAE demonstrates improved performance on Gran Chaco (increase of 0.148) and Lismore (increase of 0.046) but shows decreased effectiveness on SNP (decrease of 0.066) and Lake Copeton (decrease of 0.011). These results validate SHAZAM's fundamental approach of comparing generated and true images, while opening up further research into more optimised scoring methods. Nevertheless, the SSIM-based structural difference module provides distinct advantages through its ability to capture local structural differences while suppressing background variations when mapping hazards.

Using a flat threshold, which removes the seasonal threshold, further demonstrates the importance of seasonal modelling within SHAZAM. The drastic reduction in F1 scores on the SNP and Gran Chaco datasets, alongside the generally reduced performance across most datasets and metrics validates this need of an adaptive threshold that accounts for seasonal variations. There is one exception which is Lismore, that sees a 0.067 reduction in the F1 score. This suggests that simpler threshold strategies might be sufficient in some instances, though this comes at the cost of reduced generalisation. Collectively, these ablation results support SHAZAM's architectural decisions while providing nuanced insights into the contribution of each component. Notably, while individual components may show advantages in specific contexts, SHAZAM's full architecture provides a theoretically grounded and empirically robust approach to self-supervised change monitoring for hazard detection and mapping.

Using a flat threshold, which removes the seasonal threshold, further demonstrates the importance of seasonal modelling within SHAZAM. The substantial reduction in F1 scores on the SNP (decrease of 0.393) and Gran Chaco (decrease of 0.260) datasets, alongside generally reduced performance across the other F1 scores and AUPRC scores, validates the need for an adaptive threshold that accounts for seasonal variations. The exception is the Lismore dataset, which shows a modest improvement (increase of 0.067) in the F1 score, suggesting that simpler threshold strategies might be sufficient in some cases, although at the cost of reduced generalisation. Collectively, these ablation results support SHAZAM's architectural design while providing deeper insights into the contribution of each component. While individual components may show advantages in specific contexts, SHAZAM's full architecture provides a theoretically grounded and empirically robust approach to self-supervised change monitoring for hazard detection and mapping.

\section{Limitations \& Future Work} \label{sec:future_work}

Although SHAZAM demonstrates strong performance for the detection and mapping of various types of hazards, several key challenges remain to create an ideal operational hazard monitoring system. A key limitation is the management of cloud coverage, as the current evaluation uses datasets with less than 10\% cloud cover. While existing methods like COLD and RaVAEn implement pixel-wise cloud masking and exclusion in their original implementations, this approach results in data loss and would greatly impair SHAZAM's structural difference module. Potential solutions include using SAR images, which can penetrate clouds and operate at night. Recent generative methods for converting SAR to multispectral images \cite{du2023MTCDN, radoi2022SARandMSCD} can enhance cloud-free multispectral SITS by increasing data availability and effective temporal resolution, while still leveraging the more detailed information available in optical images.

SHAZAM's intra-annual seasonal modelling also has limitations, as it does not capture gradual changes driven by longer-term weather patterns (e.g. El Nino and La Nina). Future work could explore modelling these longer term patterns and trends, even including multi-modal approaches that integrate additional data sources to supplement the optical imagery. The temporal frequency could also be increasing by using harmonised Landsat and Sentinel-2 data, although this challenge will naturally be reduced with more frequent image capture and the growth of remote sensing archives over time.

A third core challenge is scaling to multi-region or global coverage, an essential milestone for operational hazard monitoring. This requires models that generalise across different ROIs while maintaining a strong understanding of each region's spatial structure. Future work could explore injecting geographical coordinates alongside seasonality to model changes in land cover at diverse locations simultaneously. This video anomaly detection-esque approach could replace SIU-Net with a more scalable model, and the development of SITS foundation models could also enable generalised hazard detection while supporting other downstream tasks. However, computational efficiency should remain a critical consideration when designing scalable models, as it is a core requirement for operational deployment.

While SHAZAM effectively detects and maps hazards, its operational value would be significantly enhanced if it could also classify what the hazards are. Semi-supervised and few-shot learning methods could leverage a small number of labelled images to develop hazard classification abilities, making efficient use of limited labelled data \cite{wang2020FewShotSurvey, yang2022SemiSupervisedSurvey}. Alternatively, zero-shot learning approaches could allow classification without labelled data from large, generalisable models \cite{pourpanah2022ZeroShotReview}, such as multi-model Large Language Models (LLMs). 

A critical direction for future work is onboard real-time monitoring for hazard detection and mapping. Although satellite-based systems provide global land coverage, they require significant time for data downlink and post-processing. Performing onboard anomaly detection with less-processed data greatly reduces these delays. Additionally, satellite multi-day revisit times limit rapid response capabilities in situations requiring immediate mapping. Real-time hazard detection can leverage more agile platforms such as drones and aircraft that can be deployed rapidly. These platforms offer enhanced spatial resolution due to their proximity to the ground and can be readily equipped with advanced sensors, including hyperspectral cameras that provide rich spectral information for anomaly detection. This combination of high spatial and spectral resolution, coupled with flexible deployment, enables immediate and targeted hazard detection and mapping and opens new possibilities for real-time monitoring.

\section{Conclusion} \label{sec:conclusion}

The shift from change detection to change monitoring represents a crucial advancement in hazard detection and mapping. Existing methods often require hazard-specific labelled data, lack compatibility with irregular SITS observations, or fail to distinguish unexpected changes from normal seasonal variations. SHAZAM addresses these limitations through a self-supervised change monitoring approach that integrates seasonal modelling via SIU-Net, structural difference mapping and scoring, and a seasonally adaptive threshold. SHAZAM was evaluated alongside similar models on four datasets, which covered different regions of interest (ROIs) with bushfires, extreme/out-of-season snowfall, deforestation, algal blooms, drought, and floods.

SHAZAM demonstrated superior hazard detection performance across all four datasets, with F1 score improvements between $0.066$ and $0.234$. This was primarily through its greater effectiveness at reducing missed hazards (higher recall). The consistently strong performance across all metrics demonstrated SHAZAM's ability to balance false alarms and missed detections in diverse geographical regions, while competing models performed below the random baseline. While other models could have benefited from dataset-specific threshold optimisation, SHAZAM's theoretically grounded seasonal threshold required no optimisation and was robust across all datasets. Notably, SHAZAM achieved this performance while being extremely lightweight, using the least number of learnable parameters among all comparative models (473K).

SHAZAM demonstrated superior mapping capabilities through high spatial resolution, seasonal modelling, and the ability to suppress background features while accentuating changes caused by hazards. Although existing methods showed some success, they lacked spatial resolution, seasonal awareness, or missed mapping crucial hazard features. SHAZAM effectively mapped both immediate and gradual hazards, demonstrating versatility across various types of hazards and geographic locations.

The ablation study showed that altering individual components had both benefits and drawbacks, suggesting that further optimisation is promising. However, SHAZAM's full architecture provides a theoretically grounded and empirically robust approach to hazard detection and mapping. Future work should address remaining challenges in cloud cover management, modelling long-term environmental changes, and scaling to multi-region or global coverage. The development of real-time detection capabilities and multi-modal/LLM hazard classification offers promising directions that would enhance monitoring solutions for rapid insights into emerging hazards.

\section*{Acknowledgements}
The authors would like to thank Dr. Lichao Mou for his advice, which was instrumental in the formation and early stages of this work. The authors also thank Mr. Thomas Dujardin for his valuable technical advice and expertise in remote sensing and deep learning methods, and Dr. Pu Jin for his insights on ANDT.

\bibliographystyle{IEEEtran}
\bibliography{references}

\begin{IEEEbiography}[{\includegraphics[width=1in,height=1.25in,clip,keepaspectratio]{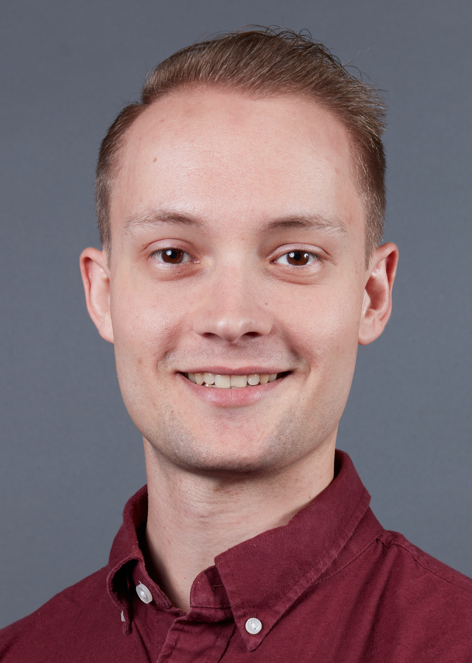}}]{Samuel Garske} 
received his Bachelor of Mechanical Engineering with Honours (2019) and his Bachelor of Commerce in business analytics (2019) from The University of Sydney, NSW, Australia. He is currently pursuing a PhD in Engineering, also at The University of Sydney. Samuel has worked as a visiting student researcher at the National Aeronautics and Space Administration (NASA) Jet Propulsion Laboratory in Pasadena, CA, U.S.A. He has also been a visiting research fellow at the Future Lab of Artifical Intelligence for Earth Observation in Munich, Germany; a joint collaboration between The Technical University of Munich (TUM) and the German Aerospace Centre (DLR). He is a student at the Australian Research Council Training Centre on CubeSats, UAVs and Their Applications (CUAVA). Sam's research interests include computer vision, anomaly detection, remote sensing, and applied machine learning for Earth observation and space applications.
\end{IEEEbiography}

\vskip -2.5\baselineskip plus -1fil

\begin{IEEEbiography} [{\includegraphics[width=1in,height=1.25in,clip,keepaspectratio]{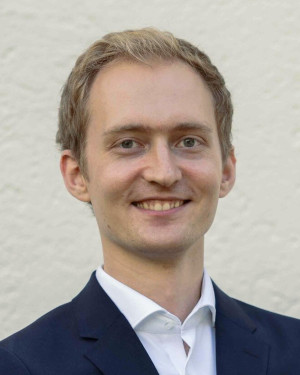}}]{Konrad Heidler} 
received the bachelor’s degree (B.Sc.) in mathematics, the
master’s degree (M.Sc.) in mathematics in data science, and the Doctorate in Engineering degree
(Dr.-Ing.) from the Technical University of Munich (TUM), Munich, Germany, in 2017, 2020, and 2024,
respectively.
He is currently a Post-Doctoral Researcher with TUM, where he is leading the working group for
visual learning and reasoning at the Chair for Data Science in Earth Observation. His research work
focuses on the application of deep learning for remote sensing in polar regions, solving reasoning tasks with deep learning, and applications of self- and semi-supervised learning in Earth observation.
\end{IEEEbiography}

\vskip -2.5\baselineskip plus -1fil

\begin{IEEEbiography} [{\includegraphics[width=1in,height=1.25in,clip,keepaspectratio]{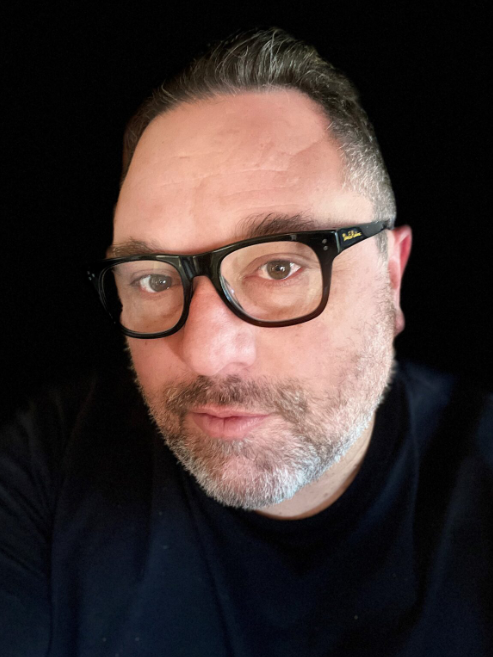}}]{Bradley Evans} 
received his Bachelor of Science with Honours in environmental science (2009), his Bachelor of Science in energy studies (2009), and his PhD in Environmental Science (2013) from Murdoch University, Western Australia, Australia. 

Bradley has served as Director of Australia’s Terrestrial Ecosystem Research Network, Ecosystem Modelling and Scaling Infrastructure, at Macquarie University and the University of Sydney. He has worked at the Australian Government Department of Defence, where he served on the AquaWatch and other committees related to Earth Observation. He has served as Director of Sydney Informatics Hub during his time at The University of Sydney. He is currently an Associate Professor in the School of Environmental and Rural Science at the University of New England, NSW, Australia since 2023. During his tenure he has established the Earth Observation Laboratory with a special focus on water and wildlife habitat (koalas), and riverine water quality. 

Bradley is an Earth observation and remote sensing specialist who is passionate about Environmental Science, Biodiversity and research applying hyperspectral imaging spectroscopy to solve real world problems. He has worked on projects associated with NASA’s OCO2 mission, working to find new ways of modelling plant growth using plant fluorescence. As a Chief Investigator of the Australian Research Council Training Centre on CubeSats, UAVs and Their Applications (CUAVA), Bradley has led the development of an OpenSource Hyperspectral Imaging Spectrometer, OpenHSI in collaboration with the Rochester Institute of Technology. He is also a community contributor to NASA JPL’s Surface Biology and Geology Study.
\end{IEEEbiography}

\newpage

%\vskip -2.5\baselineskip plus -1fil  

\begin{IEEEbiography}[{\includegraphics[width=1in,height=1.25in,clip,keepaspectratio]{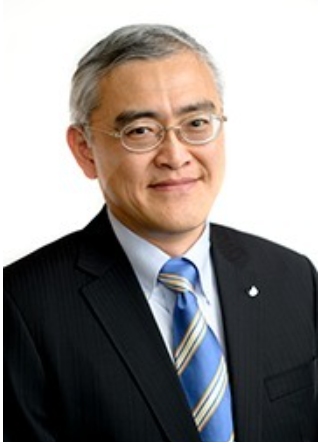}}]{KC Wong} received his Bachelor of Aeronautical Engineering (1985) and PhD in Aerospace Engineering (1994) from The University of Sydney, NSW, Australia. He has been an active full-time academic in aerospace engineering for the School of Aerospace Mechanical and Mechatronic Engineering at The University of Sydney since 1992. He is currently a Professor of Aerospace Engineering, Director for Aeronautical Engineering, and the Deputy Head of School. Prof. Wong is a pioneering UAS (Uncrewed Aircraft Systems) researcher, having worked on multidisciplinary and morphing airframe design, instrumentation, control, system integration, applications, flight testing and project management since 1988. 

He is very active in facilitating the next generation of UAS and aerospace engineers through his introduction of unique experiential learning and global engineering design opportunities. He leads a small UAS research team and has international R\&D collaborations. UAS designed and developed by his group have been used in several collaborative research projects with industry. Prof. Wong was the founding President of the industry-focused AAUS (Australian Association for Uncrewed Systems) and served in that role for seven years until 2015. He is an Associate Fellow of the AIAA (American Institute of Aeronautics and Astronautics) and a Fellow of the RaeS (Royal Aeronautical Society). He is a Chief Investigator at the Australian Research Council Training Centre on CubeSats, UAVs and Their Applications (CUAVA).
\end{IEEEbiography}

% \vskip -2.5\baselineskip plus -1fil

% \begin{IEEEbiography} [{\includegraphics[width=1in,height=1.25in,clip,keepaspectratio]{Images/Bio/Lichao.png}}]{Lichao Mou} 
% insert bio here.
% \end{IEEEbiography}

\vskip -2.5\baselineskip plus -1fil

\begin{IEEEbiography}[{\includegraphics[width=1in,height=1.25in,clip,keepaspectratio]{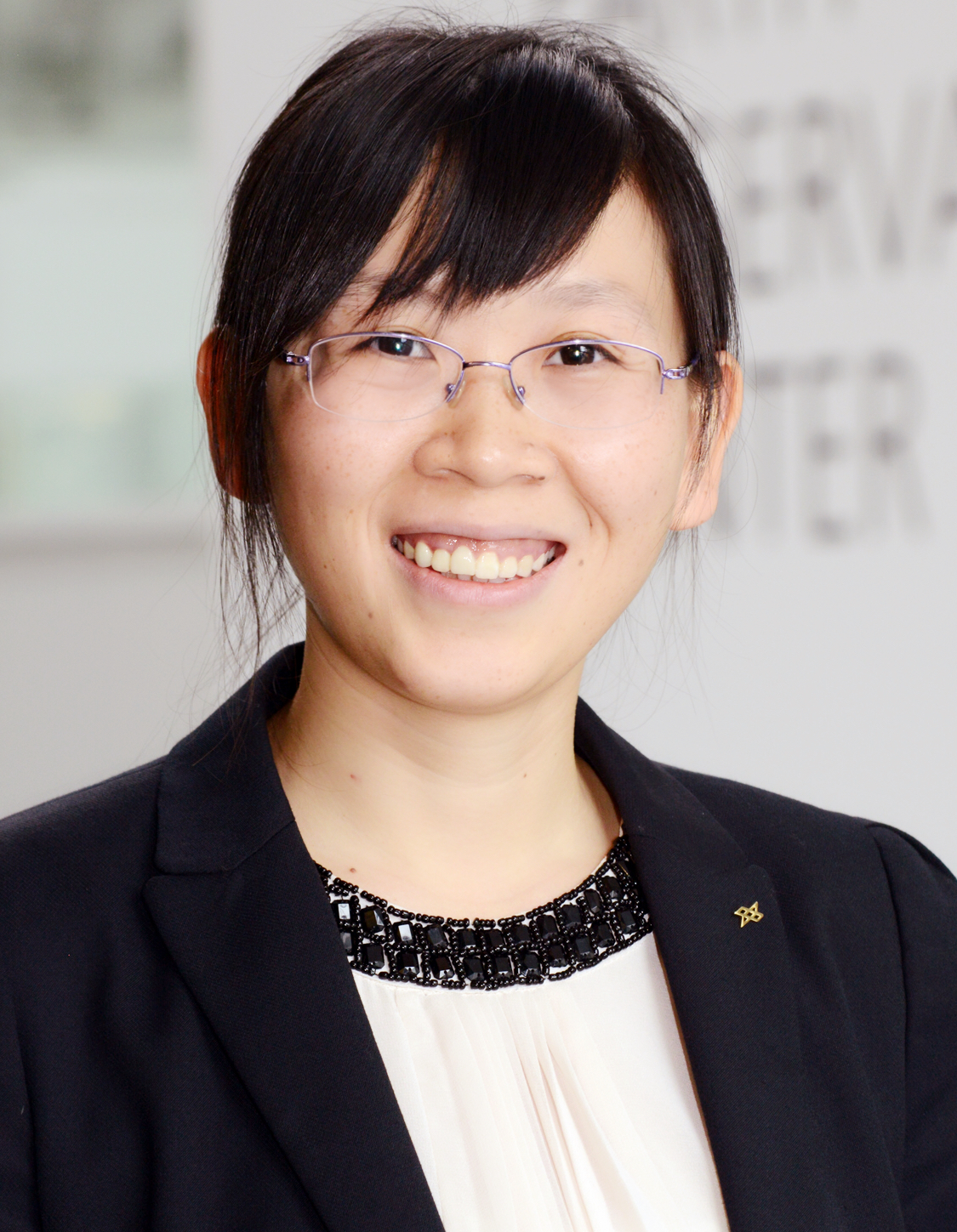}}]{Xiao Xiang Zhu}(S'10--M'12--SM'14--F'21) received the Master (M.Sc.) degree, her doctor of engineering (Dr.-Ing.) degree and her “Habilitation” in the field of signal processing from Technical University of Munich (TUM), Munich, Germany, in 2008, 2011 and 2013, respectively.
\par
She is the Chair Professor for Data Science in Earth Observation at Technical University of Munich (TUM) and was the founding Head of the Department ``EO Data Science'' at the Remote Sensing Technology Institute, German Aerospace Center (DLR). Since May 2020, she is the PI and director of the international future AI lab "AI4EO -- Artificial Intelligence for Earth Observation: Reasoning, Uncertainties, Ethics and Beyond", Munich, Germany. Since October 2020, she also serves as a Director of the Munich Data Science Institute (MDSI), TUM. From 2019 to 2022, Zhu has been a co-coordinator of the Munich Data Science Research School (www.mu-ds.de) and the head of the Helmholtz Artificial Intelligence -- Research Field ``Aeronautics, Space and Transport".  Prof. Zhu was a guest scientist or visiting professor at the Italian National Research Council (CNR-IREA), Naples, Italy, Fudan University, Shanghai, China, the University  of Tokyo, Tokyo, Japan and University of California, Los Angeles, United States in 2009, 2014, 2015 and 2016, respectively. She is currently a visiting AI professor at ESA's Phi-lab, Frascati, Italy. Her main research interests are remote sensing and Earth observation, signal processing, machine learning and data science, with their applications in tackling societal grand challenges, e.g. Global Urbanization, UN’s SDGs and Climate Change.

Dr. Zhu has been a member of young academy (Junge Akademie/Junges Kolleg) at the Berlin-Brandenburg Academy of Sciences and Humanities and the German National  Academy of Sciences Leopoldina and the Bavarian Academy of Sciences and Humanities. She is a Fellow of the Academia Europaea (the Academy of Europe). She serves in the scientific advisory board in several research organizations, among others the German Research Center for Geosciences (GFZ, 2020-2023) and Potsdam Institute for Climate Impact Research (PIK). She is an associate Editor of IEEE Transactions on Geoscience and Remote Sensing, Pattern Recognition and served as the area editor responsible for special issues of IEEE Signal Processing Magazine (2021-2023). She is a Fellow of IEEE, AAIA, and ELLIS.
\end{IEEEbiography}

\vfill

\end{document}